\documentclass[10pt]{article} % For LaTeX2e
\usepackage[preprint]{tmlr}

\usepackage{amsmath,amsfonts,bm}

\def\eqref#1{equation~\ref{#1}}
\def\1{\bm{1}}

\DeclareMathAlphabet{\mathsfit}{\encodingdefault}{\sfdefault}{m}{sl}
\SetMathAlphabet{\mathsfit}{bold}{\encodingdefault}{\sfdefault}{bx}{n}

\usepackage{booktabs}       % professional-quality tables

\usepackage{hyperref}
\usepackage{url}

\usepackage{color}
\usepackage{subcaption}
\usepackage[normalem]{ulem}
\usepackage{multirow}
\title{Meta-reinforcement learning with minimum attention}
\author{\name Shashank Gupta \email shashang@umich.edu \\
      \addr Department of EECS\\
      University of Michigan\\
      Ann Arbor, MI 48109\\\\
      \AND
      \name Pilhwa Lee \email pilhwa.lee@morgan.edu \thanks{Corresponding author}\\
      \addr Department of Mathematics\\
      Morgan State University \\
      Baltimore, MD 21251
      }

\let\AND\relax

\usepackage{mathrsfs, amsbsy, graphicx, amsmath, amssymb, algorithm, algorithmic, latexsym, amsthm}

\def\month{MM}  % Insert correct month for camera-ready version
\def\year{YYYY} % Insert correct year for camera-ready version
\def\openreview{\url{https://openreview.net/forum?id=XXXX}} % Insert correct link to OpenReview for camera-ready version

\begin{document}

\maketitle

\begin{abstract}
Minimum attention applies the least action principle in changes of control concerning state and time, first proposed by Brockett. The involved regularization is highly relevant in emulating biological control, such as motor learning. We apply minimum attention in reinforcement learning (RL) as part of the rewards and investigate its connection to meta-learning and stabilization. Specifically, model-based meta-learning with minimum attention is explored in high-dimensional nonlinear dynamics. Ensemble-based model learning and gradient-based meta-policy learning are alternately performed. Empirically, minimum attention improves fast adaptation in few shots and reduces variance from perturbations of the model and environment, compared to model-free and model-based RL baseline, and yields consistent gain when integrated into modern world models (DreamerV3, MAMBA). Furthermore, the minimum attention demonstrates an improvement in energy efficiency.
\end{abstract}

\section{Introduction}
\noindent Minimum attention considers the changes of control in state and time, first proposed by Brockett \citep{brockett1997}, with the following criterion:
\begin{equation}
\mathcal{J}(u) = \frac{1}{2} \int_0^T \int_{\Omega} \| \frac{\partial u}{\partial x} \|^2 + \| \frac{\partial u}{\partial t} \|^2 dxdt,
\end{equation}
The involved regularization is highly relevant in emulating biological control, such as motor learning. As a first step for a rigorous analysis of minimum attention, the existence of the minimum attention was shown, focused on time-variant linear feedback control in the context of motor learning \citep{lee2022}. Interestingly, a similar framework was considered in the regularization of fluid flow control by Gunzburger. In the flow control of Navier-Stokes equations, the well-posedness of boundary flow control with the changes of control in time and space was proved by the one-shot method \citep{gunzburger2000}. The minimum attention implies “regularization” and might work as stabilization in mathematical point of view, and provides a generative formalism of transition between “closed” and “open” loop controls \citep{lee2022} from the control point of view.

\noindent In this paper, we apply minimum attention to the reward of reinforcement learning and investigate its connection to meta-learning. The primary motivation is that minimum attention is a generative formalism of transition between feedforward and feedback controls while getting close to the target goals gradually in state and time \citep{lee2022}. This is conjectured to be associated with the adaptation to a new environment in the sense of meta-learning. Overall, the specific nature of the contributions is empirical in the following:
\begin{itemize}
\item Minimum attention has a significant improvement in total rewards.
\item Minimum attention has a significant reduction in variance, implying the functionality of stabilization.
\item Minimum attention has an enhancement of improved total reward in meta-testing.
\end{itemize}
% \textcolor{red}{ THIS SHOULD MOVE SOMEWHERE ELSE \noindent We emphasize that the contribution of this paper is not a new model-based RL architecture. Instead, we study minimum attention as a control-theoretic regularization framework that can be incorporated into existing reinforcement learning and model-based learning pipelines. This framing is important because the central question is whether regularizing control variation in state and time can improve learning stability, energy efficiency, and adaptation without sacrificing task reward.}
%\noindent This is indeed significant in its connection to \textcolor{blue}{robust control and learning stability.}
%
\noindent Most learning processes are through experiences, and their functionality is tested with new tasks. Deep reinforcement learning (RL) has been successful in showing competence in learning controls for complex systems \citep{mnih2015}. Many of the algorithms are model-free and take the learning process purely by observing state transitions without explicit knowledge of internal system dynamics. As a consequence, most of those algorithms require a large number of sampling and immense learning time to come to a converging stage. Model-based RL shows a promising efficiency in sampling in comparison to model-free RL, but in many cases, the converging return is lower than the asymptotic performance of model-free RL due to model bias. Consequently, the main challenges are how to tolerate the inherent model-bias, and how the learned models and control are flexible to diverse tasks in the presence of uncertainties. Here, in our scope of consideration, two major uncertainties are attributed to 1) the model is not \emph{a priori} known, and evolves in the model learning process, i.e., model-bias and 2) the model has intrinsic stochasticity or in the interaction with the environment. Model-ensemble approach with a number of models learned together shows a potential to overcome part of the issues \citep{clavera2018, kurutach2018}. In the paradigm of ``learning to learn'', meta-learning pursues adaptation of learning with few shots confronting new tasks \citep{finn2017}, showing a promising learning performance coupled to a model-ensemble approach \citep{clavera2018}.
In our strategy to handle the uncertainty of model learning, we also incorporate an ensemble of model systems in the course of model learning \citep{kurutach2018}, utilizing data collected from the real environment. 
\begin{figure}[hbt]
\begin{center}
\includegraphics[angle=0, width=1.0\textwidth]{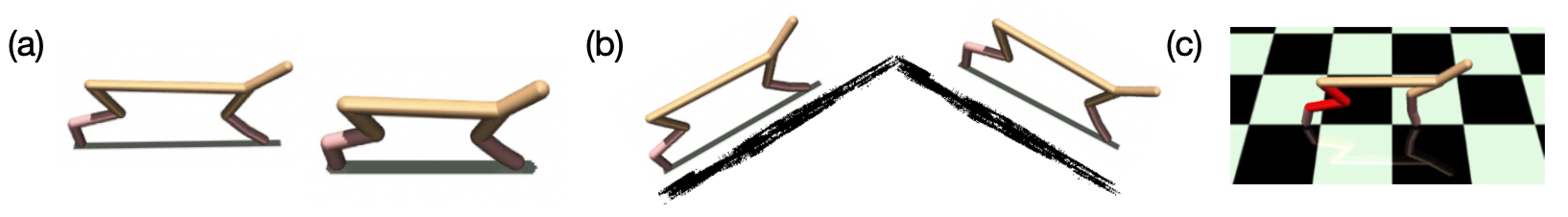}
\end{center}
\caption{Meta-learning from model and environmental perturbations. (a) Half-Cheetah with varying body masses \citep{KiminLee2020}, (b) Adaptation to uphill and downhill, (c) One hind leg crippled \citep{nagabandi2019}}
\label{figure1}
\end{figure}

\noindent We emphasize that the contribution of this paper is not a new model-based RL architecture. Instead, we study minimum attention as a control-theoretic regularization framework that can be incorporated into existing reinforcement learning and model-based learning pipelines. The central question is whether regularizing control variation in state and time can improve learning stability, energy efficiency, and adaptation without sacrificing task reward. We use MB-MPO as the primary backbone because it is a well-understood model-based meta-learning framework, suitable for controlled analysis of feedback, feedforward, variance, energy, and adaptation; to verify that the benefit is not specific to MB-MPO, we also incorporate the same regularizer into modern world-model methods, including MAMBA and DreamerV3.

%The goal of this paper is not to propose a new model-based RL architecture. Instead, we study minimum attention as a control-theoretic regularizer that can be incorporated into existing RL and model-based learning pipelines. We use MB-MPO as the primary backbone because it is a well-understood model-based meta-learning framework, making it suitable for controlled analysis of feedback, feedforward, variance, energy, and adaptation. To verify that the benefit is not specific to MB-MPO, we also incorporate the same regularizer into modern world-model methods, including MAMBA and DreamerV3. The resulting improvements support our main claim: minimum attention helps stabilize and simplify learning by reducing unnecessary control variation while preserving, and often improving, task reward.

%\noindent Overall, related works on model-based RL and meta-RL, and the associated regularizations are covered in Section 2. We formulate the ensemble model-based meta-policy learning with minimum attention in Section 3. In the main results of Section 4, the training/testing of high-dimensional dynamical systems is compared with the state-of-the art model-free and model-based RLs and analyzed in the feedback and feedforward terms and adaptation in model or environmental perturbations.

\section{Related Works}
\textbf{Model-based reinforcement learning}  Model-based RL improves sample efficiency by learning dynamics and using the learned model for policy updates, but can suffer from model bias.
%\sout{Once initially unknown or partially known models are reduced in model bias and uncertainty, and are well identified,} 
Model-based RL approaches can populate imaginary exploration and show outperforming learning curves with smaller samples for policy training. Taking a meta-learning with gradient-descent of a finite number model learning, i.e., ensemble model-based RL shows competent performances \citep{kurutach2018, clavera2018}. In the efforts of meta-learning and generalization, Lee, et al. takes the dynamics model as a global one, which is learned by characterizing a context latent vector for the local dynamics and then by taking conditional prediction \cite{KiminLee2020}. Du, et al. used neural-ODE in a model-based RL for semi-Markov decision processes in continuous-time dynamics \cite{du2020}. Recent world-model methods provide stronger model-based backbones than earlier ensemble-based approaches. DreamerV3 learns latent dynamics and trains policies through imagined trajectories across diverse domains \citep{hafner2023mastering}, while MAMBA adapts world-model learning to meta-RL settings with task variation \citep{rimon2024mamba}.\\\\
\textbf{Meta-Reinforcement learning} Model-agnostic meta-learning is by two-level gradient-based optimization in the sense of predictor-corrector schemes \citep{finn2017}. For a rapid adaptation to a new environment, it is possible to take off-line learning \citep{clavera2018} or online learning \citep{nagabandi2019} at the test stage in model-based reinforcement learning. In the off-policy actor-critic methods, meta-critic is rapidly learned online for a single task, rather than slowly over a family of tasks \citep{WeiZhou2020}. In the technique of generalization, a loss function can be extended to encourage the context latent vector to be trained for predicting both forward and backward dynamics \citep{KiminLee2020}.\\\\
\textbf{Regularization in reinforcement learning and meta-learning} Regularization is important transforming non-convex profiles to convex and sparse discreteness to continuous and differentiable ones. In machine learning, robustness and domain generalization are fundamentally correlated, but "robustness" is neither necessary nor sufficient for transferability; rather, regularization is a more fundamental perspective for understanding domain transferability \citep{XiaojunXu2022}.
In the inference of nonlinear state-space models, Corenflos, et al. overcome the issue of traditional resampling encountering non-differentiable loss functions in partial filter methods with entropic regularization \citep{Corenflos2021}. Regularization with Jacobian term stabilizes the random and adversarial input perturbations in the classification without severely degrading generalization \citep{hoffman2019}. Biased regularization overcomes complexities in distributions of tasks via a conditioning function which maps task-side information into a meta-parameter vector relevant for the specific task at hand \citep{denevi2020}. 

Regularization has been widely used in RL to improve sample efficiency, robustness, and training stability. Smooth Regularized Reinforcement Learning (SR2L) \citep{shen2020deep} encourages nearby states to induce nearby actions, effectively constraining local policy sensitivity with respect to state perturbations. Spectral-normalization \citep{bjorck2021towards} and Lipschitz-based approaches \citep{zheng2023model} instead constrain network layers or value functions to stabilize actor--critic training and reduce model-induced Bellman error in model-based RL. Randomized policy smoothing \citep{kumar2021policy} provides certified robustness against adversarial observation perturbations, while demonstration-regularized RL \citep{tiapkin2023regularized} constrains the learned policy toward an expert or behavior-cloned prior.

Minimum attention (MA) is related to these approaches but differs in its object and motivation. Rather than regularizing only the value function, network weights, observation robustness, or deviation from an external policy prior, MA directly regularizes the learned control law. Its spatial term penalizes state-dependent feedback sensitivity, while its temporal term penalizes rapid variation of the control over learning or rollout time. Thus, MA can be viewed as a control-theoretic regularizer that couples feedback and feedforward structure, rather than only a generic smoothness or Lipschitz penalty.

\section{Ensemble model and meta-policy learning with minimum attention}

\subsection{Mathematical formulation}

Given the first-order nonlinear system, $\dot{x} = f(x,u,t),$
%\begin{eqnarray}
%\dot{x} = f(x,u,t),
%\label{eqn:xdotfxu}
%\end{eqnarray}
where $x \in \mathcal{X} \subseteq \mathbb{R}^{n}$ is the state,
$u \in \mathcal{U} \subseteq \mathbb{R}^{m}$ is the control, and $f$ is
continuously differentiable with respect to $x$ and $u$. %The corresponding stochastic differential equation is $dx = f(x,u)dt + \sigma(x,t)dW_t.$
The corresponding stochastic differential equation is the following:
\begin{eqnarray}
dx = f(x,u)dt + \sigma(x,t)dW_t. \label{SDE}
\end{eqnarray}
where $\sigma(x,t)$ is the diffusion coefficient and $W_t$ is a standard Wiener process.

\subsection{Model learning}
The model
$\{ \hat{f}_{\theta_\mathcal{M}} \}$
parameterized by $\theta_\mathcal{M}$ is learned by the following loss function, similar to the approach of the probabilistic dynamics model \citep{clavera2018} the one-step prediction residual $e$ is defined as:
\begin{eqnarray}
e(X_n, X_{n+1}) = \frac{X_{n+1} - X_n}{\Delta t_n} - \hat{f}_{\theta_\mathcal{M}}(X_n, u(X_n)),\\
\mathcal{L}(\theta_\mathcal{M}) = \frac{1}{| R |} \sum_{ \{ (X_n, X_{n+1}) \} \in R} || e(X_n, X_{n+1}) ||^2, \label{loss_model}
\end{eqnarray}
where the data pool $R$ with the transitions $(X_n, X_{n+1})$ in the finite horizon is collected from the observation in the real environment.

\subsection{Meta-learning with model-ensemble}
% \subsection{Formulation of control: linearization with stochastic terms}
The control $\mathbf{u}$ is linearized in state of $\mathbf{x}$ with feedback gain of $\mathbf{K(t)}$ and feedforward term of $\mathbf{v(t)}$:
\begin{eqnarray}
\mathbf{u}(\mathbf{x}, t) &=& \mathbf{K(t)}\mathbf{x} + \mathbf{v(t)} + \epsilon, \label{control_with_noise}
\end{eqnarray}
where the stochastic term of control, $\epsilon$ is formulated by Gaussian noise or Ornstein-Uhlenbeck process \citep{uhlenbeck_ornstein}:
\begin{eqnarray}
d \mathbf{\epsilon} &=& \theta_{\rm OU} (\mathbf{\mu} - \mathbf{\epsilon}) dt + \mathbf{\sigma}_{\epsilon} dW_{\epsilon}. \label{Ornstein_Uhlenbeck}
\end{eqnarray}
where $\theta_{\rm OU}$ and $\sigma_{\epsilon}$ are the involving relaxation and stochasticity constants and $\mu$ is the mean-reversion target of the Ornstein-Uhlenbeck process. Intuitively, $K(t)$ is the state-feedback gain — how strongly the policy reacts to deviations in $x$ — while $v(t)$ is the feedforward (open-loop) bias driving the system toward the goal independent of the current state; minimum attention regularizes both terms jointly.

The control hyperparameter $\theta_u$ is constituted with the hyperparameters for $K(t)$ and $v(t)$. %$u = u_{\theta_u}(x)$ is defined by a feed-forward neural network with the parameter $\theta_u$ and deterministically by the state $x$. 
Overall, the augmented hyperparameter $\theta =\left[ \theta_u ~\theta_\mathcal{M} \right]^T$.
%\theta_u \\ 
%\theta_\mathcal{M}
%%
%\begin{eqnarray}
%\theta
%=\left[ \begin{array}{c}
%\theta_u \\ 
%\theta_\mathcal{M}
%\end{array} \right]. \label{neural_ODE}
%\end{eqnarray}
%
\noindent The meta-learning by one-step gradient-ascent in policy optimization proceeds by two steps as follows:
\begin{eqnarray}
r_{\rm reg}(x,u_{\theta}) &=& r(x,u_{\theta}) - \alpha (\| \frac{\partial u_{\theta}}{\partial x} \|^2 + \| \frac{\partial u_{\theta}}{\partial t} \|^2), \\
\mathcal{J}_i(\theta)
&=& \mathbb{E}_{\mathcal{T}_i} \left( \int_0^{T} r_{\rm reg}(x,u_{\theta}) dt \right),
\end{eqnarray}
where the sample trajectories $\mathcal{T}_i$ are generated by model ${\mathcal{M}_i}$ and the control $u_{\theta}$. 

We utilize Soft Actor-Critic (SAC) \citep{haarnoja2018} as the primary off-policy optimizer. The regularized objective follows the maximum entropy framework:
\begin{eqnarray}
%\begin{split}
    \mathcal{J}(\theta) &=& \mathbb{E}_{(s_t, u_t) \sim \mathcal{D}} [ \sum_{t} \gamma^t (r(s_t, u_t)  + \beta H(u_\theta(\cdot|s_t)) - \alpha (\| \frac{\partial u_{\theta}}{\partial x} \|^2 + \| \frac{\partial u_{\theta}}{\partial t} \|^2) ] \label{SAC_formula} \\
%\end{split}
\theta'_{u,i} &=& \theta_u + \beta \nabla_{\theta_u} \mathcal{J}_i(\theta), 
\label{eq_sac}
\end{eqnarray}
%\sout{SAC is our primary backbone, the framework is general; we have evaluated on-policy methods including Vanilla Policy Gradient (VPG) and Trust Region Policy Optimization (TRPO) \citep{schulman2015}, finding consistent performance improvements across both paradigms.}. 

% 090726 Concretely, Eq. \ref{SAC_formula} is used for the fast, per-model inner-loop adaptation described in Algorithm 1, while the outer meta-update in Eq. \ref{meta_objective} is instantiated with SAC's reparameterized actor and critic updates \citep{haarnoja2018}, trained on the pooled imaginary transitions $\{\mathcal{T}'_i\}_{i=1}^M$ generated across the ensemble.

%\textcolor{red}{We first present the inner-loop adaptation step using the vanilla policy gradient (VPG) method \citep{peters2006} for clarity of exposition, the same regularized objective is instantiated with Soft Actor-Critic (SAC) \citep{haarnoja2018} in Section 3.4, which is the optimizer used for all results reported in this paper.} Based on the vanilla policy gradient method:
%
%\begin{eqnarray}
%\nabla_{\theta_u} \mathcal{J}_i(\theta) &=& -\mathbb{E}_{\mathcal{T}_i} ( \sum_{k=0}^T \nabla_{\theta_u} \log P_i(x_k, t_k; \theta) (\sum_{l=k}^T \gamma^l r_{\rm reg}(x_l,u_{\theta}(x_l)) - b_k) ), \label{VPG} \nonumber \\
%\begin{eqnarray}
%\theta'_{u,i} &=& \theta_u + \beta \nabla_{\theta_u} \mathcal{J}_i(\theta), 
%\end{eqnarray}
%where $b_k$ is a baseline (e.g., the value function estimate) used for variance reduction. 
where $H$ is the policy entropy (derived from the stochastic term $\epsilon$ in Eq. \ref{control_with_noise}). This is the first one-step gradient-ascent for adaptation of policy. Next is integrating the return utilizing the whole model-ensemble $\{ {\mathcal{M}_i} \}$:
\begin{eqnarray}
r_{\rm reg}(x,u_{\theta'_i}(x)) &=& r(x,u_{\theta'_i}(x)) \nonumber - \alpha (\| \frac{\partial u_{\theta'_i}}{\partial x} \|^2 + \| \frac{\partial u_{\theta'_i}}{\partial t} \|^2), \nonumber \\
\max_{\theta_u} && \frac{1}{M} \sum_i^M \mathcal{J}_i (\theta') ~~ {\rm s.t.}  ~~~~
\mathcal{J}_i(\theta') = \mathbb{E}_{\mathcal{T}'_i} \left( \int_0^{T} r_{\rm reg}(x,u_{\theta'_i}(x)) dt \right) , \label{meta_objective}  
\end{eqnarray}
where the sample imaginary trajectories $\mathcal{T}'_i$ are generated by the model ${\mathcal{M}_i}$ and the adapted control $u_{\theta'_i}$. %This outer maximization is stated here in optimizer-agnostic form; Section 3.4 instantiates it concretely using SAC, whose maximum-entropy objective (Eq. \ref{eq_sac}) is trained on the pooled trajectories $\{\mathcal{T}'_i\}_{i=1}^M$ in place of the explicit $(1/M)\sum_i$ average shown above.

SAC is the backbone used for all results reported in this paper. The framework is otherwise agnostic to the underlying optimizer, we additionally verified consistent improvements when substituting on-policy methods, including VPG and Trust Region Policy Optimzation (TRPO) \citep{schulman2015}, though those results are not the ones presented in Section 4. 

\subsection{Discrete-time Approximation of Control Derivatives}
While the minimum attention criterion in Eq. (1) is defined in continuous time and state, policy optimization in discrete MDPs requires a consistent approximation. For a policy $u_{\theta}$ at state $x_{t}$, where $\theta_k$ denotes the control parameters $\theta_u$ at update step $k$, we compute the spatial and temporal derivatives as follows:

\noindent \textbf{Spatial Derivative:} The Jacobian norm is estimated via auto-differentiation through the policy network, using stochastic dimension sampling for computational efficiency (Appendix~A.7):
\begin{equation}
\| \frac{\partial u_\theta}{\partial x} \|^2 \approx \left| \nabla_x u_\theta(x_t) \right|^2_F
\end{equation}

\noindent \textbf{Temporal Derivative:} Since $\partial u_{\theta} / \partial t$ represents the change in policy over learning time for a fixed state, we use a discrete-time difference between the current policy parameters $\theta_k$ and the parameters $\theta_{k-\Delta t}$ from the last time the transition at $x_t$ was sampled from the replay buffer $\mathcal{D}$:
\begin{equation}
\| \frac{\partial u_\theta}{\partial t} \|^2 \approx \left| \frac{u(x_t; \theta_k) - u(x_t; \theta_{k-\Delta t})}{\Delta t} \right|^2
\end{equation}
where $\Delta t$ is the number of policy updates since that transition was last sampled from $\mathcal{D}$; transitions are matched by exact identity, since $\mathcal{D}$ stores discrete transitions rather than a continuous state space, so revisits correspond to the same stored transition being resampled across minibatches rather than continuous-state recurrence. This formulation ensures that the regularization effectively penalizes erratic changes in both the control landscape and the learning trajectory.

\begin{algorithm}[hbt]
\caption{Meta-learning with minimum attention}
\begin{algorithmic}
\STATE{Initialize $\theta_i = [\theta_\mathcal{M}^i ~\theta_u]$; $\theta_\mathcal{M}^i$ for model ${\mathcal{M}_i}$ and $\theta_u$ for the control.}
\STATE{The model ensemble $\mathcal{M}$ = $\{{\mathcal{M}_i}\} = \{ \hat{f}_i\}_{i=1}^M$}
\STATE{$R_i \gets \emptyset$}
\STATE{$\mathcal{D} \gets \emptyset$}
\STATE{\textit{$R_i$ holds real environment transitions for model learning; $\mathcal{D}$ pools imaginary, regularized transitions from all ensemble models for the SAC policy update.}}
\FOR{$epoch=1$ to $T_{\rm epoch}$}
\STATE{}
\COMMENT{Model learning}
\FOR{$i=1$ to $M$}
\STATE{Sample trajectories $\{ X_n^i \}$ from the environment with $[\theta_\mathcal{M}^i ~\theta'_{u,i}]$, and $R_i \gets \{ X_n^i \}$:}
\STATE{Train ${\mathcal{M}_i}$ using $R_i$ and $\mathcal{L}(\theta_\mathcal{M})$ in Eq. (\ref{loss_model}).}
\ENDFOR
\STATE{}
\COMMENT{Meta-policy learning}
\FOR{$i=1$ to $M$}
\STATE{Generate $\{ x_j^k \}$ with $x_j^0 \sim \rho_i(\cdot, 0)$ and the sample size $N$}.
\STATE{This is to sample model-based "imaginary" trajectories $\mathcal{T}_i$ from ${\mathcal{M}_i}$ using $\theta_i$.}
\STATE{$\mathcal{J}_i(\theta) \gets \frac{1}{N} \sum_{k,j} r(x_j^k,u_{\theta}(x_j^k,t_k))$}
\STATE{$\mathcal{J}_i(\theta) \gets -\frac{\alpha}{N} \sum_{k,j} (\| \frac{\partial u_{\theta}}{\partial x} \|^2 + \| \frac{\partial u_{\theta}}{\partial t} \|^2)(x_j^k,t_k)$}
\STATE{Adapt parameters $\theta'_{u,i} = \theta_u + \beta \nabla_{\theta_u} \mathcal{J}_i(\theta)$ using a single SAC.} %VPG step (Eq. \ref{VPG}) for fast adaptation.}
\STATE{Sample model-based trajectories $\mathcal{T}'_i$ from ${\mathcal{M}_i}$ using adapted $u_{\theta'_i}$:}
\STATE{Generate $\{ x_j^k \}$ with $x_j^0 \sim \rho'_i(\cdot, 0)$ and the sample size $N$.}
\STATE{$\mathcal{J}_i(\theta') \gets \frac{1}{N} \sum_{k,j} r(x_j^k,u_{\theta'_i}(x_j^k,t_k))$}
\STATE{$\mathcal{J}_i(\theta') \gets -\frac{\alpha}{N} \sum_{k,j} (\| \frac{\partial u_{\theta'}}{\partial x} \|^2+\| \frac{\partial u_{\theta'}}{\partial t} \|^2)(x_j^k,t_k)$}
\STATE{Attach the regularized reward $\mathcal{J}_i(\theta')$ to each pooled transition and store in the SAC replay buffer $\mathcal{D}$.}
\ENDFOR
% \STATE{Train $\theta_u$ using $\frac{1}{M} \sum_i \mathcal{J}_i (\theta')$ in Eq. (\ref{meta_objective}) with TRPO.}
\STATE Update policy parameters $\theta_u$ via SAC (Eq. \ref{meta_objective}). % (Eq. \ref{eq_sac}), training on the pooled imaginary transitions $\{\mathcal{T}'_i\}_{i=1}^M$ across the ensemble.
%\STATE \textit{Note: The inner-loop adaptation uses a single VPG step for computational efficiency; the outer meta-update uses SAC (Eq. \ref{eq_sac}), the optimizer used for all results reported in this paper. This two-tier structure was also verified to work with TRPO substituted for SAC in the outer loop.}
\ENDFOR
\end{algorithmic}
\label{Monte_Carlo_method} \label{algorithm1}
\end{algorithm}

\section{Main Results and Discussion}
We explore the performance of the proposed ensemble model-based meta-learning with minimum attention to show the following:
\begin{itemize}
\item Is the proposed algorithm competent in the required training sampling in comparison to other model-based RL?
\item Does the proposed algorithm reduce the variance in learning curves? 
\item Does the proposed algorithm improve generalization in comparison to other model-based meta-policy RL? 
\end{itemize}
\noindent We conduct a series of experiments across several MuJoCo environments to rationalize the proposed method. The evaluation is designed to demonstrate that the minimum attention regularization 1) improves asymptotic performance and learning stability, 2) induces policies that are smoother and more energy-efficient by learning a more structured control strategy, and 3) enhances an agent's ability to rapidly adapt to out-of-distribution (OOD) perturbations.

\noindent We present the sensitivity, ablation, and maturation analysis for HalfCheetah in detail; headline comparison results are reported for all four agents in Section 4.2, with corresponding maturation analysis for Hopper, Walker2D, and Humanoid provided in the appendix.
\subsection{Experimental setup: high-dimensional dynamical systems}

\noindent Since our focus is the regularizer rather than a new model-based architecture, we select MB-MPO as the primary experimental backbone. MB-MPO is well suited for this purpose because it explicitly separates model learning, policy adaptation, and meta-policy optimization, allowing us to analyze how minimum attention affects feedback sensitivity, feedforward variation, energy, and adaptation under controlled perturbations. Our method integrates the minimum attention regularization into the actor-critic updates of MB-MPO \citep{clavera2018}, using SAC \citep{haarnoja2018} for policy optimization. We compare against the original MB-MPO and a purely model-free SAC as baselines. Additionally, we evaluate the compatibility of our regularizer with contemporary world models, specifically MAMBA \citep{rimon2024mamba} and DreamerV3 \citep{hafner2023mastering}, to test whether the regularizer is complementary to newer architectures.

\noindent We mainly explore motor learning and control of four agents of MuJoCo in the OpenAI Gym \citep{brockman2016openai}: Half-Cheetah (17-dim states and 6-dim controls), Walker2D (17-dim states and 6-dim control), Hopper (11-dim states and 3-dim controls), and HumanoidTruncated (45-dim states and 17-dim controls). These four Gym environments are used for the core MB-MPO comparisons. The MAMBA and DreamerV3 integration experiments in Section 4.6 instead use the DeepMind Control suite \citep{tassa2018deepmind}, following the task conventions of \citet{rimon2024mamba} and \citet{hafner2023mastering} respectively (Humanoid-dir, HalfCheetah-run). We note this distinction explicitly since Gym and DMC differ in episode and reward conventions despite sharing the MuJoCo physics engine. The governing equations for MuJoCo dynamics are formulated by Eqs. (\ref{SDE}, \ref{control_with_noise}, \ref{Ornstein_Uhlenbeck}). 
\noindent The involved learning processes are alternated between model learning and meta-policy learning, and with or without regularization of minimum attention laws. In regard to hyperparameters, key hyperparameters for the Model-Based Meta-Policy Optimization (MB-MPO) with minimum attention implementation and the baseline are described in Appendix A. In general evaluation, we have used 10 random seeds for training episodes. For the final evaluation of a trained agent, we run 10 episodes per seed, i.e., with the randomized initial configuration. Unless otherwise specified, results are averaged over 10 random seeds, with error bands and $\pm$ values indicating one standard deviation. A comprehensive description of all network architectures and hyperparameters is provided in Appendix A. Average total reward, standard deviation/error of reward across seeds/episodes are shown in Figure \ref{figure2} and \ref{meta_training_different_models}.\\ 

%\noindent In the meta-training, to analyze the effect of attention weight, we trained agents with attention weight values $\alpha \in  \{0, 0.01, 0.05, 1.0, 5.0\}$ while keeping all other hyperparameters same. In order to analyze the effects of the number of ensemble models, $M$, we trained 4 different sizes of $M \in \{1, 5, 15, 20\}$. To characterize the effect of the imaginary trajectory size $N$ in the MB-MPO inner loop, we trained the motor learning with 4 different values of $N \in \{16, 128, 256, 512\}$ (Figure \ref{model_ensemble_imaginary}).\\

\noindent In regards to specific metrics,  1) average feedback norm: during evaluation, we run the trained agent for 10 evaluation episodes and average $|| \partial u / \partial x ||^2$ per step, 2) average feedforward norm: define as the averaged $|| \partial u / \partial t ||^2$ over 10 evaluation episodes, 3) average energy: define total sum $|| u^2 ||$ per step as $``$control cost$"$. Their time courses are shown in Figure \ref{time_profile_half_cheetah}. Heatmaps of the Jacobian norm (Figure \ref{heatmap_feedback}, \ref{fig:hopper_heatmap}, \ref{fig:walker2d_heatmap}, \ref{fig:humanoid_heatmap}) and jerk norm (Figure \ref{heatmap_feedforward}, \ref{fig:hopper_heatmap_jerk}, \ref{fig:walker2d_heatmap_jerk}, \ref{fig:humanoid_heatmap_jerk}) between the baselines are shown to qualitatively judge the impact of minimum attention regularization.\\

\noindent In the meta-testing, experiments focus on out-of-distribution tasks and fast adaptation during meta-loop test time evaluation. For each of these experiments, we note the total reward, standard deviation between runs, average feedback norm, average feedforward norm, and average used energy. With the meta-trained agent, we have evaluated the impact of minimum attention regularization on the learning process by conducting qualitative analysis by visualizing policy sensitivity across different observation of the state with a specific dimension of the environment. The goal is to pick the dimensions that are critical for their respective locomotion and balance challenges. Specifically, we have picked Half-Cheetah, and three projections are plotted by heatmap: Torso height versus forward velocity, Forward velocity versus vertical velocity, and Forward velocity versus angular velocity. Similarly, the projected dimensions for other agents are specified in Appendix A. 
\subsection{Comparison with state-of-the-art RL algorithms}
\begin{figure}[t]
\centering
\begin{subfigure}{0.45\textwidth}
% \begin{subfigure}{0.35\textwidth}
%\begin{subfigure}[b]{0.49\linewidth}  
    \centering
    \includegraphics[width=\linewidth]{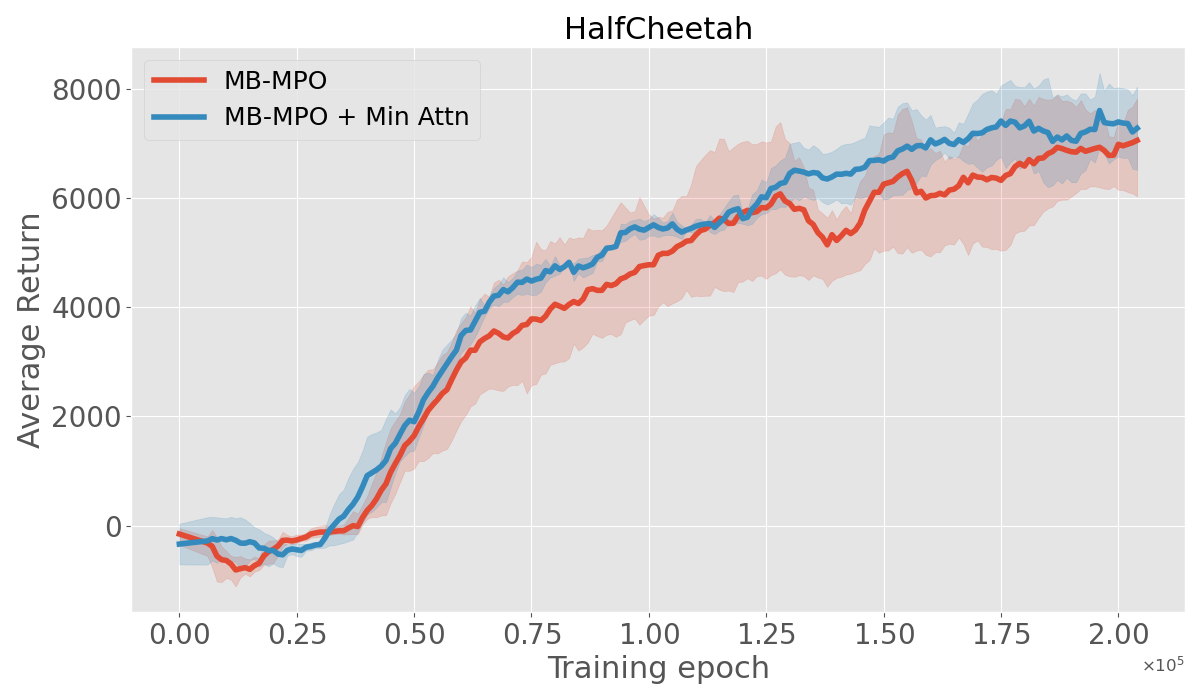}
    \label{fig:figure2a}
\end{subfigure}
\hfill 
\begin{subfigure}{0.45\textwidth}
% \begin{subfigure}{0.35\textwidth}
%\begin{subfigure}[b]{0.49\linewidth}  
    \centering
    \includegraphics[width=\linewidth]{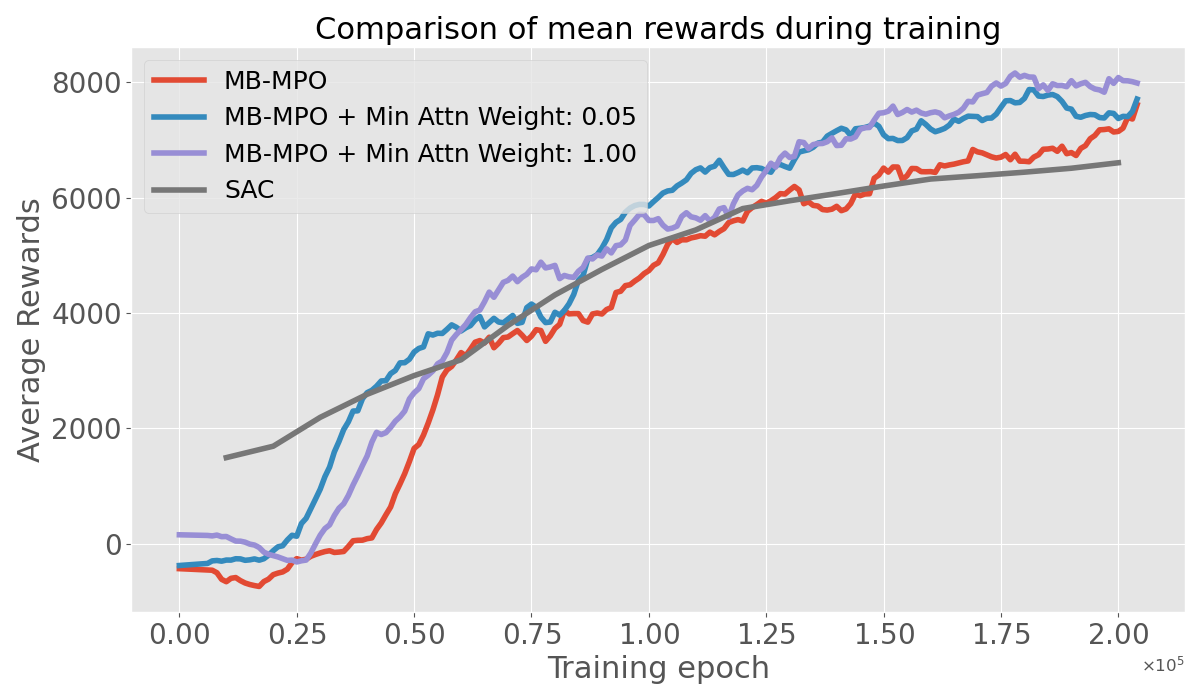} 
    \label{fig:figure2b}
\end{subfigure}

\caption{(a) Training of Half-Cheetah. There is a reduction of variance from minimum attention, (b) Comparison of learning curves among model-free RL (SAC, \citep{haarnoja2018}) and model-based RL (MB-MPO, \citep{clavera2018}, the proposed MB-MPO + minimum attention, weighting factor $\alpha = 0.05$ and $1$). The experiment is run with 10 episodes.}
\label{figure2}
\end{figure}

\begin{figure}[bt]
\centering
\begin{subfigure}[b]{0.32\textwidth}
% \begin{subfigure}[b]{0.48\textwidth} 
    \centering
    \includegraphics[width=\textwidth]{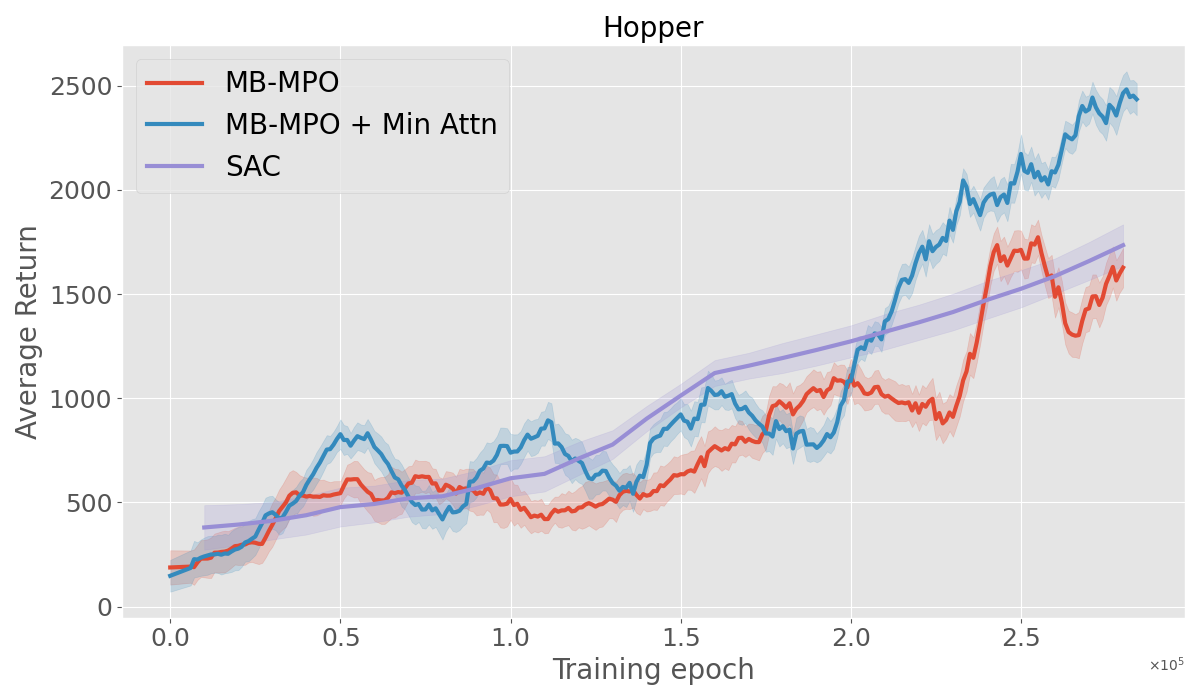}
    \label{fig:figure7a}
\end{subfigure}
\hfill 
\begin{subfigure}[b]{0.32\textwidth}
% \begin{subfigure}[b]{0.48\textwidth}  
    \centering
    \includegraphics[width=\textwidth]{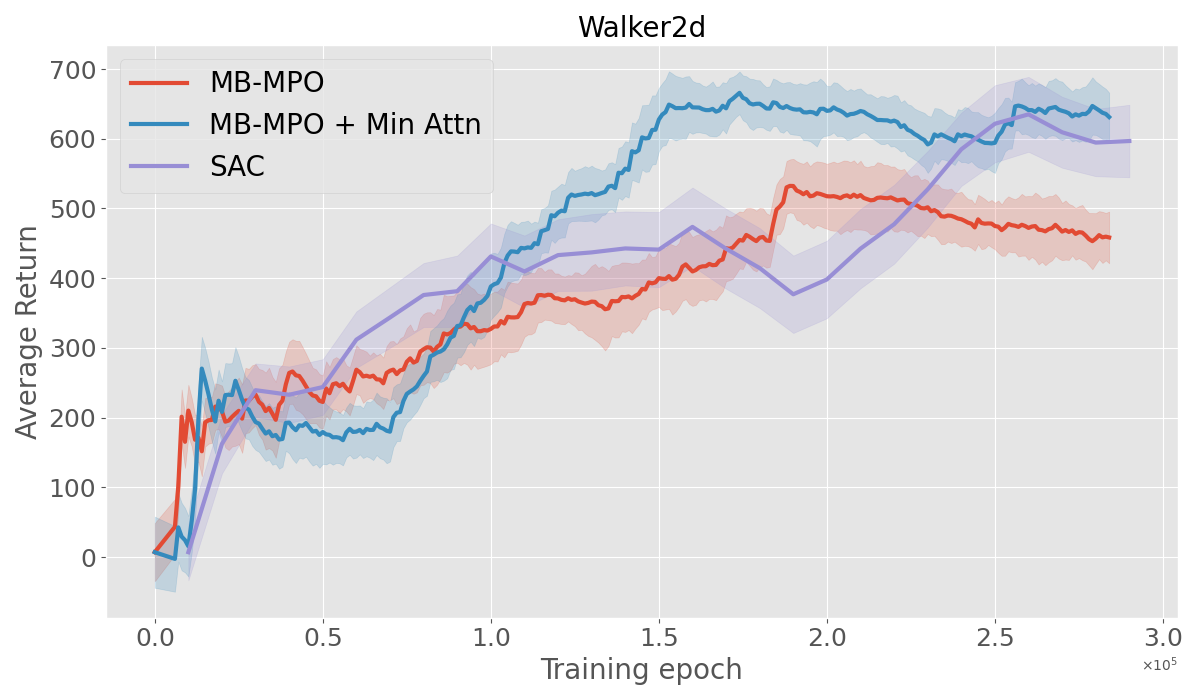} 
    \label{fig:figure7b}
\end{subfigure}
\hfill 
\begin{subfigure}[b]{0.32\textwidth}
% \begin{subfigure}[b]{0.48\textwidth}  
    \centering
    \includegraphics[width=\textwidth]{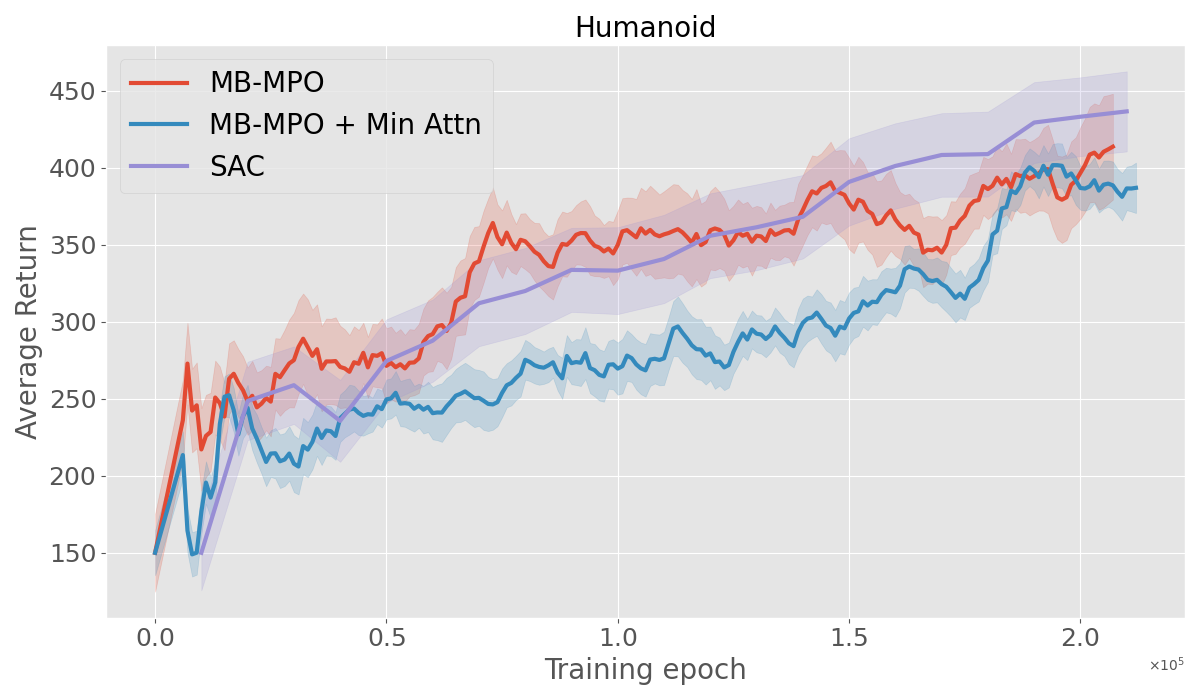} 
    \label{fig:figure7c}
\end{subfigure}
\caption{Meta-Training with different models, (a) Hopper, (b) Walker2D, (c) Humanoid. The model-based RL (MB-MPO) is employed with and without minimum attention ($\alpha = 1.0$) and compared with SAC.}
\label{meta_training_different_models}
\end{figure}

The comparison is mostly based on the performance on standard locomotion tasks. We consider a model-free RL, SAC (Soft Actor-Critic) \citep{haarnoja2018} and a model-based RL, MB-MPO (Model-Based Meta-Policy Optimization) \citep{clavera2018}. Firstly, we show the training of Half-Cheetah (Figure \ref{figure2}) in comparison of learning curves among the model-free RL (SAC), the model-based RL (MB-MPO), and the proposed MB-MPO with minimum attention, where the weighting factor is $\alpha = 0.01$, $0.05$, and $1$. We have experienced a statistically significant reduction of variance from minimum attention. To demonstrate the generality of our approach, Figure \ref{meta_training_different_models} presents the learning curves for Hopper, Walker2D, and Humanoid. For Hopper and Walker2D, our method (MB-MPO + minimum attention, blue) converges to a higher average return than the vanilla MB-MPO baseline (red). For the more complex Humanoid task, training reward is comparable between the two (Table \ref{humanoid:meta_training}: $498 \pm 38$ vs.\ $575 \pm 177$), but minimum attention exhibits substantially lower variance throughout training, visible as a consistently tighter error band in Figure \ref{meta_training_different_models}c. This variance reduction translates into a large advantage at meta-test time: minimum attention nearly doubles the meta-testing reward on Humanoid ($580 \pm 34$ vs.\ $315 \pm 67$; Table \ref{humanoid:meta_testing}) while again showing much lower variance, suggesting that the more conservative, lower-variance training policy generalizes substantially better under out-of-distribution embodiment changes.

% \noindent The outperformance of MB-MPO with minimum attention is quantified in numbers in Table \ref{table:sota_and_ablation}, \ref{table:total_reward_feedback_feedforward} and \ref{overall_comaprison}. Table \ref{table:total_reward_feedback_feedforward} provides a detailed quantitative summary of the final performance on HalfCheetah after 200K iterations. In the standard meta-training setting, the best minimum attention factor ($\alpha=1.0$) achieves a 45\% higher reward than MB-MPO, while simultaneously reducing the average feedback norm (Jacobian) by 28\% and energy consumption by 15\%. This demonstrates that the regularization does not trade performance for smoothness, but rather enables the discovery of policies that are both higher-performing and more efficient. 
% %Figure \ref{figure2} presents the learning curves on HalfCheetah, comparing our method against model-based (MB-MPO) and model-free (SAC) baselines. 
% Our approach not only matches the sample efficiency of MB-MPO but also converges to a significantly higher average reward. Notably, the learning curves of our method also exhibit reduced variance across episodes, suggesting a more stable and reliable learning process, a key benefit for robust control. 
%

\noindent The outperformance of MB-MPO with minimum attention is quantified in numbers in Tables \ref{table:total_reward_feedback_feedforward}, \ref{overall_comaprison}, and \ref{table:sota_and_ablation}. Table \ref{table:total_reward_feedback_feedforward} provides a detailed quantitative summary of the final performance on HalfCheetah after 200K iterations across three regularization strengths ($\alpha = 1.0, 0.01, 0.05$), and Table \ref{overall_comaprison} summarizes total reward across all four environments in both meta-training and meta-testing. In the standard meta-training setting, the best minimum attention factor ($\alpha=1.0$) achieves a 45\% higher reward than MB-MPO on HalfCheetah, while simultaneously reducing the average feedback norm (Jacobian) by 28\% and energy consumption by 15\%. This demonstrates that the regularization does not trade performance for smoothness, but rather enables the discovery of policies that are both higher-performing and more efficient. Our approach not only matches the sample efficiency of MB-MPO but also converges to a significantly higher average reward. Notably, the learning curves of our method also exhibit reduced variance across episodes, suggesting a more stable and reliable learning process, a key benefit for robust control.

\begin{table*}[hbt]
\caption{Half-Cheetah: \textbf{Meta-training}: total reward, feedback ($||\partial u / \partial x||^2$, Jacobian), feedforward ($||\partial u/ \partial t||^2$, jerk), and energy ($||u||^2$) at final stage of iterations (200K). Comparison of the model-based RL (MB-MPO) without and with regularization of minimum attention. The bold typeset represents the best outperformance in the variation of $\alpha$. \textbf{Meta-testing}: the agent of Half-Cheetah, one leg crippled, at final stage of iterations (200K).}
\label{table:total_reward_feedback_feedforward}
\begin{center}
  \small
  \begin{tabular}{|l|c|c|c|c|}
    \hline    
           & MB-MPO & Ours ($\alpha = 1.0$) & Ours ($\alpha = 0.01$) & Ours ($\alpha = 0.05$)    \\
    \hline
    \textbf{Meta-training} & & & &\\
    \hline
      Average total rewards   &  6692  $\pm$ 318 & \textbf{9721 $\pm$ 128} & 7545 $\pm$ 123 & 8385 $\pm$ 179  \\
     \hline
     Average feedback normalized  & 783 $\pm$ 135 & \textbf{567 $\pm$ 53} & 590 $\pm$ 88 & 637 $\pm$ 56  \\
    \hline
     Average feedforward normalized  & 5.16 $\pm$ 0.11 & \textbf{5.01 $\pm$ 0.07} & 5.02 $\pm$ 0.10 & 5.05 $\pm$ 0.07     \\
    \hline
     Average energy normalized  & 4.54 $\pm$ 0.13 & \textbf{3.85 $\pm$ 0.05} & 3.94 $\pm$ 0.03 & 3.91 $\pm$ 0.02     \\
    \hline
    \textbf{Meta-testing}  & & & &\\
    \hline
      Average total rewards   &  6356  $\pm$ 132 & \textbf{6822 $\pm$ 109} & 6514 $\pm$ 133 & 6590 $\pm$ 125  \\
     \hline
     Average feedback normalized  & 673 $\pm$ 75.6 & \textbf{625 $\pm$ 34.7} & 650 $\pm$ 77.4 & 720 $\pm$ 53.9  \\
    \hline
     Average feedforward normalized  & 7.03 $\pm$ 0.04 & 6.76 $\pm$ 0.04 & \textbf{4.87 $\pm$ 0.04} & 7.07 $\pm$ 0.04     \\
    \hline
%     Average energy normalized  & 372 $\pm$ 2.71 & \textbf{366 $\pm$ 1.48} & 334 $\pm$ 1.12 & 359 $\pm$ 1.58     \\
         Average energy normalized  & 3.72 $\pm$ 0.027 & 3.66 $\pm$ 0.015 & \textbf{3.34 $\pm$ 0.011} & 3.59 $\pm$ 0.016     \\      
         \hline   
  \end{tabular}
\end{center}
\end{table*}
\begin{table}[hbt]
\caption{Comparison of the model-based RL (MB-MPO) with regularization of minimum attention in meta-training and testing. Total reward at the epoch of 200K. For Hopper, the thigh is paralyzed. For Walker2D, the right thigh is paralyzed.} \label{overall_comaprison}
\begin{center}
  \small
  \begin{tabular}{|l|c|c|c|c|}
    \hline
           & Half-Cheetah & Hopper &  Walker2D  & Humanoid\\
    \hline
      MB-MPO training &  6692  $\pm$ 318 & 2475 $\pm$ 3.81 & 2399 $\pm$ 694  & \textbf{575 $\pm$ 177}\\
     \hline
      MB-MPO meta-testing  & 6356 $\pm$ 132 & 467 $\pm$ 100 & 523 $\pm$ 174 & 315 $\pm$ 67\\
    \hline
      MB-MPO + Min Attn training & \textbf{9721 $\pm$ 128} & \textbf{2825 $\pm$ 1.25} & \textbf{3038 $\pm$ 148} & 498 $\pm$ 38 \\
    \hline
      MB-MPO + Min Attn meta-testing & \textbf{6822 $\pm$ 109} & \textbf{485 $\pm$ 61} & \textbf{1123 $\pm$ 152} & \textbf{580 $\pm$ 34} \\
    \hline     
  \end{tabular}
\end{center}
\end{table}
\noindent To address the rapid advancement of the field, we integrated our regularizer into modern world models, including \textbf{MAMBA} and \textbf{DreamerV3}. Table \ref{table:sota_and_ablation} (right) summarizes final-stage performance for both integrations, evaluated on the DeepMind Control suite tasks Humanoid-dir (MAMBA) and HalfCheetah-run (DreamerV3), following the task conventions of \citet{rimon2024mamba} and \citet{hafner2023mastering} respectively. On MAMBA, minimum attention improves final reward on Humanoid-dir from $2453 \pm 110$ to $2553 \pm 76$ at 30M steps, with the largest relative gain early in training ($467 \pm 288$ vs.\ $759 \pm 231$ at 2.5M steps; Appendix Table \ref{table:appendix_mamba}), suggesting the regularizer also accelerates early exploration in a latent-space world model, not only in the ensemble-based MB-MPO setting. We further evaluate DreamerV3 integration across three DeepMind Control tasks: HalfCheetah-run, reacher-hard, and finger-turn-hard. Minimum attention improves final-stage reward on HalfCheetah-run ($575 \pm 123$ vs.\ $625 \pm 77$ at 500K steps) and shows a substantially larger improvement on finger-turn-hard ($826 \pm 40$ vs.\ $906 \pm 20$ at 500K steps), a roughly 10\% gain in mean reward alongside a near-halving of variance. On reacher-hard, minimum attention shows a modest reduction in variance ($20$ vs.\ $17$ std) but does not improve, and is nominally slightly below, final-stage mean reward ($942 \pm 20$ vs.\ $931 \pm 17$ at 500K steps). Taken together, these results indicate that minimum attention is complementary to DreamerV3 on two of the three tasks tested, with the magnitude and direction of benefit varying by task rather than being uniform across domains.

\noindent Section \ref{sec:ablations} presents ablation studies verifying the necessity of both components of the minimum attention regularizer and characterizing its sensitivity to ensemble size and imaginary rollout length.

\begin{table*}[t]
\centering
\scriptsize
\setlength{\tabcolsep}{3.2pt}
\renewcommand{\arraystretch}{1.05}

\begin{minipage}[t]{0.49\textwidth}
  \vspace{0pt}
  \centering
  \resizebox{\linewidth}{!}{%
  \begin{tabular}{llcc}
    \toprule
    \multicolumn{4}{c}{\textbf{HalfCheetah main results}} \\
    \midrule
    \textbf{Setting} & \textbf{Metric} & \textbf{MB-MPO} & \textbf{Ours (MA)} \\
    \midrule
    Meta-train 
      & Reward $\uparrow$ & 6692 $\pm$ 318 & \textbf{9721 $\pm$ 128} \\
      & Feedback $\downarrow$ & 783 $\pm$ 135 & \textbf{567 $\pm$ 53} \\
      & Energy $\downarrow$ & 4.54 $\pm$ 0.13 & \textbf{3.85 $\pm$ 0.05} \\
    \midrule
    Meta-test 
      & Reward $\uparrow$ & 6356 $\pm$ 132 & \textbf{6822 $\pm$ 109} \\
      & Feedback $\downarrow$ & 673 $\pm$ 75.6 & \textbf{625 $\pm$ 34.7} \\
%      & Energy $\downarrow$ & 372 $\pm$ 2.71 & \textbf{366 $\pm$ 1.48} \\
      & Energy $\downarrow$ & 3.72 $\pm$ 0.27 & \textbf{3.66 $\pm$ 0.15} \\

    \bottomrule
  \end{tabular}
  }
\end{minipage}
\hfill
\begin{minipage}[t]{0.48\textwidth}
\vspace{0pt}
  \centering
  \resizebox{\linewidth}{!}{%
  \begin{tabular}{lccc}
    \toprule
    \multicolumn{4}{c}{\textbf{Modern world models}} \\
    \midrule
    \textbf{Task} & \textbf{Steps} & \textbf{Baseline} & \textbf{+ MA} \\
    \midrule
    Humanoid (MAMBA) & 30M & 2453 $\pm$ 110 & \textbf{2553 $\pm$ 76} \\
    HalfCheetah (DreamerV3) & 500K & 575 $\pm$ 123 & \textbf{625 $\pm$ 77} \\
    reacher-hard (DreamerV3) & 500K & \textbf{942 $\pm$ 20} & 931 $\pm$ 17 \\
    finger-turn-hard (DreamerV3) & 500K & 826 $\pm$ 40 & \textbf{906 $\pm$ 20} \\
    \bottomrule
  \end{tabular}
  }

  \vspace{0.15em}

  % \resizebox{\linewidth}{!}{%
  % \begin{tabular}{lccc}
  %   \toprule
  %   \multicolumn{4}{c}{\textbf{HalfCheetah ablation}} \\
  %   \midrule
  %   \textbf{Metric} & \textbf{Full MA} & \textbf{Jac.-only} & \textbf{Temp.-only} \\
  %   \midrule
  %   Reward $\uparrow$ & \textbf{9721 $\pm$ 128} & 6484 $\pm$ 148 & 6475 $\pm$ 135 \\
  %   Feedback $\downarrow$ & 567 $\pm$ 53 & \textbf{128 $\pm$ 24.5} & 736.2 $\pm$ 84.9 \\
  %   Feedforward $\downarrow$ & 5.01 $\pm$ 0.07 & 9.54 $\pm$ 0.35 & \textbf{4.94 $\pm$ 0.04} \\
  %   \bottomrule
  % \end{tabular}
  % }
\end{minipage}

\vspace{-0.2em}
\caption{
Summary of main results. Left: MA improves HalfCheetah reward while reducing feedback and energy in meta-training and crippled-back meta-testing. Right: MA improves MAMBA and DreamerV3 baselines.
}
\label{table:sota_and_ablation}
\end{table*}

\subsection{Sensitivity Analysis: Feedback and Feedforward Control Strategies}
To understand how minimum attention shapes the learned policy, we analyze the resulting control strategy from two complementary perspectives. The spatial term $\|\partial u/\partial x\|^2$ measures state-dependent feedback sensitivity, while the temporal term $\|\partial u/\partial t\|^2$ measures temporal variation in the control signal. Together, these quantities describe how the policy allocates reactive feedback and feedforward control over states and time. To further interpret the learned controller, we analyze the decomposition $u(x,t)=K(t)x+v(t)$, where $K(t)$ captures state-dependent feedback and $v(t)$ captures the feedforward bias. Compared with the vanilla policy, the MA-regularized policy shows narrower dispersion in both $\|K(t)\|_F$ and $\|v(t)\|_2$, suggesting lower variability in the learned control structure. Importantly, MA does not simply suppress feedback: localized increases in $\|K(t)\|_F$ remain during early and transient phases, indicating that the policy can still apply corrective feedback when needed. The feedforward term shows a less monotonic pattern, but its variability is also more constrained. Overall, this supports the view that MA organizes feedback and feedforward control into a less variable but still adaptive policy.

\begin{figure*}[hbt]
    \centering
    \includegraphics[width=0.95\textwidth]{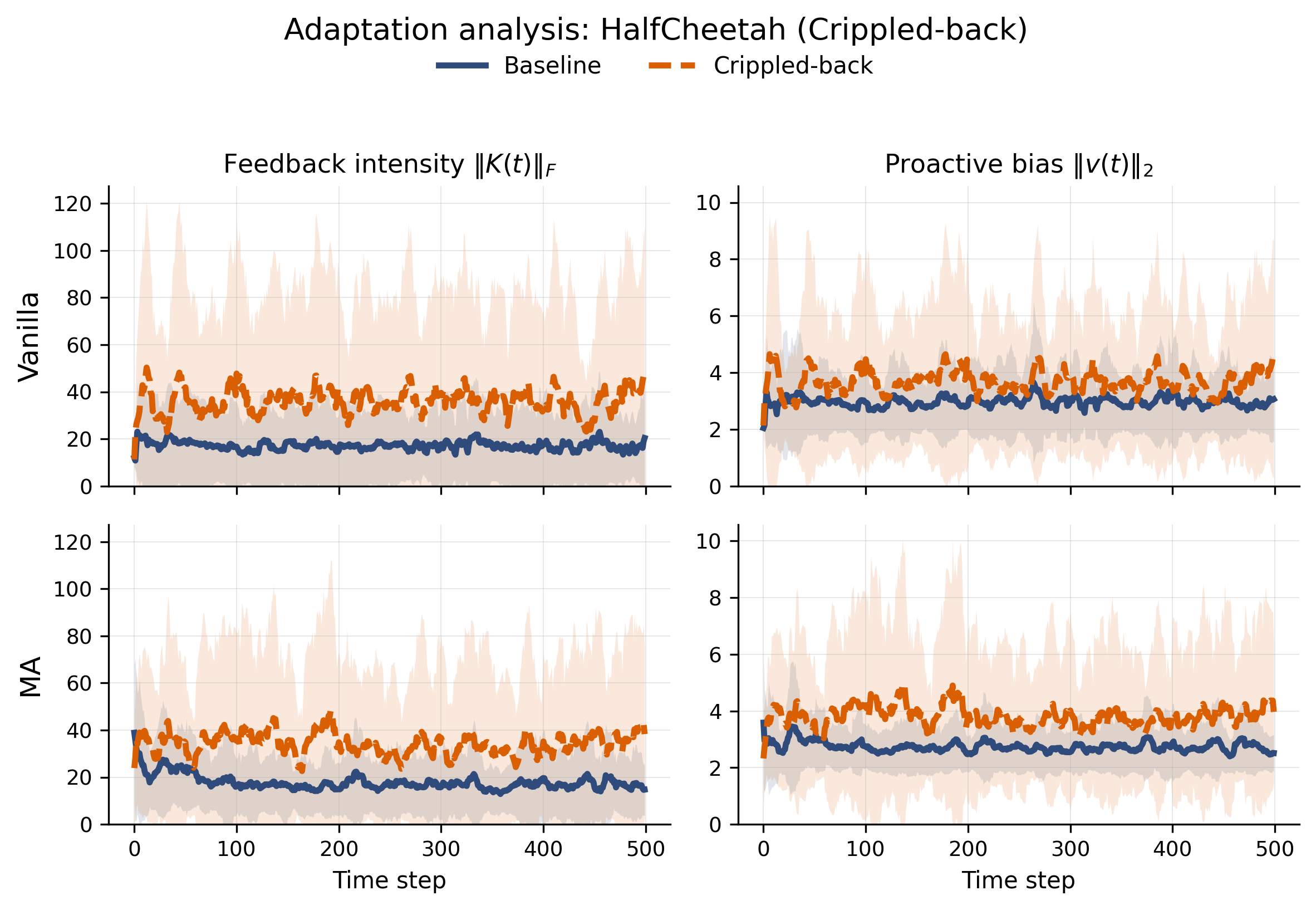}
    \caption{
    Control decomposition $u(x,t)=K(t)x+v(t)$ under crippled-back perturbation, comparing vanilla and MA policies through $\|K(t)\|_F$ and $\|v(t)\|_2$.
    }
    \label{fig:kv_adaptation_crippled}
\end{figure*}

\noindent This same decomposition remains informative under distribution shift. Figure \ref{fig:kv_adaptation_crippled} shows the control decomposition $u(x,t)=K(t)x+v(t)$ under a crippled-back perturbation, comparing vanilla and MA policies through $\|K(t)\|_F$ and $\|v(t)\|_2$; the MA policy retains the same narrower, more structured dispersion observed in the unperturbed setting even as the perturbation forces both policies to adjust their feedback and feedforward effort. Additional maturation and mass-perturbation decompositions are provided in Appendix \ref{app:kv_additional}.

\begin{figure}[bt]
\centering
\begin{subfigure}[b]{0.49\textwidth}
% \begin{subfigure}[b]{0.3\textwidth}  
    \centering
    \includegraphics[width=\textwidth]{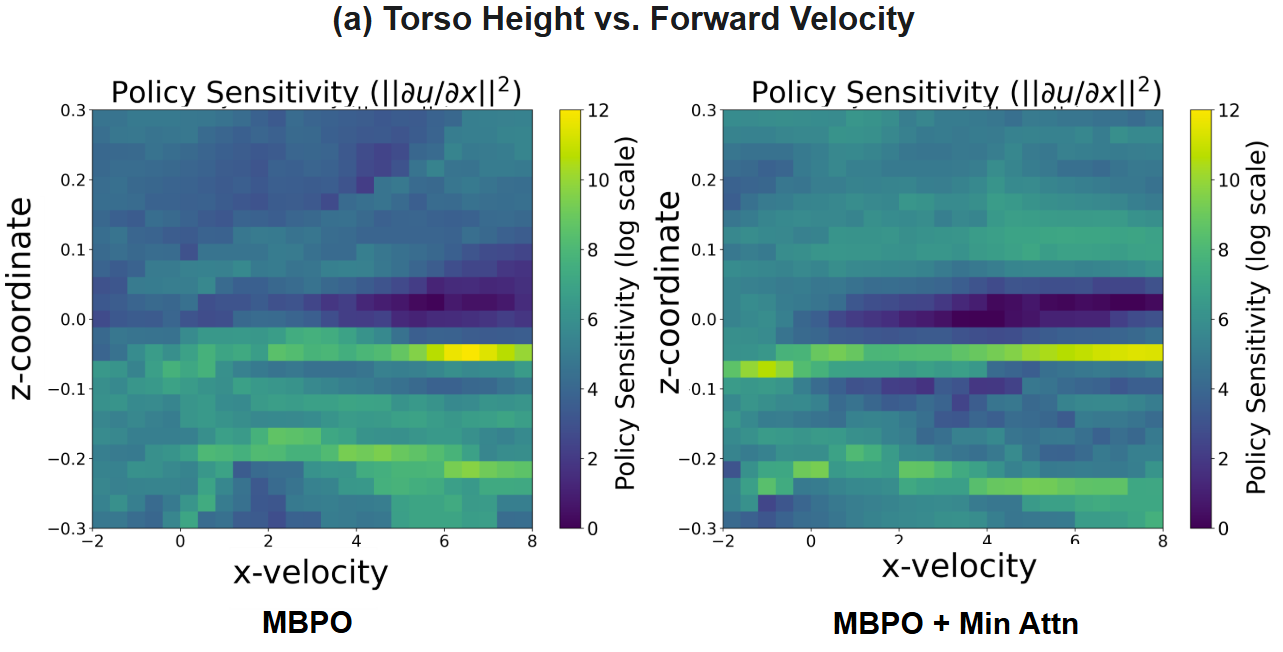}
\end{subfigure}
\hfill 
\begin{subfigure}[b]{0.49\textwidth}
% \begin{subfigure}[b]{0.3\textwidth}  
    \centering
    \includegraphics[width=\textwidth]{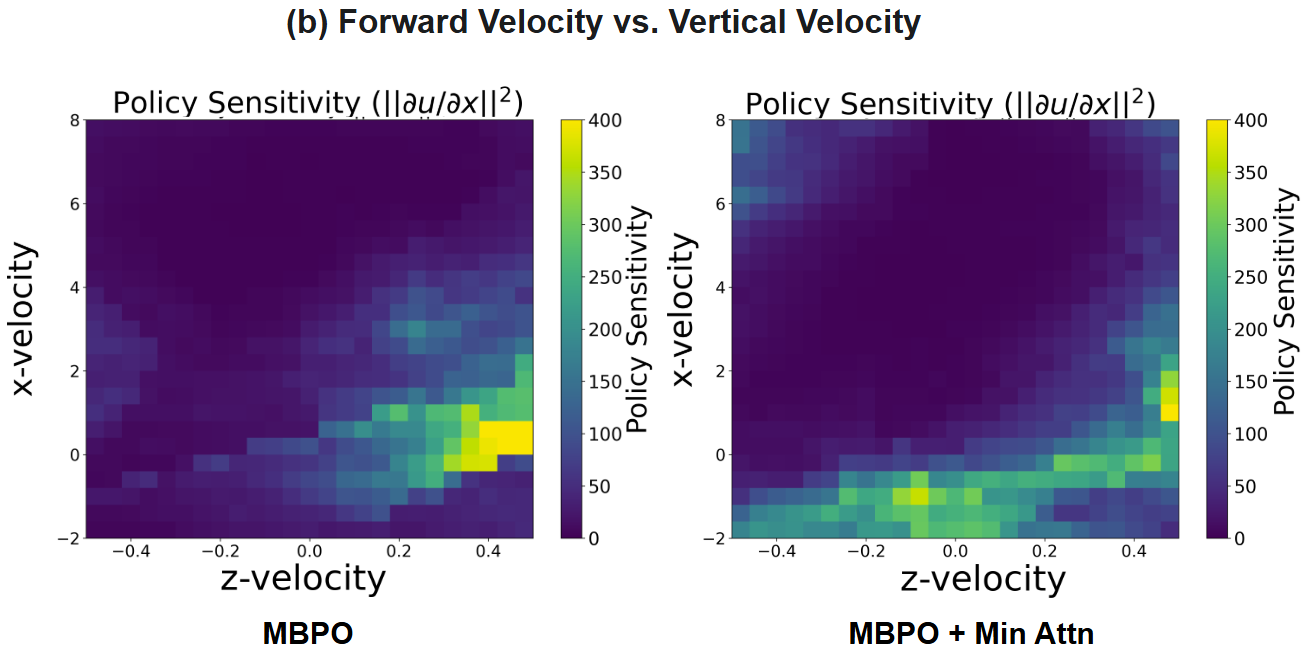} 
\end{subfigure}
\hfill 
\begin{subfigure}[b]{0.49\textwidth}
% \begin{subfigure}[b]{0.3\textwidth}  
    \centering
    \includegraphics[width=\textwidth]{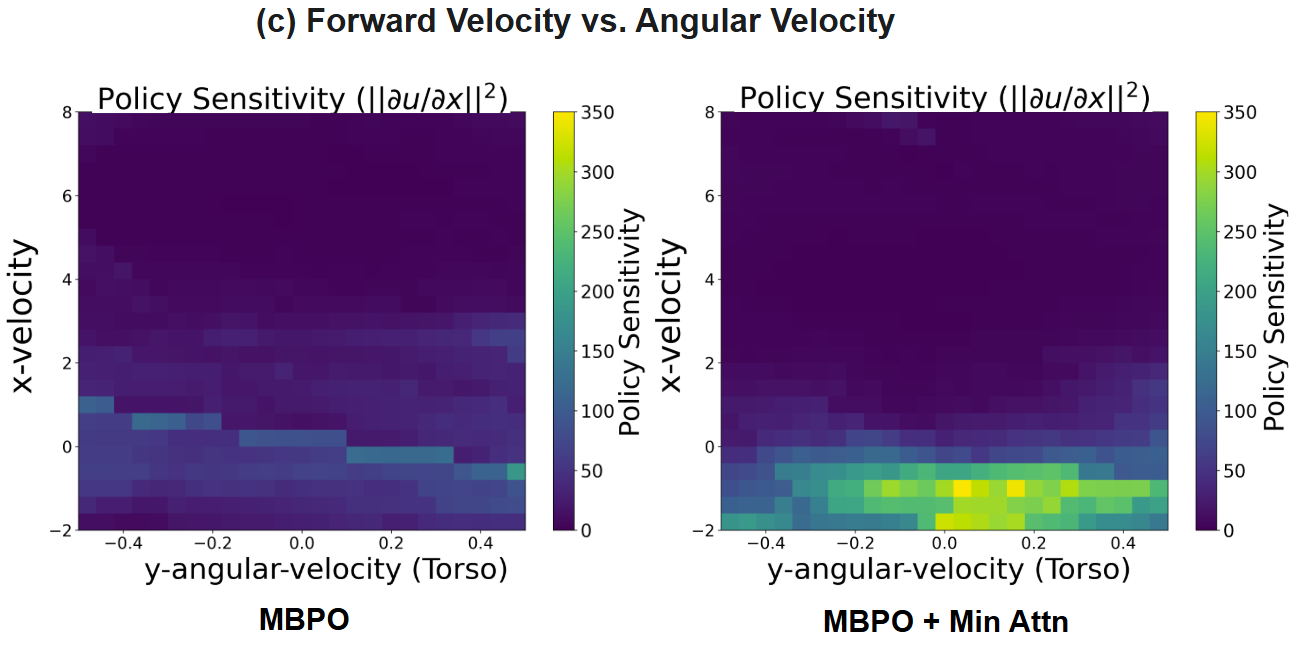}
\end{subfigure}

\caption{Feedback sensitivity heatmaps for HalfCheetah. We visualize $\|\partial u/\partial x\|^2$ over three state-space projections: (a) forward velocity vs. torso height, (b) forward velocity vs. vertical velocity, and (c) forward velocity vs. torso angular velocity. Minimum attention reduces localized sensitivity peaks and yields a more distributed feedback profile in task-relevant regions.}
%Heatmaps of policy sensitivity ($||\partial u / \partial x||^2$) for HalfCheetah, comparing vanilla MB-MPO and our method with minimum attention. \textbf{(a)} Torso height vs. forward velocity: Our method learns a simpler strategy with a stable base posture, reducing sensitivity at high speeds. \textbf{(b)} Forward velocity vs. vertical velocity: Our method distributes sensitivity over a wider range of vertical velocities, avoiding sharp reactive hotspots. \textbf{(c)} Forward velocity vs. angular velocity: Our method induces a proactive stabilization mechanism, showing higher sensitivity to torso roll at low speeds.}
\label{heatmap_feedback}
\end{figure}

\noindent Figure \ref{heatmap_feedback} compares the spatial sensitivity (feedback norm, Jacobian) of the vanilla MB-MPO policy against our regularized method across three state-space projections, revealing a consistent narrative:\\
\textbf{$\bullet$ In locomotion (Figure \ref{heatmap_feedback}a, height vs.\ velocity):} The vanilla policy learns a complex, reactive strategy with high sensitivity at high speeds. Our regularization encourages a simpler, more robust base posture, drastically reducing sensitivity and demonstrating a more efficient control strategy.\\
\textbf{$\bullet$ In vertical motion (Figure \ref{heatmap_feedback}b, velocity vs.\ Z-velocity):} The vanilla policy relies on a concentrated, high-magnitude ``hotspot" of sensitivity to manage the takeoff phase. Our method distributes this control effort over a wider range of vertical velocities, replacing the sharp, reactive correction with a smoother, more continuous control function.\\
\textbf{$\bullet$ In stabilization (Figure \ref{heatmap_feedback}c, velocity vs.\ angular velocity):} A key strategic difference is revealed. While the vanilla policy is largely passive to torso roll at low speeds, our regularized policy exhibits sensitivity in regions associated with torso roll at low speeds, suggesting a more anticipatory corrective pattern in this projection.

\begin{figure}[hbt]
\centering
\begin{subfigure}[b]{0.49\textwidth}
%\begin{subfigure}[b]{0.35\textwidth}  
    \centering
    \includegraphics[width=\textwidth]{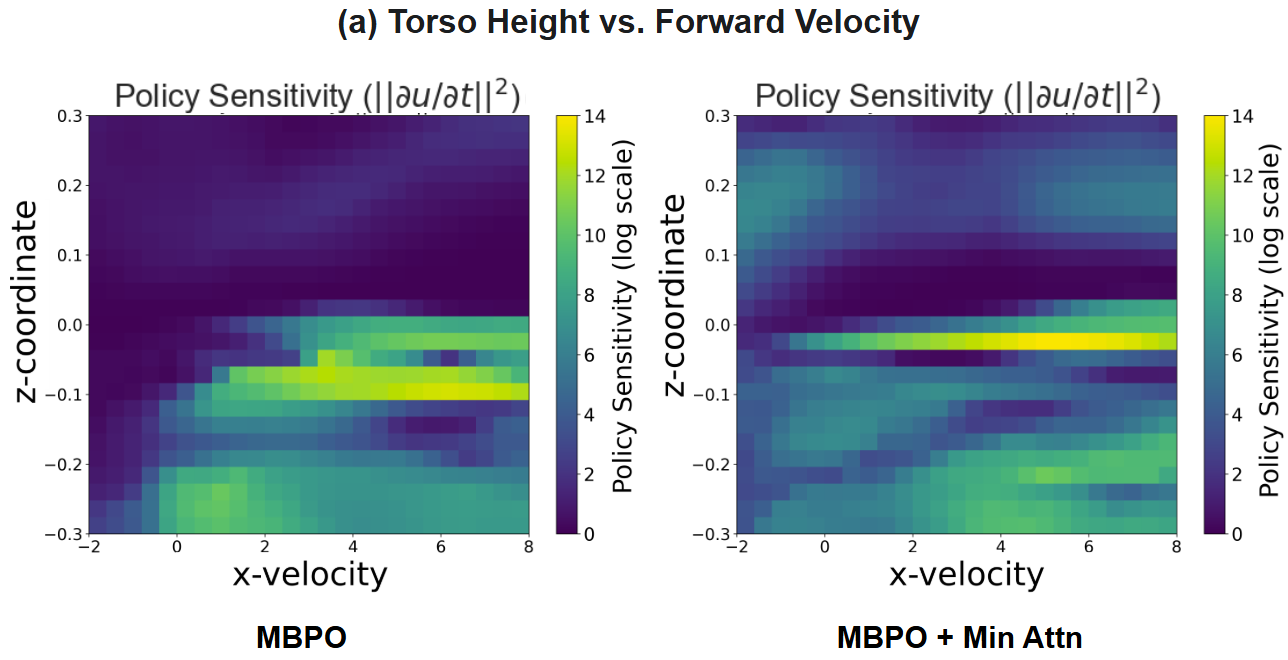}
\end{subfigure}
\hfill 
\begin{subfigure}[b]{0.49\textwidth}
%\begin{subfigure}[b]{0.35\textwidth}  
    \centering
    \includegraphics[width=\textwidth]{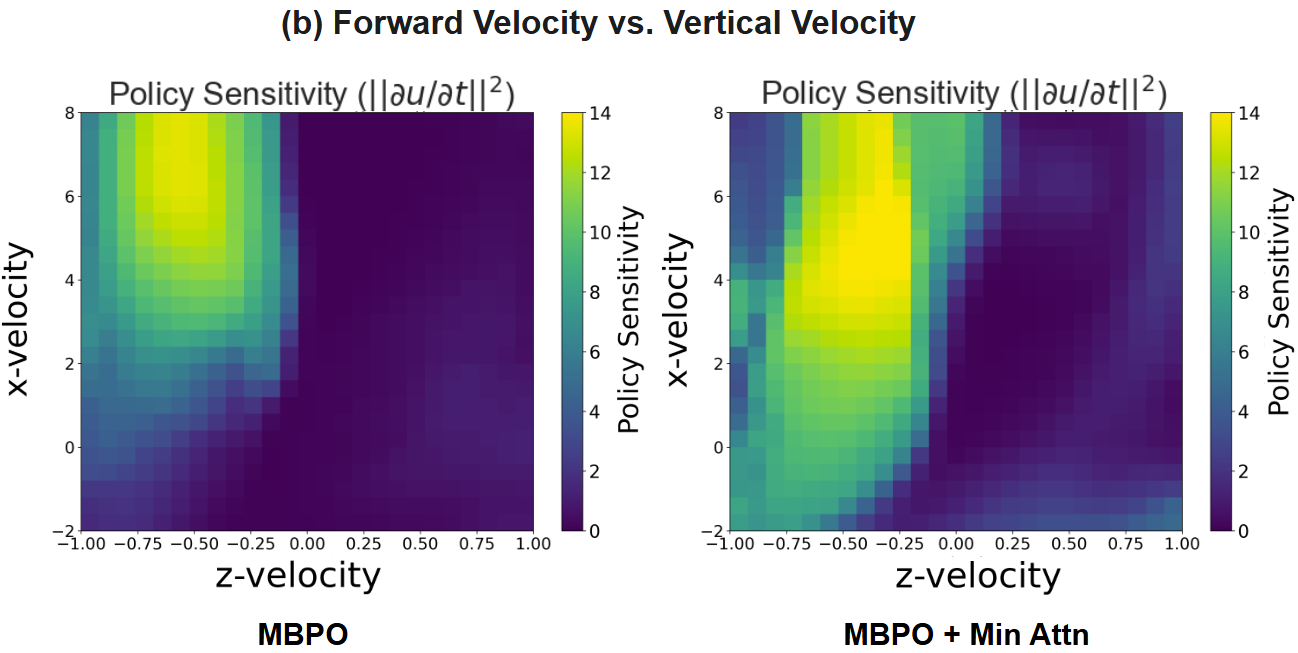} 
\end{subfigure}
\hfill 
\begin{subfigure}[b]{0.49\textwidth}
%\begin{subfigure}[b]{0.35\textwidth}  
    \centering
    \includegraphics[width=\textwidth]{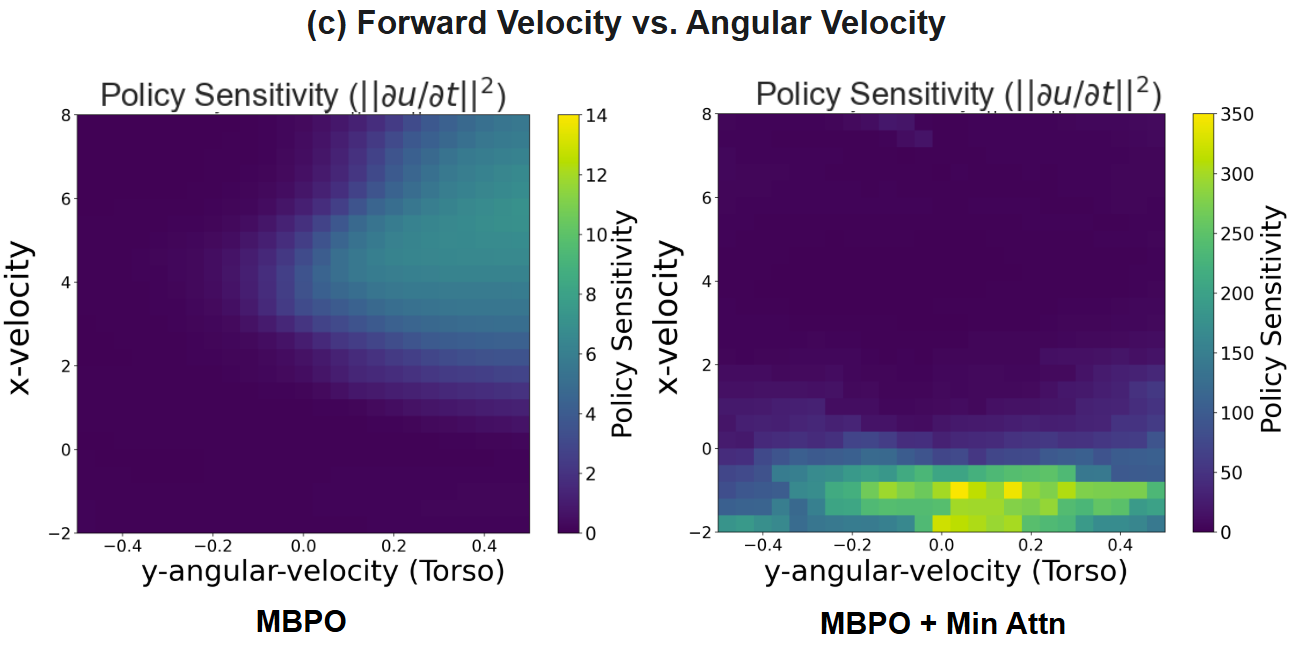}
\end{subfigure}
\caption{Heatmaps of feedforward ($||\partial u / \partial t||^2$) for HalfCheetah. Our method concentrates propulsive jerk and introduces a novel, proactive stabilization behaviour against torso roll at low speeds.}
\label{heatmap_feedforward}
\end{figure}

Figure \ref{heatmap_feedforward} shows the analogous analysis for temporal sensitivity (jerk). The regularized policy concentrates its propulsive jerk into more precise phases of the gait cycle, while introducing proactive, stabilizing jerk to counter torso roll at low speeds---a behavior absent in the baseline. Taken together, these visualizations show a shift from a reactive, sporadic control strategy to a proactive, structured, and efficient one.

Across both spatial and temporal projections, both policies identify a similar region (low Z-coord, high forward velocity) as requiring high sensitivity, likely corresponding to maintaining balance at high speed and near critical phases of locomotion---this agreement suggests these are genuinely critical regions dictated by the task dynamics, with regularization primarily affecting the {\it magnitude} rather than the {\it location} of high-sensitivity behavior. The vanilla policy's high Jacobian norms in these regions suggest an aggressive, reactive strategy, optimal for pure reward maximization but potentially less robust or energy-efficient; the attention-regularized policy remains sensitive where needed but avoids the same extreme magnitude, consistent with smoother transitions and improved energy efficiency.

\begin{figure}[tb]
\centering
% \begin{subfigure}[b]{0.49\textwidth}
\begin{subfigure}[b]{0.32\textwidth} 
   \centering
   \includegraphics[width=\textwidth]{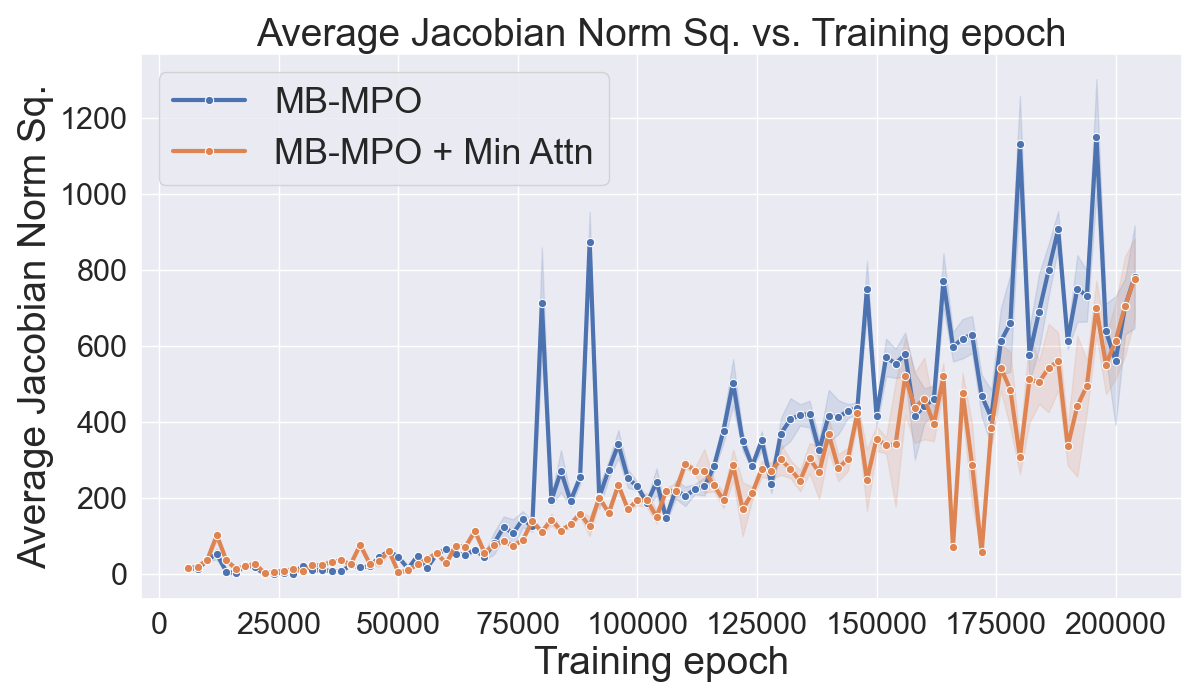}
   \label{fig:figure3a}
\end{subfigure}
\hfill 
% \begin{subfigure}[b]{0.49\textwidth} 
\begin{subfigure}[b]{0.32\textwidth} 
   \centering
   \includegraphics[width=\textwidth]{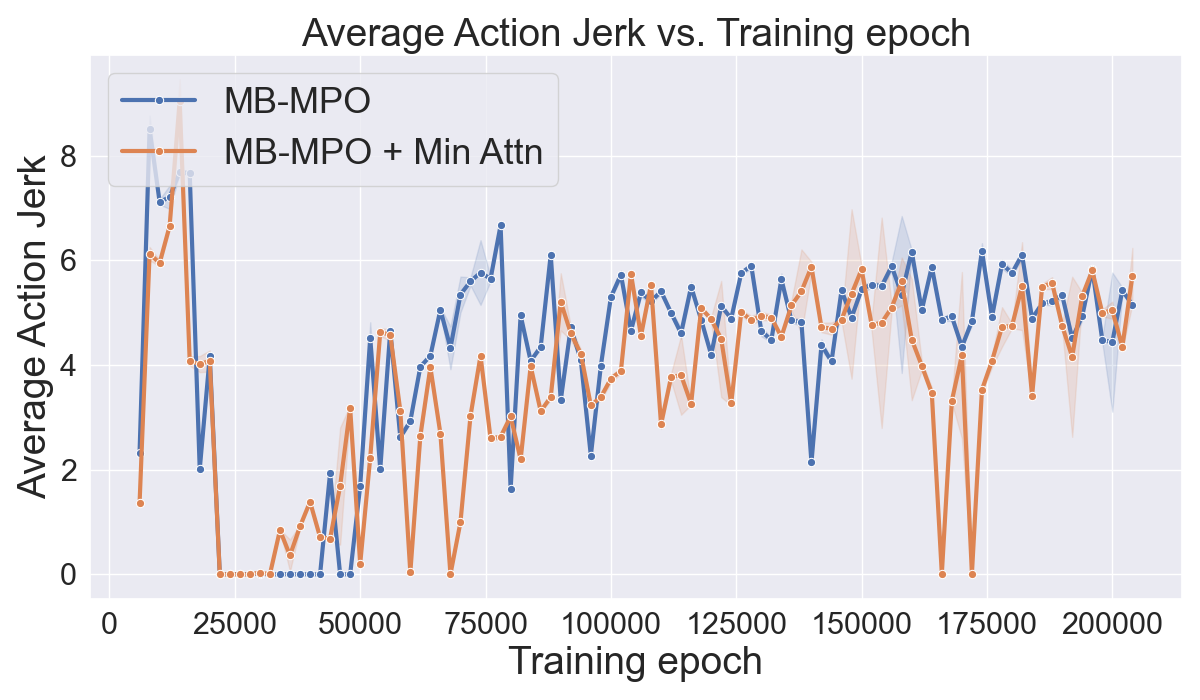} 
   \label{fig:figure3b}
\end{subfigure}
\hfill 
% \begin{subfigure}[b]{0.49\textwidth}
\begin{subfigure}[b]{0.32\textwidth} 
   \centering
   \includegraphics[width=\textwidth]{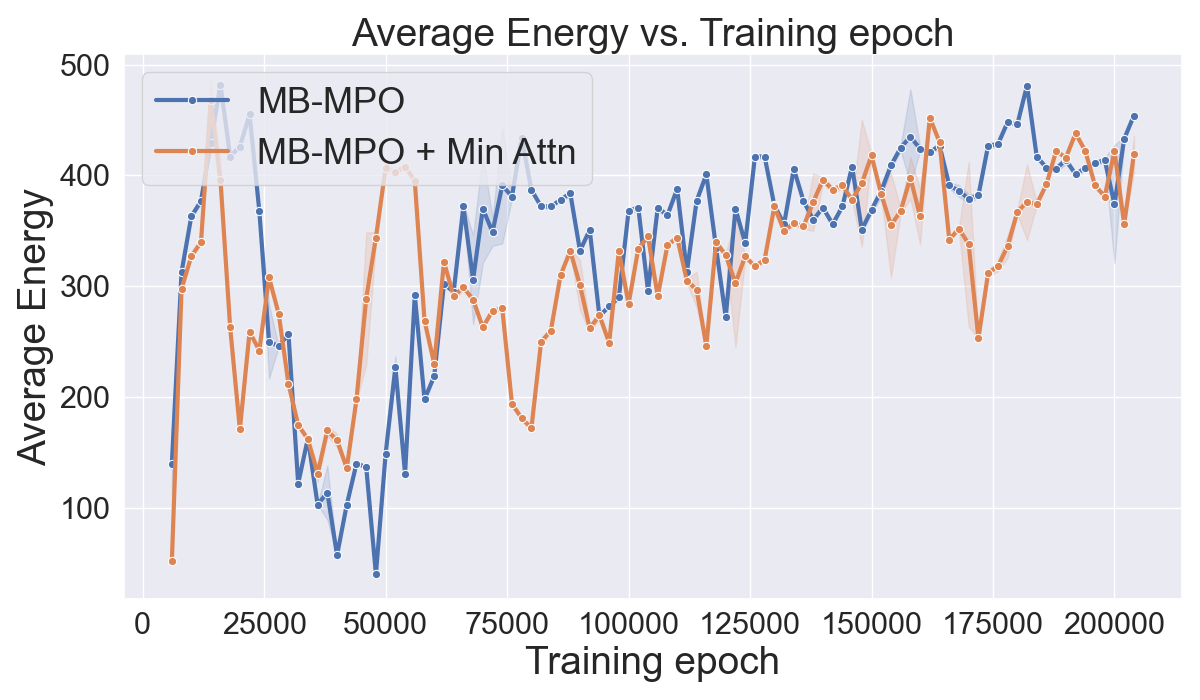} 
   \label{fig:figure3c}
\end{subfigure}

\caption{Half-Cheetah: Time profiles for the feedback, feedforward, and energy in training. The comparison of feedback ($||\partial u / \partial x||^2$, Jacobian), feedforward ($||\partial u/ \partial t||^2$, Jerk), and energy ($||u||^2$) between MB-MPO and MB-MPO with minimum attention ($\alpha = 1.0$).}
\label{time_profile_half_cheetah}
\end{figure}

\noindent This same structuring effect is visible not only in the final trained policy but throughout training. Figure \ref{time_profile_half_cheetah} shows time profiles for the feedback, feedforward, and energy terms over the course of training. The benefits of minimum attention regularization are present from the early stages of learning: the average feedback norm for our method remains consistently below the baseline after an initial exploration phase, and energy consumption is also markedly reduced. This confirms that minimum attention actively shapes the policy towards smoother and more efficient solutions throughout the entire learning process, not only at convergence. As with the sensitivity heatmaps above, we do not expect MA to minimize every auxiliary metric independently: the regularizer changes the trade-off between reward, feedback sensitivity, temporal variation, and control energy, and in some cases one auxiliary quantity may increase while reward or energy improves, indicating a different allocation of feedback and feedforward effort rather than uniform suppression of all control variation.

\subsection{Adaptation to Out-of-Distribution Perturbations}
We evaluate the meta-learning functionality of minimum attention under out-of-distribution (OOD) perturbations, first establishing performance on standard locomotion tasks and then testing adaptation under three perturbation types. For the Half-Cheetah agent, model perturbation is realized by a crippled leg and by mass gain or loss; environmental perturbation is posed by variation in floor angle, emulating running uphill and downhill.

\begin{figure}[tb]
\centering
% \begin{subfigure}[b]{0.49\textwidth}
\begin{subfigure}[b]{0.3\textwidth}
%\begin{subfigure}[b]{0.28\textwidth} 
    \centering
    \includegraphics[width=\textwidth]{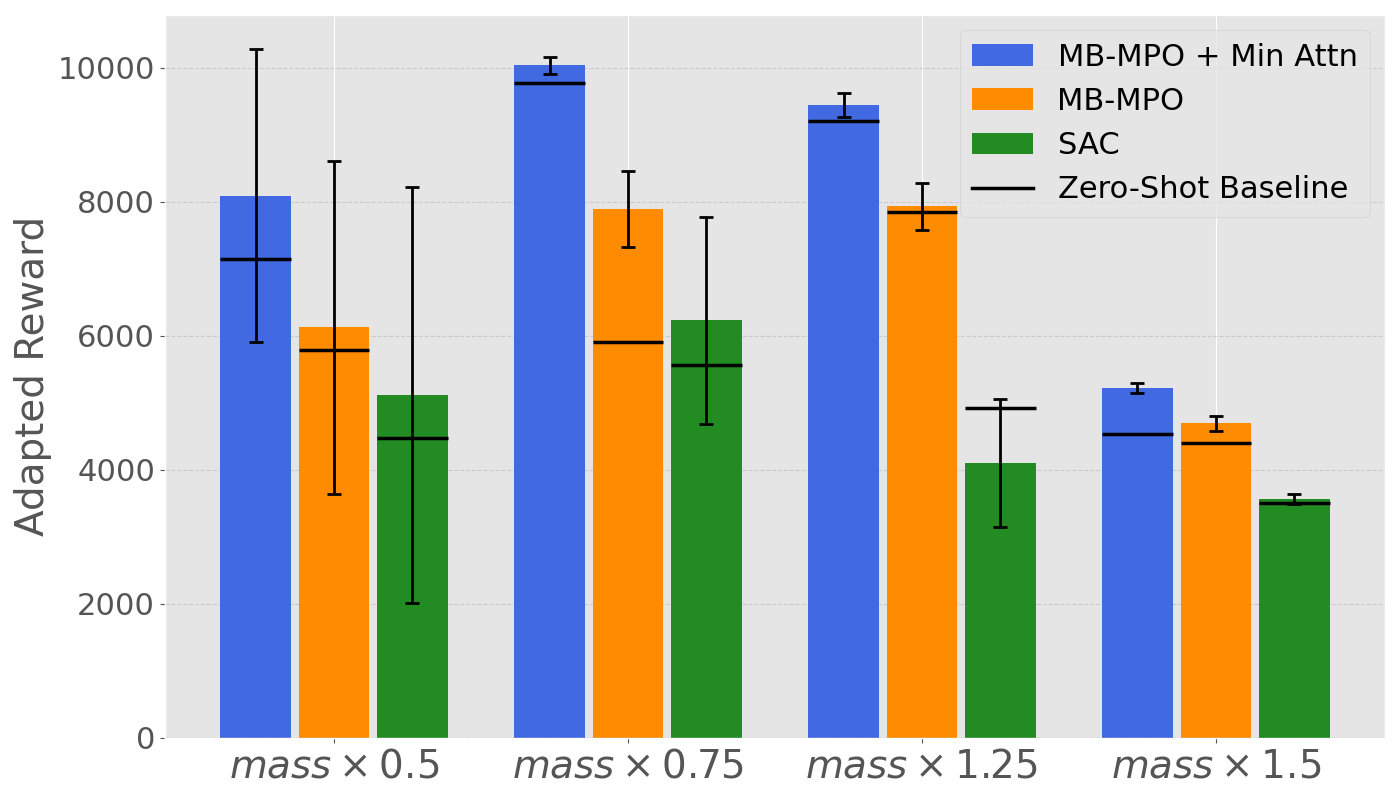}
    \subcaption{Body weight}
    \label{fig:figure5a}
\end{subfigure}
\hfill 
% \begin{subfigure}[b]{0.49\textwidth}
\begin{subfigure}[b]{0.3\textwidth}
%\begin{subfigure}[b]{0.28\textwidth} 
    \centering
    \includegraphics[width=\textwidth]{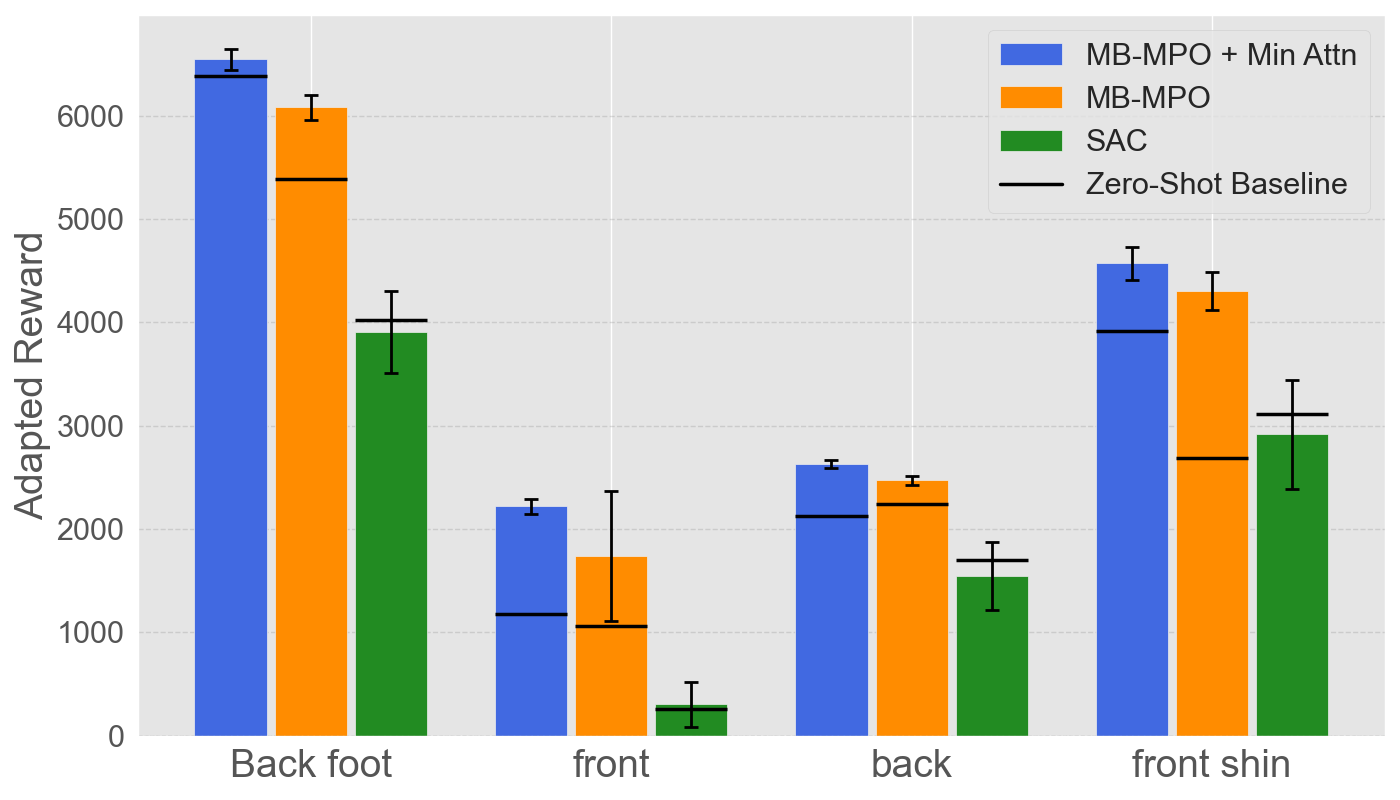} 
    \subcaption{Crippled leg}
    \label{fig:figure5b}
\end{subfigure}
\hfill 
% \begin{subfigure}[b]{0.49\textwidth}
\begin{subfigure}[b]{0.3\textwidth}
%\begin{subfigure}[b]{0.28\textwidth}
    \centering
    \includegraphics[width=\textwidth]{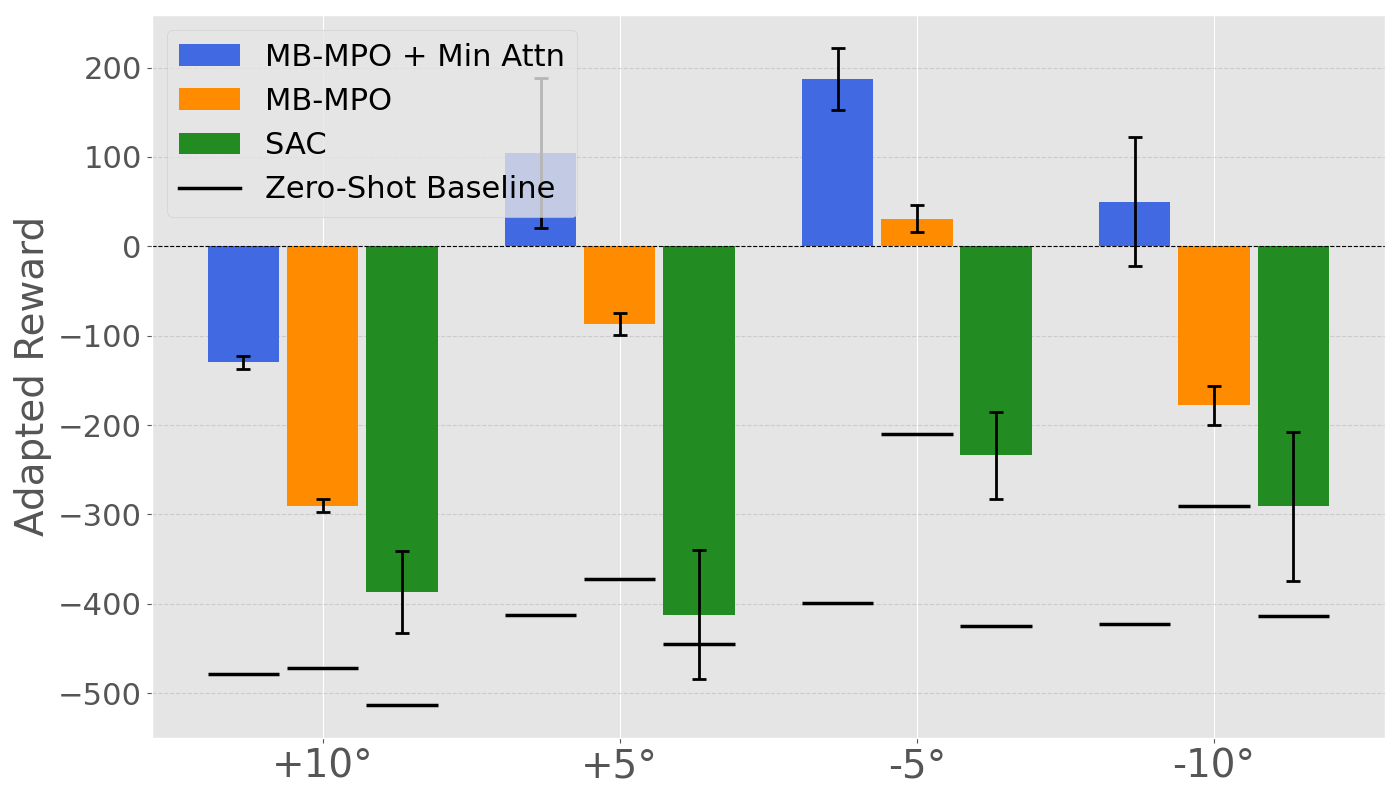}
    \subcaption{Environment transition}
    \label{fig:figure5c}
\end{subfigure}
\caption{Half-Cheetah: Meta-testing under OOD perturbations: (a) body mass change (Easy OOD tasks), (b) crippled legs (Medium OOD tasks), (c) uphill/downhill terrain (Difficult OOD tasks). We compare SAC, MB-MPO, and MB-MPO + minimum attention (MA, $\alpha=1.0$) over 10 evaluation steps. MA improves adaptation performance and reduces variance across perturbation settings.}
\label{meta-testing}
\end{figure}

\noindent In the meta-testing scenario where the agent must adapt to a crippled leg, the performance benefits of minimum attention are maintained or even amplified: Half-Cheetah with minimum attention achieves a higher reward ($6822 \pm 109$) than the vanilla baseline ($6356 \pm 132$) and does so with a lower feedback norm and feedforward norm (jerk). This suggests that the smoother policies learned through minimum attention provide a better prior for rapid adaptation to significant out-of-distribution embodiment changes. Figure \ref{meta-testing} shows the competence of minimum attention across all three perturbation types, with outperforming total rewards and reduced variance relative to both SAC and vanilla MB-MPO.

\subsection{Ablations}
\label{sec:ablations}

We conduct two ablation studies: one isolating the necessity of the two components of the minimum attention regularizer, and one characterizing sensitivity to the ensemble size $M$ and imaginary rollout sample size $N$.

\noindent \textbf{Necessity of spatial and temporal regularization.} To distinguish minimum attention from simpler smoothness regularizers and verify the necessity of the Brockett formula's components, we compare the full formulation against a Jacobian-only baseline (a state-smoothness regularizer) and a temporal-only baseline (regularizing control variation over time). As shown in Table \ref{table:ablation_synergy}, the full formulation (Ours, MA) significantly outperforms both single-component baselines, confirming the importance of regularizing changes in both state and time jointly rather than either alone.

\begin{table}[hbt]
\caption{Ablation study on HalfCheetah (200K iterations, $\alpha = 1.0$). We compare the full minimum attention (MA) regularizer against spatial-only (Jacobian) and temporal-only components. Bold indicates the best value per row.}
\label{table:ablation_synergy}
\begin{center}
  \small
  \begin{tabular}{|l|c|c|c|}
    \hline
    \textbf{Metric} & \textbf{Ours (Full MA)} & \textbf{Jacobian-only} & \textbf{Temporal-only} \\
    \hline
    Reward $\uparrow$ & \textbf{9721 $\pm$ 128} & 6484 $\pm$ 148 & 6475 $\pm$ 135 \\
    \hline
    Feedback $\downarrow$ & 567 $\pm$ 53 & \textbf{128 $\pm$ 24.5} & 736.2 $\pm$ 84.9 \\
    \hline
    Feedforward $\downarrow$ & 5.01 $\pm$ 0.07 & 9.54 $\pm$ 0.35 & \textbf{4.94 $\pm$ 0.04} \\
    \hline
    Energy $\downarrow$ & 3.85 $\pm$ 0.05 & \textbf{3.00 $\pm$ 0.03} & 3.09 $\pm$ 0.03 \\
    \hline
  \end{tabular}
\end{center}
\end{table}

\begin{figure}[b]
\centering
\begin{subfigure}[b]{0.48\textwidth}
%\begin{subfigure}[b]{0.35\textwidth}
%\begin{subfigure}{0.28\textwidth}  
    \centering
    \includegraphics[width=\textwidth]{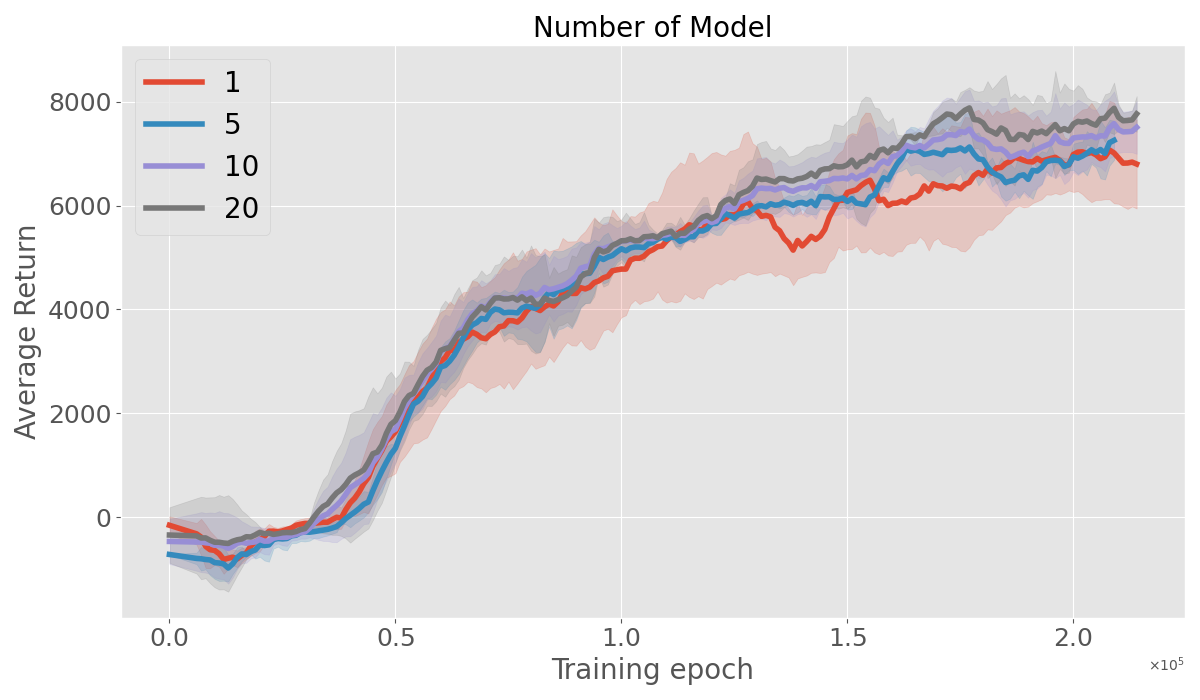}
    \label{fig:figure4a}
\end{subfigure}
\hfill 
\begin{subfigure}[b]{0.48\textwidth}
%\begin{subfigure}[b]{0.35\textwidth}
%\begin{subfigure}{0.28\textwidth}   
    \centering
    \includegraphics[width=\textwidth]{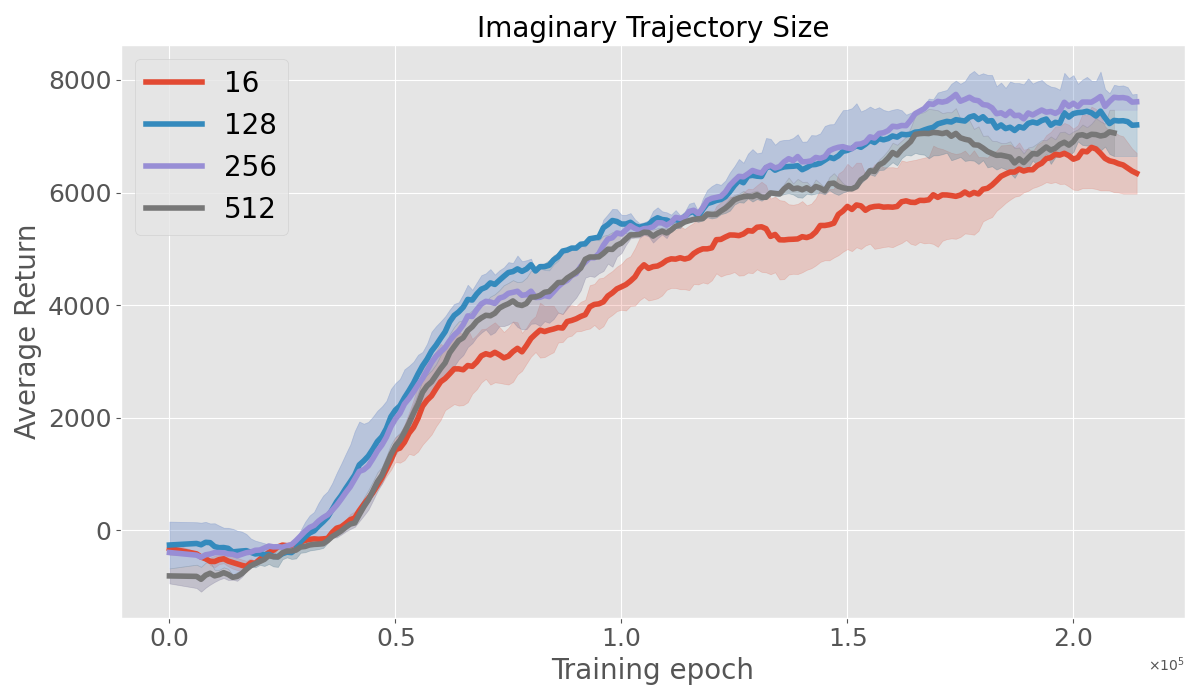} 
    \label{fig:figure4b}
\end{subfigure}
\caption{Half-Cheetah: Dependency on the ensemble model size and imaginary rollouts. When the ensemble model size of $M$ increases to 20,  the total reward also increases. The imaginary sample size $N$ provides the benefit of increased total reward to $N=128$, but decreases when $N$ is turned to 256.}
\label{model_ensemble_imaginary}
\end{figure}

\noindent \textbf{Ensemble size and imaginary rollout length.} Figure \ref{model_ensemble_imaginary} shows the influence of the ensemble model size $M$ and the imaginary sample size $N$ on HalfCheetah. Larger ensembles consistently improve reward, with total reward increasing as $M$ grows to 20. The imaginary sample size shows a non-monotonic effect: total reward increases up to $N=128$, then decreases as $N$ is further increased to 256.

\subsection{Maturation of Learned Control Strategies}

\begin{figure}[hbt]
\begin{center}
\includegraphics[angle=0, width=1.0\textwidth]{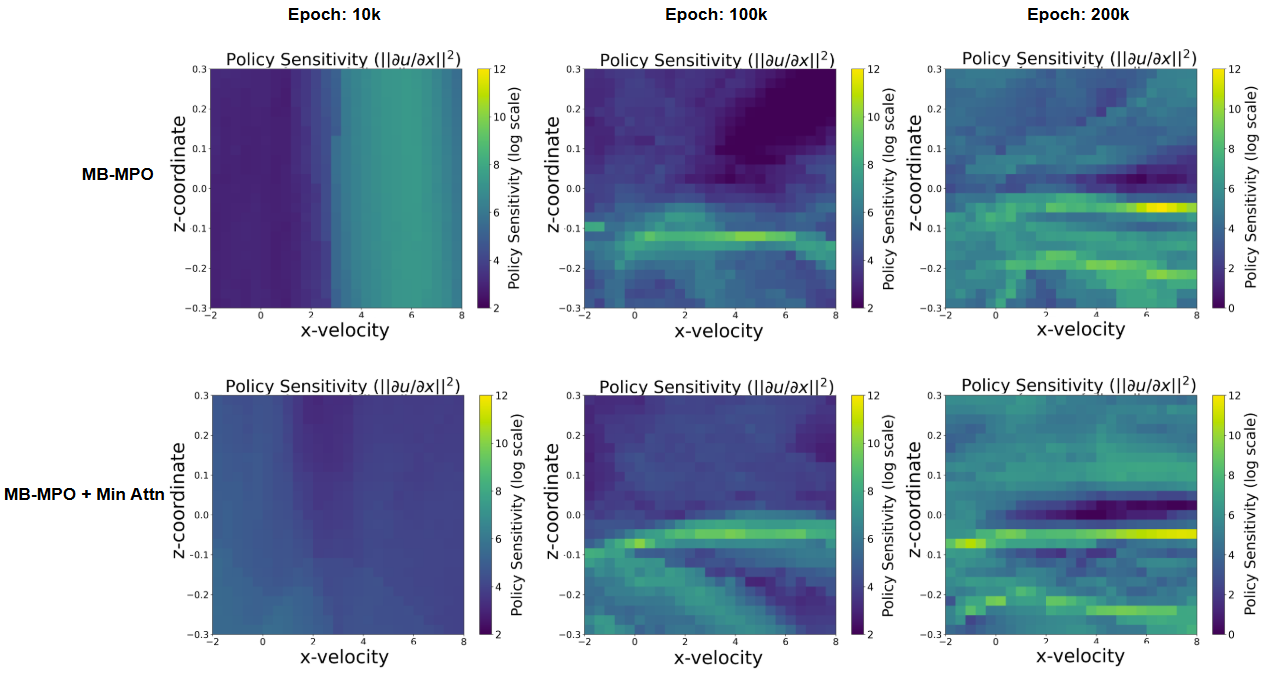}
\end{center}
\caption{Half Cheetah: Evolution of spatial sensitivity (feedback norm, $||\partial u/\partial x||^2$). \textbf{Top Row (MB-MPO):} Converges to a broadly reactive strategy. \textbf{Bottom Row (Ours):} Converges to a sparse, structured strategy, focusing sensitivity on a critical phase.}
\label{heatmap_feedback_evolution}
\end{figure}

\begin{figure}[hbt]
\begin{center}
\includegraphics[angle=0, width=1.0\textwidth]{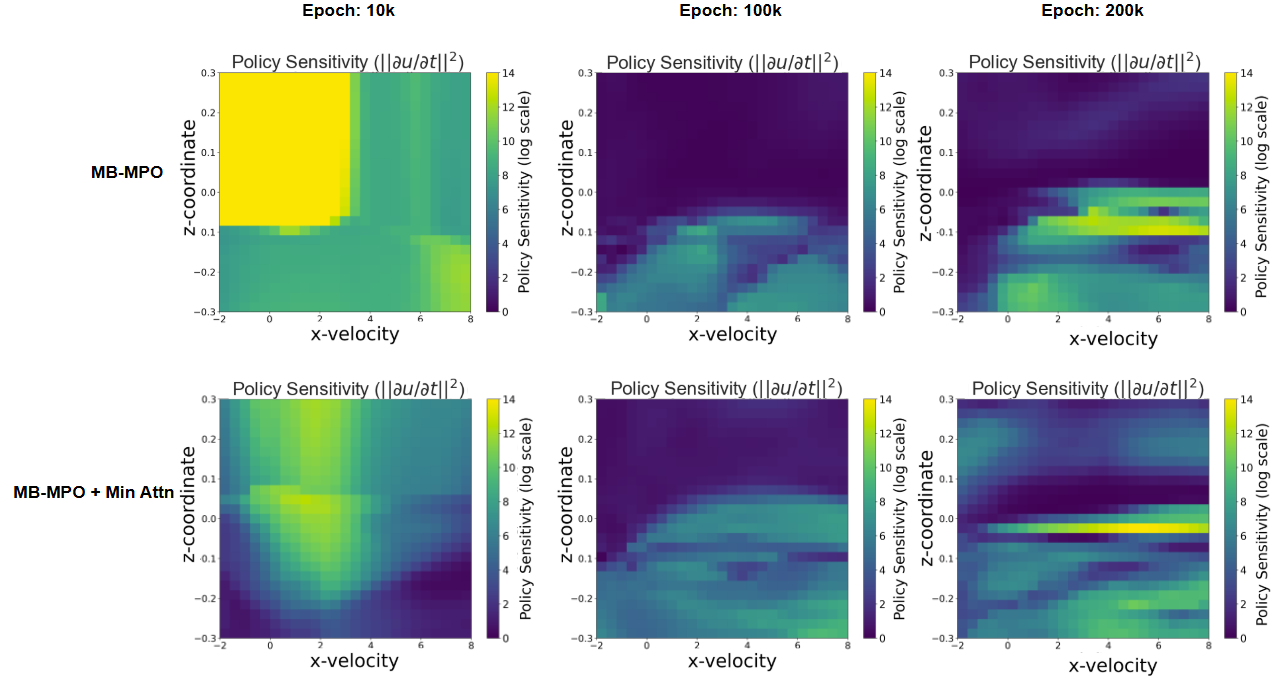}
\end{center}
\caption{Half Cheetah: Evolution of temporal sensitivity (feedforward norm/jerk, $||\partial u/\partial t||^2$). \textbf{Top Row (MB-MPO):} Progresses from chaotic flailing to a moderately jerky running gait. \textbf{Bottom Row (Ours):} The regularizer shapes the policy to be smoother overall, concentrating its action changes into a single, precise phase of the gait cycle.}
\label{heatmap_feedforward_evolution}
\end{figure}

To understand how the final control strategies emerge, we visualize the evolution of both spatial sensitivity (feedback/Jacobian norm) and temporal sensitivity (feedforward/jerk) at different stages of training. Figures \ref{heatmap_feedback_evolution} and \ref{heatmap_feedforward_evolution} show the policy heatmaps for HalfCheetah at 10K, 100K, and 200K epochs for both vanilla MB-MPO and our regularized method.

\noindent \textbf{Maturation of feedback (Jacobian) strategy.} At the early stage of training (10K epochs), both policies exhibit broad, unstructured sensitivity. As training progresses, a clear divergence in strategy appears. The vanilla MB-MPO policy develops a wide, high-sensitivity band at high forward velocities, indicating it has learned a reactive strategy that requires constant, high-gain corrections to maintain a running gait. In contrast, our regularized policy converges to a much sparser and more structured landscape, becoming largely insensitive across most of the state space and focusing its sensitivity into a narrow, precise band corresponding to the apex of its running motion.

\noindent \textbf{Maturation of feedforward (jerk) strategy.} The vanilla MB-MPO policy begins with extremely high, undifferentiated jerk at low speeds (10K epochs), suppresses this initial jerk as it learns, but retains a moderately high level of action change during the low-crouch running phase (200K epochs). The minimum attention policy follows a different developmental path: while it also explores with high jerk initially, minimum attention quickly shapes the policy to be temporally smoother overall, converging by 200K epochs to a highly specialized strategy where jerk is concentrated into a very sharp, narrow band at a specific torso height---consistent with an efficient, ballistic motion where the agent ``coasts" with smooth actions for most of the gait cycle.

\noindent Tables \ref{half_cheetah:meta_training} and \ref{half_cheetah:meta_testing} quantify this maturation across early, middle, and final training stages. Corresponding maturation analysis for Hopper, Walker2D, and Humanoid is provided in Appendix (\ref{table:hopper:meta-training}, \ref{table:hopper:meta-testing}, \ref{table:Walker2D:meta-training}, \ref{table:Walker2D:meta-testing}, \ref{humanoid:meta_training}, \ref{humanoid:meta_testing}).

\begin{table}[hbt]
\caption{Half-Cheetah: Meta-training on the maturation of meta-learning} \label{half_cheetah:meta_training}
\begin{center}
  \small
  \begin{tabular}{|l|c|c|c|c|}
    \hline
           & MB-MPO & MA ($\alpha = 1.0$) & MA ($\alpha = 0.01$) & MA ($\alpha = 0.05$)    \\
    \hline
    Early stage of epoch: 10K & & & &\\
    \hline    
      Average total rewards   &  -1084  $\pm$ 73 & \textbf{672 $\pm$ 67} & -427 $\pm$ 68 & -579 $\pm$ 170  \\
     \hline
     Average feedback normalized  & 38 $\pm$ 3.95 & \textbf{13 $\pm$ 0.35} & 36 $\pm$ 5.91 & 26 $\pm$ 1.91  \\
    \hline
     Average feedforward normalized  & 7.11 $\pm$ 0.06 & 7.33 $\pm$ 0.04 & \textbf{5.99 $\pm$ 0.05} & 6.86 $\pm$ 0.06     \\
    \hline
%     Average energy normalized  & 363 $\pm$ 1.68 & \textbf{320 $\pm$ 1.55} & 327 $\pm$ 1.51 & 350 $\pm$ 0.96     \\
        Average energy normalized  & 3.63 $\pm$ 0.017 & \textbf{3.20 $\pm$ 0.016} & 3.27 $\pm$ 0.15 & 3.50 $\pm$ 0.010     \\
    \hline
    Middle stage of epoch: 100K & &.  &\\
    \hline
      Average total rewards   &  5531  $\pm$ 127 & \textbf{6564 $\pm$ 57} & 4242 $\pm$ 68 & 5870 $\pm$ 117  \\
     \hline
     Average feedback normalized  & 235 $\pm$ 25.1 & 212 $\pm$ 17.7 & \textbf{191 $\pm$ 13.9} & 232 $\pm$ 25.4  \\
    \hline
     Average feedforward normalized  & 5.28 $\pm$ 0.03 & 5.01 $\pm$ 0.07 & \textbf{3.78 $\pm$ 0.11} & 5.69 $\pm$ 0.08     \\
    \hline
%     Average energy normalized  & 367 $\pm$ 2.16 & 344 $\pm$ 2.15 & \textbf{284 $\pm$ 1.56} & 356 $\pm$ 2.20     \\
       Average energy normalized  & 3.67 $\pm$ 0.022 & 3.44 $\pm$ 0.022 & \textbf{2.84 $\pm$ 0.016} & 3.56 $\pm$ 0.022     \\
    \hline        
  \end{tabular}
\end{center}
\end{table}

\begin{table}[hbt]
\caption{Half-Cheetah: Meta-testing on the maturation of meta-learning} \label{half_cheetah:meta_testing}
\begin{center}
  \small
  \begin{tabular}{|l|c|c|c|c|}
    \hline
           & MB-MPO & Ours ($\alpha = 1.0$) & Ours ($\alpha = 0.01$) & Ours ($\alpha = 0.05$)    \\
    \hline
    Early stage of epoch: 10K & & & &\\
    \hline    
      Average total rewards   &  -1245  $\pm$ 801 & 99 $\pm$ 180 & 90 $\pm$ 102 & \textbf{100 $\pm$ 139}  \\
     \hline
     Average feedback normalized  & \textbf{13.6 $\pm$ 9.23} & 21.5 $\pm$ 1.36 & 19.2 $\pm$ 6.8 & 15.3 $\pm$ 1.65  \\
    \hline
     Average feedforward normalized  & 10.5 $\pm$ 0.07 & 8.12 $\pm$ 0.15 & \textbf{5.99 $\pm$ 0.14} & 8.07 $\pm$ 0.08     \\
    \hline
%     Average energy normalized  & 425 $\pm$ 5.68 & 330 $\pm$ 4.54 & \textbf{320 $\pm$ 2.05} & 350 $\pm$ 3.15     \\
        Average energy normalized  & 4.25 $\pm$ 0.057 & 3.30 $\pm$ 0.045 & \textbf{3.20 $\pm$ 0.021} & 3.50 $\pm$ 0.032     \\
    \hline
    Middle stage of epoch: 100K & & & &\\
    \hline    
      Average total rewards   & 4907   $\pm$ 69 & \textbf{4953 $\pm$ 85} & 4920 $\pm$ 50 & 4950 $\pm$ 138  \\
     \hline
     Average feedback normalized  & \textbf{164 $\pm$ 11.5} & 301 $\pm$ 12.7 & 222 $\pm$ 10.5 & 185 $\pm$ 10.1  \\
    \hline
     Average feedforward normalized  & 6.15 $\pm$ 0.03 & 5.45 $\pm$ 0.05 & \textbf{3.68 $\pm$ 0.03} & 6.1 $\pm$ 0.04     \\
    \hline
%     Average energy normalized  & 325 $\pm$ 1.54 & 302 $\pm$ 1.42 & \textbf{230 $\pm$ 1.72} & 340 $\pm$ 2.9     \\
        Average energy normalized  & 3.25 $\pm$ 0.015 & 3.02 $\pm$ 0.014 & \textbf{2.30 $\pm$ 0.017} & 3.40 $\pm$ 0.029     \\
    \hline        
  \end{tabular}
\end{center}
\end{table}

\section{Conclusion}
The formalism of minimum attention is shown to have a significant signature of dexterity to mitigate the variance in training learning curves and meta-testing. This is an important feature for designing lower-variance and more energy-efficient learned controllers. Also, the regularization of minimum attention, even though with an implication of conservative minimal changes in state and time, demonstrates few sampling in out-of-distribution meta-testing and domain adaptation. 
Our results are empirical and primarily evaluated on MuJoCo locomotion tasks with dense rewards and continuous control. Although these environments are useful for analyzing feedback, feedforward variation, energy, and perturbation adaptation, they do not establish formal safety guarantees or transfer to all robotic settings. Lower variance and smoother control profiles should therefore be interpreted as evidence of improved learning stability, not as proof of safe behavior. In addition, the connection between minimum attention and few-shot adaptation is empirical in this work; theoretical analysis for the boundedness and stability of minimum attention based on control Lyapunov functions remains as future work, where Thompson sampling can be one reference. Finally, the regularization weight $\alpha$ is tuned per environment family, and tasks with changing reward functions are not studied here. The regularization of minimum attention can be further explored in its application in algorithmic debiasing. It is intriguing whether singularities and rare events can be degenerate in the open avenues of language models from minimum attention. The motor learning and control from agents of Half-Cheetah, Hopper, Walker2D, and Humanoid can be further extended to other models. It needs more exploration to characterize how this formalism works in model-free RLs and to overcome the limitations of gradient-based meta-learning via multi-task learning. 

\section*{Acknowledgements}
PL gives thanks to Frank C. Park introduding the formalism of minimum attention by Roger Brockett. SG and PL were partially supported by NIH Data Science and NIMHD grant, 3U54MD013376-04S3. PL was partially supported by Center for Equitable AI and Machine Learning Systems (CEAMLS) Affilated Project (ID: 10072401), and NIH OT2 Award, 1OT2OD042825-01.
\bibliographystyle{plainnat}
\bibliography{main}

\clearpage
\appendix

\section{Technical Appendices and Supplementary Material}

\setcounter{page}{1}
\setcounter{equation}{0}
\setcounter{algorithm}{0}
\setcounter{figure}{0}
\setcounter{table}{0}

\renewcommand{\theequation}{S\arabic{equation}}
\renewcommand{\thealgorithm}{S\arabic{algorithm}}
\renewcommand{\thetable}{S\arabic{table}}
\renewcommand{\thefigure}{S\arabic{figure}}

% \subsection{Discrete-time Approximation of Control Derivatives}
% While the minimum attention criterion in Eq. (1) is defined in continuous time and state, policy optimization in discrete MDPs requires a consistent approximation. For a policy $u_{\theta}$ at state $x_{t}$, we compute the spatial and temporal derivatives as follows:

% \noindent \textbf{Spatial Derivative: The Jacobian is computed analytically via auto-differentiation through the policy network:
% \begin{equation}
% \| \frac{\partial u_\theta}{\partial x} \|^2 \approx \left| \nabla_x u_\theta(x_t) \right|^2_F
% \end{equation}
% \textbf{Temporal Derivative:} Since $\partial u_{\theta} / \partial t $
%  represents the change in policy over learning time for a fixed state, we use a discrete-time difference between the current policy parameters $\theta_k$ and the previous update $\theta_{k-1}$
% \begin{equation}
% \| \frac{\partial u_\theta}{\partial t} \|^2 \approx \left| \frac{u(x_t; \theta_k) - u(x_t; \theta_{k-1})}{\Delta t} \right|^2
% \end{equation}
% where $\Delta t$ is the number of policy updates between visits to state $x_t$
% This formulation ensures that the regularization effectively penalizes erratic changes in both the control landscape and the learning trajectory.
% }

%\begin{enumerate}
%
\subsection{Setup / General Configuration}
\begin{itemize}
\item Environments used
\begin{itemize}
\item environments: "Half-Cheetah-v3", "Hopper-v3", “Walker2d-v3”, "Humanoid-v2" from OpenAI Gym.
\item We normalized the env if the actions are not already in [-1,1].
\item We have also normalized the rewards during training.
\end{itemize}
\item Software and Hardware
\begin{itemize}
\item All experiments are done in PyTorch and use OpenAI Gym, MuJoCo.
\item Hardware used for training is "NVIDIA A30"
\end{itemize}
\item Baselines
\begin{itemize}
\item Baseline algorithms "standard MB-MPO”, and “SAC”.
\item For the baseline we used the open-source implementation with minute changes to satisfy our requirements. We used unstable baseline repository \textbf{(link)}
\end{itemize}
\end{itemize}

\subsection{Algorithm Implementation Details (MB-MPO + Minimum Attention)}
\begin{itemize}
\item Core MB-MPO Components
\begin{itemize}
\item For inner and outer loop policy updates we have experimented with SAC. %PPO, TRPO and SAC.
\item SAC is practically more suitable for the updates.
\item We have experimented with 4 different imaginary trajectory size: [16, 128, 256, 512].
\end{itemize}
\item Minimum Attention Regularization Specification
\begin{itemize}
\item The regularization term is $|| \partial u_{\theta} / \partial x ||^2 + || \partial u_{\theta} / \partial t ||^2$
\item The regularization term is added to the reward with weight $``\alpha"$.
\item We have experimented for $``\alpha"$ taking values $[0.01, 0.05, 1.0, 5.0]$ and reported the results.
\end{itemize}
\item Meta-Learning Setup
\begin{itemize}
\item``Meta-learning” loop is that of Model-Agnostic Meta-Learning (MAML) structure (inner adaptation on models, outer update of meta-parameters).
\item Tasks (models) are sampled randomly from the task list, the mini-batch of $R_i$ in Algorithm 1.
\item Specifications of the inner loop and outer loop is in parameter file.
\end{itemize}
\item Model Ensemble Setup
\begin{itemize}
\item We have experimented with four different models of ensemble size: $[1, 5, 15, 20]$ and reported the results.
\end{itemize}
\end{itemize}

\subsection{Neural Network Architectures}

The following architecture is for Half-Cheetah-v3 training.
\begin{itemize}
\item Policy network (actor): The policy network is a multi-layer perceptron (MLP) that maps states to the parameters of a Gaussian action distribution.
\begin{itemize}
\item Input layer: matches the observation dimension of the specific environment (e.g., 17 for Half-Cheetah-v3).
\item Hidden layers: two hidden layers, each with 256 units.
\item Hidden activation: ReLU is used as the activation function after each hidden layer.
\item Output layer (mean): produces the mean of the Gaussian distribution, with dimensionality matching the action space of the environment. This layer uses an identity activation. The output is subsequently passed through a tanh function and scaled/biased to the environment's action range.
\item Output layer (log standard deviation): the log standard deviation (log std) for the Gaussian distribution is produced by a separate output head. The log std values are clamped (e.g., between -20 and 2, a common SAC practice.
\item Reparameterization trick: used (reparameterize: true) for backpropagation through the sampling process.
\item Log probability stabilization: employed (stablize log prob: true) using the standard SAC transformation for tanh-squashed Gaussian distributions.
\end{itemize}
\item Q-networks (critics): two separate Q-networks are used (typical for SAC to mitigate overestimation bias), each an MLP that maps state-action pairs to a Q-value.
\begin{itemize}
\item Input layer: matches the sum of observation and action dimensions.
\item Hidden layers: two hidden layers, each with 256 units.
\item Hidden activation: ReLU.
\item Output layer: a single unit producing the Q-value, using an identity (identity) activation.
\end{itemize}
\item Dynamics model ensemble: the dynamics model consists of an ensemble of ensemble size probabilistic neural networks. Each network in the ensemble predicts the change in state ($\Delta$ state) and the reward.
\begin{itemize}
\item Input layer: matches the sum of observation and action dimensions (from the previous time step).
\item Hidden layers: four hidden layers, each with 200 units.
\item Hidden activation: swish.
\item Output layer: predicts $\Delta$ state and reward. The output dimension for $\Delta$ state matches the observation dimension, plus one for the reward. This layer uses an identity (identity) activation. dynamics models predicts mean and log variance for $\Delta$ state to represent uncertainty. 
\item	Ensemble size: 7 networks.
\item Elite models: 5 elite models are selected based on holdout loss for making predictions during rollouts.
\item Input/output normalization: dynamics models normalize inputs and de-normalize outputs. 
\end{itemize}

\end{itemize}
\subsection{Hyperparameters}

Key hyperparameters for the MB-MPO implementation and the baseline are detailed below. Unless specified otherwise (e.g., for attention $\alpha$ regularization), these values were kept constant across experiments.

\begin{itemize}
\item General and SAC agent parameters:
\begin{itemize}
\item Optimizer (for policy, Q-networks, alpha): Adam
\item	Policy network learning rate: 3e-4
\item Q-network learning rate: 3e-4
\item Discount factor ($\gamma$): 0.99
\item Target smoothing coefficient ($\gamma$): 0.005
\item Initial alpha (entropy coefficient): 0.2 (if not automatically tuned from the start)
\item Reward scale: 1.0
\item	Entropy tuning (SAC alpha):
\begin{itemize}
\item Automatic Tuning: Enabled (automatic tuning: true)
\item Target Entropy: -3 (equal to -action dim for Half-Cheetah)
\item Alpha Optimizer Learning Rate: 3e-4
\end{itemize}
\end{itemize}
\item Dynamics model parameters:
\begin{itemize}
\item Ensemble Size: 7
\item Number of Elite Models: 5
\item Optimizer: Adam
\item Learning Rate: 0.001
\item Batch Size: 256
\item Holdout Ratio (for model validation): 0.2
\item Weight Decay Coefficients: $[2.5e-5, 5e-5, 7.5e-5, 7.5e-5, 0.0001]$ (applied to different layers).
\end{itemize}
\item MB-MPO and Trainer Parameters:
\begin{itemize}
\item Total Training Epochs: 400
\item Environment Steps per Epoch: 1000
\item Agent Update Batch Size: 256 (composed of real and model data)
\item Model/Environment Data Ratio for Agent Updates: 0.95 (meaning 95\% model data, 5\% real data per agent batch)
\item Model Training Interval: 250
\item Number of Agent Updates per Environment Step: 2
\item Rollout batch size (initial states for model rollouts): 100,000
\item Rollout mini-batch Size (for processing rollouts): 10,000
\item Warmup timesteps (random actions): 5000
\item Real Environment Replay Buffer Size: 1,000,000
\item Model-Generated Data Replay Buffer Size (per task/model if meta): 2,000,000
\end{itemize}
\item Meta-Learning Parameters:
\begin{itemize}
\item Number of Training Tasks: 5
\item Number of Tasks Sampled per Meta-Epoch: 5
\item Number of Tasks per Meta-Gradient Update: 5
\item In tasks are small variations in Gravity and Friction  of the surface. 
\end{itemize}
\item Minimum attention regularization:
\begin{itemize}
\item attention $\alpha$: Varied across experiments (e.g., 0 for vanilla, and other values like 0.01, 0.05, 0.1, etc., for adjustment in regularization). This is the main parameter we change.
\end{itemize}
\end{itemize}

\subsection{Evaluation protocol and metrics}
\begin{itemize}
\item General evaluation:
\begin{itemize}
\item We have used 10 random seeds for training runs.
\item For final evaluation of a trained agent, we run for 10 episodes run per seed.
\item Metrics reported: Average total reward, standard deviation/error of reward across seeds/episodes.
\end{itemize}
\item Specific metrics (as discussed)::
\begin{itemize}
\item Average Jacobian norm: during evaluation we run the trained agent for 10 evaluation episode and  averaged $|| \partial u / \partial x ||^2$ per step over 10 evaluation episodes.
\item Average action jerk: define as averaged $|| \partial u / \partial t ||^2$ over 10 evaluation episodes.
\item Average Energy: define as $``$control cost$"$ sum($u^2$) per step.
\end{itemize}
\item Quantification of regularization:
\begin{itemize}
\item Heatmap of the Jacobian norm between the baselines to qualitative judge the impact of minimum attention regularizer.
\end{itemize}
\end{itemize}

\subsection{Specific Experiments Conducted}
\begin{itemize}
\item Meta training:
\begin{itemize}
\item To analyze the effect of attention weight, we trained agents with attention weight values $\alpha \in  \{0, 0.01, 0.05, 1.0, 5.0\}$ while keeping all other hyperparameters same.
\item To analyze the effects of the number of ensemble models, we trained 4 different sizes = $[1, 5, 15, 20]$.
\item To analyze the effect of the imaginary trajectory size in the MB-MPO core loop, we trained 4 different agents $[16, 128, 256, 512]$.
\item Analysis of policy sensitivity during training
\end{itemize}
\item Meta testing:
\begin{itemize}
\item Experiments for ``out of distribution" task and fast adaptation during meta loop test time evaluation.
\item Perturbation in agent: crippled leg, mass change

\item Perturbation in the environment: slope change
\end{itemize}

\item For each of these experiments we note the total reward, standard deviation between runs, average Jacobian norm, average action jerk, and average energy consumed.
\item Heatmap:
\begin{itemize}
\item With the meta trained agent we have evaluated the impact of minimum attention regularizer on the learning process by conducting qualitative analysis by visualizing policy sensitive across different observational dimension of the environment. The goal is to pick dimension that are critical for their respective locomotion and balance challenges. Specifically we have picked
\item For Half-Cheetah 
\begin{itemize}
\item Z-coordinate versus X-velocity
\item X-velocity versus Z-velocity
\item X-velocity versus Y-angular velocity (torso)
\end{itemize}
\item For Hopper 
\begin{itemize}
\item Torso angle versus torso height
\item Torso angle versus torso angular velocity
\item Torso X-velocity versus torso angle
\end{itemize}
\item For Walker2D 
\begin{itemize}
\item Torso height versus torso angle
\item Torso angle  versus torso angular velocity
\item Torso X-velocity versus torso angle
\end{itemize}
\item For Humanoid 
\begin{itemize}
\item Torso height versus torso forward velocity
\item Right hip angle versus left hip angle
\end{itemize}

\end{itemize}
\end{itemize}
%
%\end{enumerate}
%
%
%
\subsection{The details of the algorithm implementation}
In regard to core Model-Based Meta-Policy Optimization (MB-MPO) components, for inner and outer loop policy updates, we have experimented with Proximal Policy Optimization (PPO) \citep{schulman2017}, Trust Region Policy Optimization (TRPO), and Soft Actor-Critic (SAC) \citep{haarnoja2018}. 
We have experimented with 4 different imaginary trajectory sample sizes $N$: 16, 128, 256, and 512. Specifically, the regularization term $|| \partial u_{\theta} / \partial x ||^2 + ||\partial u_{\theta} / \partial t ||^2$, is added to the policy loss with weight $``\alpha"$. We have experimented for $``\alpha"$ taking values 0.01, 0.05, 1.0, and 5.0.\\

\noindent The ``meta-learning” loop essentially follows that of Model-Agnostic Meta-Learning (MAML) structure (inner loop for the perturbation of models, outer loop for the optimization of meta-parameters): tasks (models) are sampled randomly from the task list, the mini-batch of $R_i$ in Algorithm 1. Specifications of the inner loop, and outer loop are in parameters (Appendix A). For model ensemble setup, we have experimented with four different models of ensemble size: 1, 5, 15, and 20. Some details of the neural network architecture are provided in the Appendix A (Dynamics model ensemble).

Quantitatively, the Jacobian term ($\partial u / \partial x$) requires one additional backward pass through the policy network per update, adding approximately 10-12\% to the update time. However, by utilizing stochastic estimation and updating the attention cost every 10 steps, the amortized computational overhead is reduced to less than 2\% of total training time, effectively addressing efficiency concerns raised in evaluation.

\section{Detailed Analysis of Meta-RL with Minimum Attention}
This section provides the full per-environment breakdown with maturation heatmaps and meta-training/meta-testing tables for Hopper, Walker2D, and Humanoid, complementing the HalfCheetah analysis and summary results presented in the main text.

\subsection{Extended Results on Modern World Models}
To supplement the empirical evaluation in the main text, we provide the full performance breakdown for the integration of Minimum Attention (MA) into state-of-the-art world models: MAMBA and DreamerV3. These experiments demonstrate that our regularizer is complementary to latent-space world models, providing gains in both sample efficiency and asymptotic performance.

\begin{table}[hbt]
\caption{Full evaluation of \textbf{MAMBA} on Humanoid-dir. Results report Average Total Rewards $\pm$ standard deviation across 3 random seeds. We evaluate two regularization strengths ($\alpha$) against the unregularized baseline.}
\label{table:appendix_mamba}
\begin{center}
  \small
  \begin{tabular}{|l|c|c|c|}
    \hline
    \textbf{Training Steps} & \textbf{Mamba Baseline} & \textbf{+ MA ($\alpha=0.05$)} & \textbf{+ MA ($\alpha=0.01$)} \\
    \hline
    30M (Final) & 2453 $\pm$ 110 & \textbf{2553 $\pm$ 76} & 2498 $\pm$ 123 \\
    \hline
    10M (Mid) & 1985 $\pm$ 156 & \textbf{2127 $\pm$ 121} & 1965 $\pm$ 136 \\
    \hline
    2.5M (Early) & 467 $\pm$ 288 & \textbf{759 $\pm$ 231} & 415 $\pm$ 245 \\
    \hline
  \end{tabular}
\end{center}
\end{table}

% \begin{table}[hbt]
% \caption{Full evaluation of \textbf{DreamerV3} on HalfCheetah-run. Results report Average Total Rewards $\pm$ standard deviation. Note the significant performance gain in early and final stages of learning.}
% \label{table:appendix_dreamer}
% \begin{center}
%   \small
%   \begin{tabular}{|l|c|c|}
%     \hline
%     \textbf{Training Steps} & \textbf{DreamerV3 Baseline} & \textbf{+ MA ($\alpha=0.05$)} \\
%     \hline
%     500K (Final) & 575 $\pm$ 123 & \textbf{625 $\pm$ 77} \\
%     \hline
%     250K (Mid) & \textbf{408 $\pm$ 86} & 394 $\pm$ 81 \\
%     \hline
%     100K (Early) & 257 $\pm$ 56 & \textbf{308 $\pm$ 59} \\
%     \hline
%   \end{tabular}
% \end{center}
% \end{table}
\begin{table}[hbt]
\caption{Full evaluation of \textbf{DreamerV3} across three DeepMind Control tasks. Results report Average Total Rewards $\pm$ standard deviation across 10 random seeds. Note that the benefit of minimum attention varies by task and training stage.}
\label{table:appendix_dreamer}
\begin{center}
  \small
  \begin{tabular}{|l|l|c|c|}
    \hline
    \textbf{Task} & \textbf{Training Steps} & \textbf{DreamerV3 Baseline} & \textbf{+ MA ($\alpha=0.05$)} \\
    \hline
    \multirow{3}{*}{HalfCheetah-run}
      & 100K (Early)  & 257 $\pm$ 56  & \textbf{308 $\pm$ 59} \\
      & 250K (Mid)    & \textbf{408 $\pm$ 86}  & 394 $\pm$ 81 \\
      & 500K (Final)  & 575 $\pm$ 123 & \textbf{625 $\pm$ 77} \\
    \hline
    \multirow{3}{*}{reacher-hard}
      & 100K (Early)  & \textbf{917 $\pm$ 35} & 906 $\pm$ 42 \\
      & 250K (Mid)    & 874 $\pm$ 20  & \textbf{930 $\pm$ 17} \\
      & 500K (Final)  & \textbf{942 $\pm$ 20} & 931 $\pm$ 17 \\
    \hline
    \multirow{3}{*}{finger-turn-hard}
      & 100K (Early)  & \textbf{341 $\pm$ 40} & 329 $\pm$ 42 \\
      & 250K (Mid)    & 660 $\pm$ 118 & \textbf{826 $\pm$ 96} \\
      & 500K (Final)  & 826 $\pm$ 40  & \textbf{906 $\pm$ 20} \\
    \hline
  \end{tabular}
\end{center}
\end{table}
\noindent As observed in Table \ref{table:appendix_mamba}, a relevant strength of minimum attention ($\alpha$) is essential for high-dimensional stability in Humanoid-dir. DreamerV3 results (Table \ref{table:appendix_dreamer}) show a more task-dependent pattern: minimum attention improves final-stage reward on HalfCheetah-run and finger-turn-hard, with the largest gains typically emerging in the middle-to-late stages of training, while reacher-hard shows comparable performance between baseline and regularized variants throughout.

\begin{figure*}[t]
    \centering
    \begin{subfigure}[b]{0.32\textwidth}
        \centering
        \includegraphics[width=\textwidth]{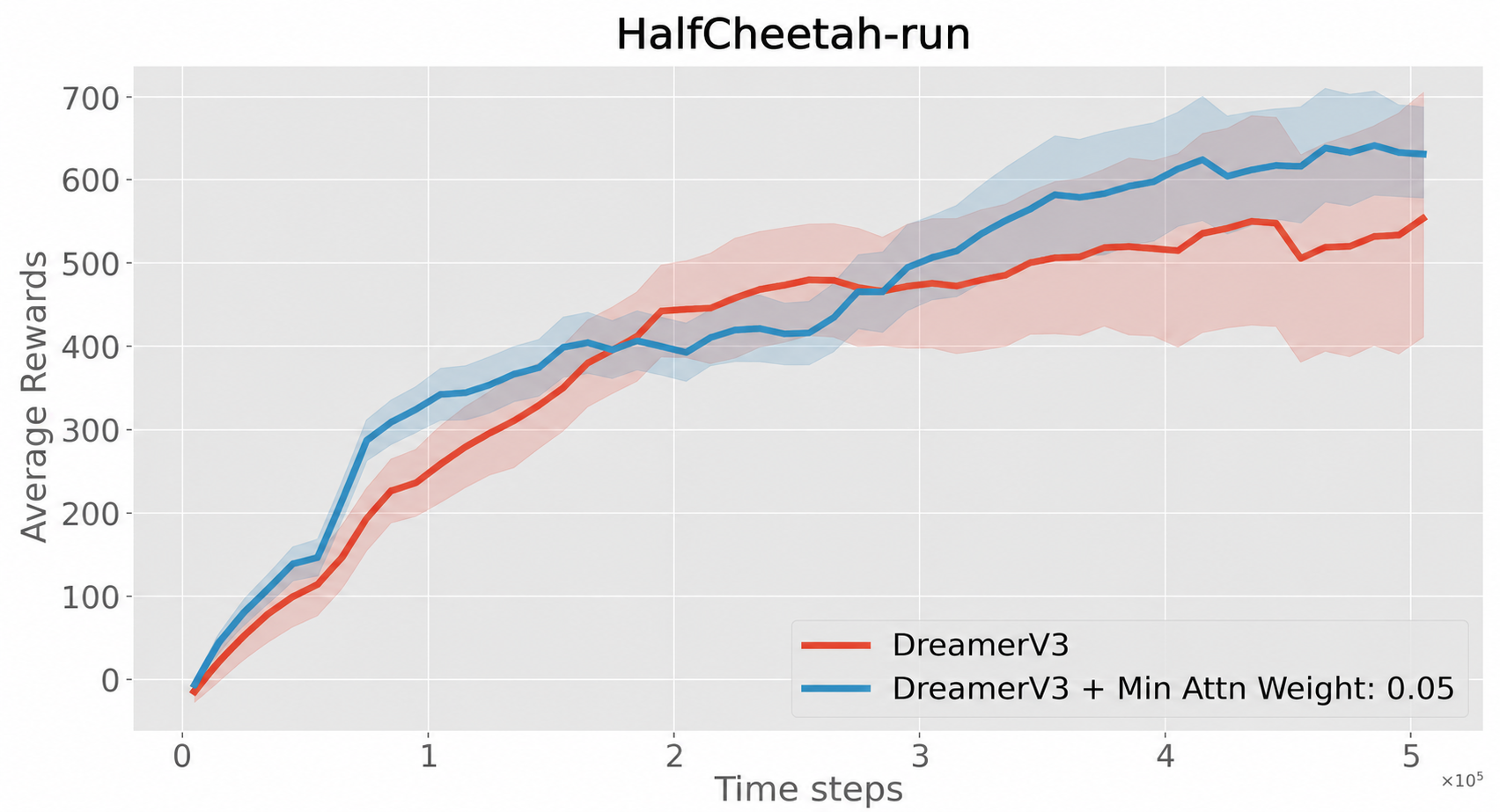}
        \caption{HalfCheetah-run}
    \end{subfigure}
    \hfill
    \begin{subfigure}[b]{0.32\textwidth}
        \centering
        \includegraphics[width=\textwidth]{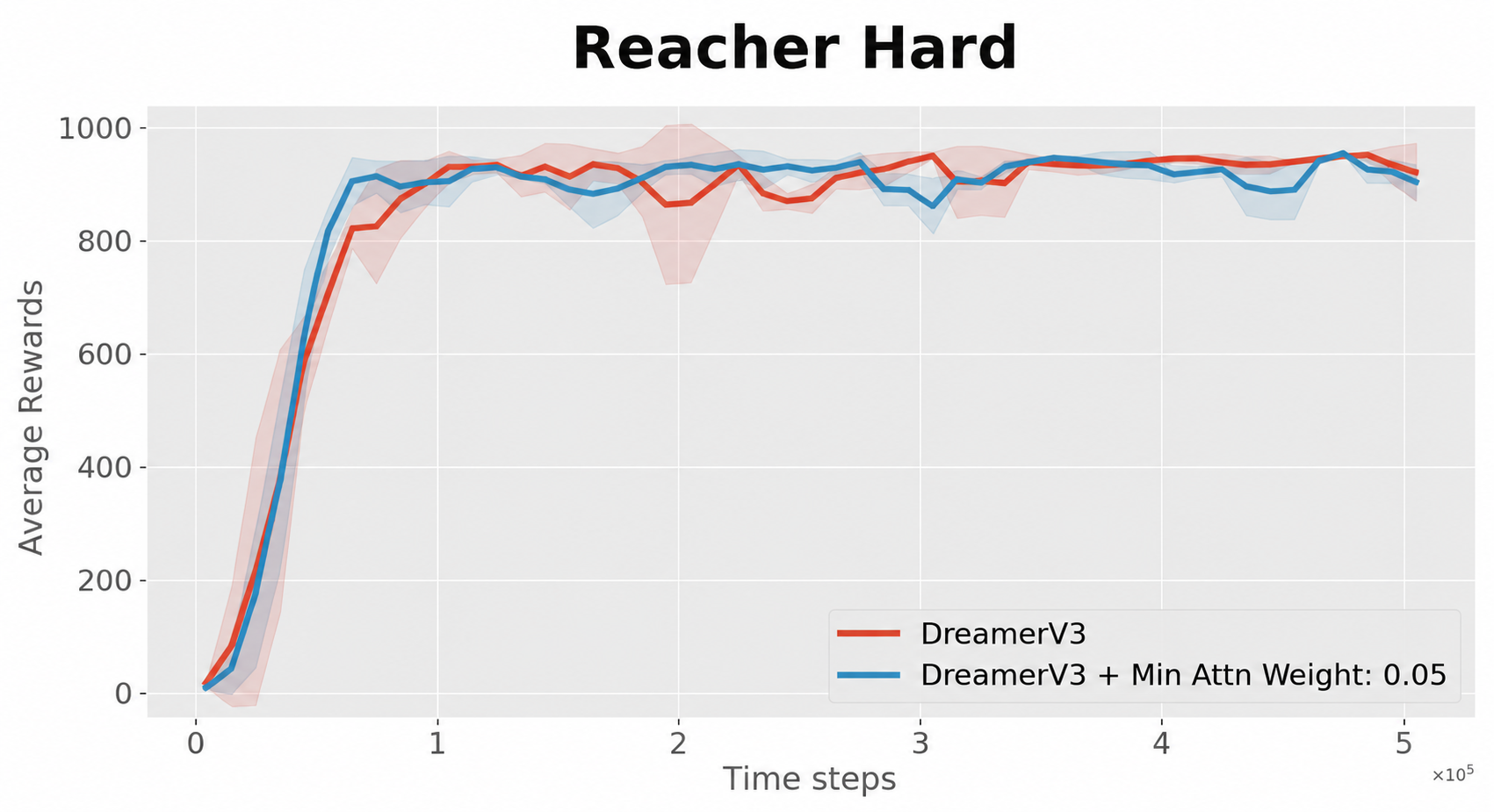}
        \caption{reacher-hard}
    \end{subfigure}
    \hfill
    \begin{subfigure}[b]{0.32\textwidth}
        \centering
        \includegraphics[width=\textwidth]{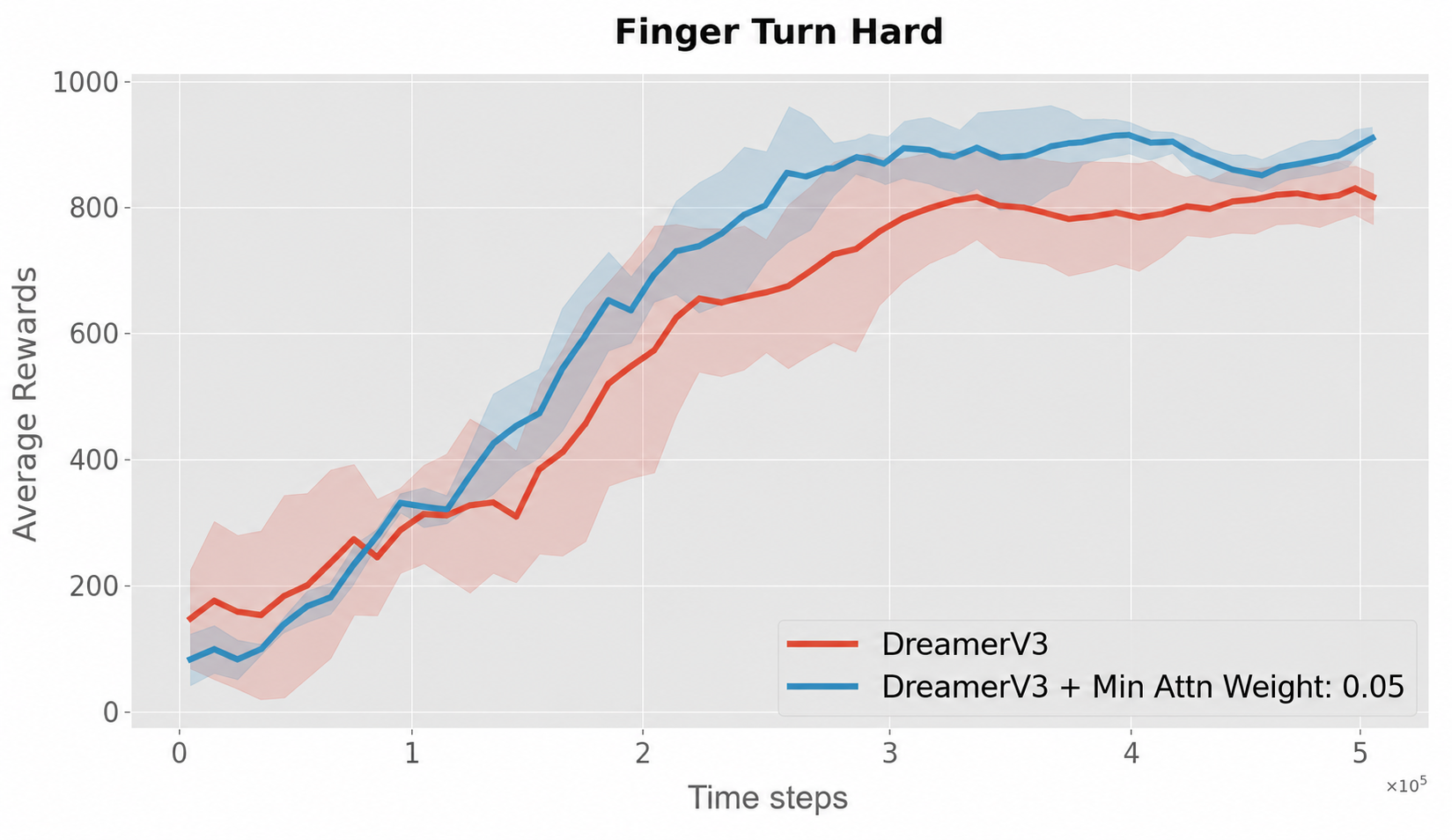}
        \caption{finger-turn-hard}
    \end{subfigure}
    \caption{DreamerV3 training curves across three DeepMind Control tasks, comparing baseline against minimum attention ($\alpha=0.05$). Shaded bands show $\pm 1$ standard deviation across 10 seeds.}
    \label{fig:dreamer_extra_curves}
\end{figure*}

Figure \ref{fig:dreamer_extra_curves} shows the corresponding training curves across all three DreamerV3 tasks. On HalfCheetah-run and finger-turn-hard, minimum attention's curve separates from the baseline and shows a sustained advantage, consistent with the reward gains reported in Table \ref{table:appendix_dreamer}). On reacher-hard, the two curves largely overlap throughout training, with minimum attention showing a modestly narrower band; neither curve shows a sustained advantage over the other, consistent with the near-parity final-stage numbers.

\subsection{Half Cheetah (Figure \ref{model_ensemble_imaginary}, \ref{heatmap_feedback_evolution}, \ref{heatmap_feedforward_evolution}, Table \ref{half_cheetah:meta_training}, \ref{half_cheetah:meta_testing}) }

\subsubsection*{Maturation of learned control strategies}
To understand how the final control strategies emerge, we visualize the evolution of both spatial sensitivity (feedback/Jacobian norm) and temporal sensitivity (feedforward/jerk) at different stages of training. Figures \ref{heatmap_feedback_evolution} and \ref{heatmap_feedforward_evolution} show the policy heatmaps for HalfCheetah at 10K, 100K, and 200K epochs for both vanilla MB-MPO and our regularized method.\\

\begin{itemize}

\item  Maturation of feedback (Jacobian) strategy (Figure \ref{heatmap_feedback_evolution}):
At the early stage of training (10K epochs), both policies exhibit broad, unstructured sensitivity. As training progresses, a clear divergence in strategy appears. The vanilla MB-MPO policy (top row) develops a wide, high-sensitivity band at high forward velocities, indicating it has learned a reactive strategy that requires constant, high-gain corrections to maintain a running gait. In contrast, our regularized policy (bottom row) converges to a much sparser and more structured landscape. It learns to be largely insensitive across most of the state space, focusing its sensitivity into a narrow, precise band corresponding to the apex of its running motion. This shows the minimum attention actively guides the policy towards a simpler, more efficient state-action mapping.

\item Maturation of feedforward (jerk) strategy (Figure \ref{heatmap_feedforward_evolution}):
The evolution of temporal sensitivity is even more striking. The vanilla MB-MPO policy (top row) begins with extremely high, undifferentiated jerk at low speeds (10K epochs), essentially an exploratory behavior. As it learns, it suppresses this initial jerk but retains a moderately high level of action change during the low-crouch running phase (200K epochs).

\noindent The minimum attention policy (bottom row) follows a different developmental path. While it also explores with high jerk initially, the minimum attention quickly shapes the policy to be temporally smoother overall. By the final stage (200K epochs), it has learned a highly specialized strategy: the jerk is concentrated into a very sharp, narrow band at a specific torso height. This suggests the agent has learned an efficient, ballistic motion, where it "coasts" with smooth actions for most of the gait cycle and applies a single, decisive, and well-timed control change at the critical take-off point.\\

\end{itemize}

\subsubsection*{Additional feedback--feedforward decomposition}
\label{app:kv_additional}

In addition to the crippled-back adaptation analysis in the main paper, we
include further feedback--feedforward decompositions for maturation and mass
perturbation. These plots use the same control decomposition
$u(x,t)=K(t)x+v(t)$, where $\|K(t)\|_F$ measures the state-dependent feedback
intensity and $\|v(t)\|_2$ measures the feedforward open-loop control.

\begin{figure*}[t]
    \centering
    \includegraphics[width=0.95\textwidth]{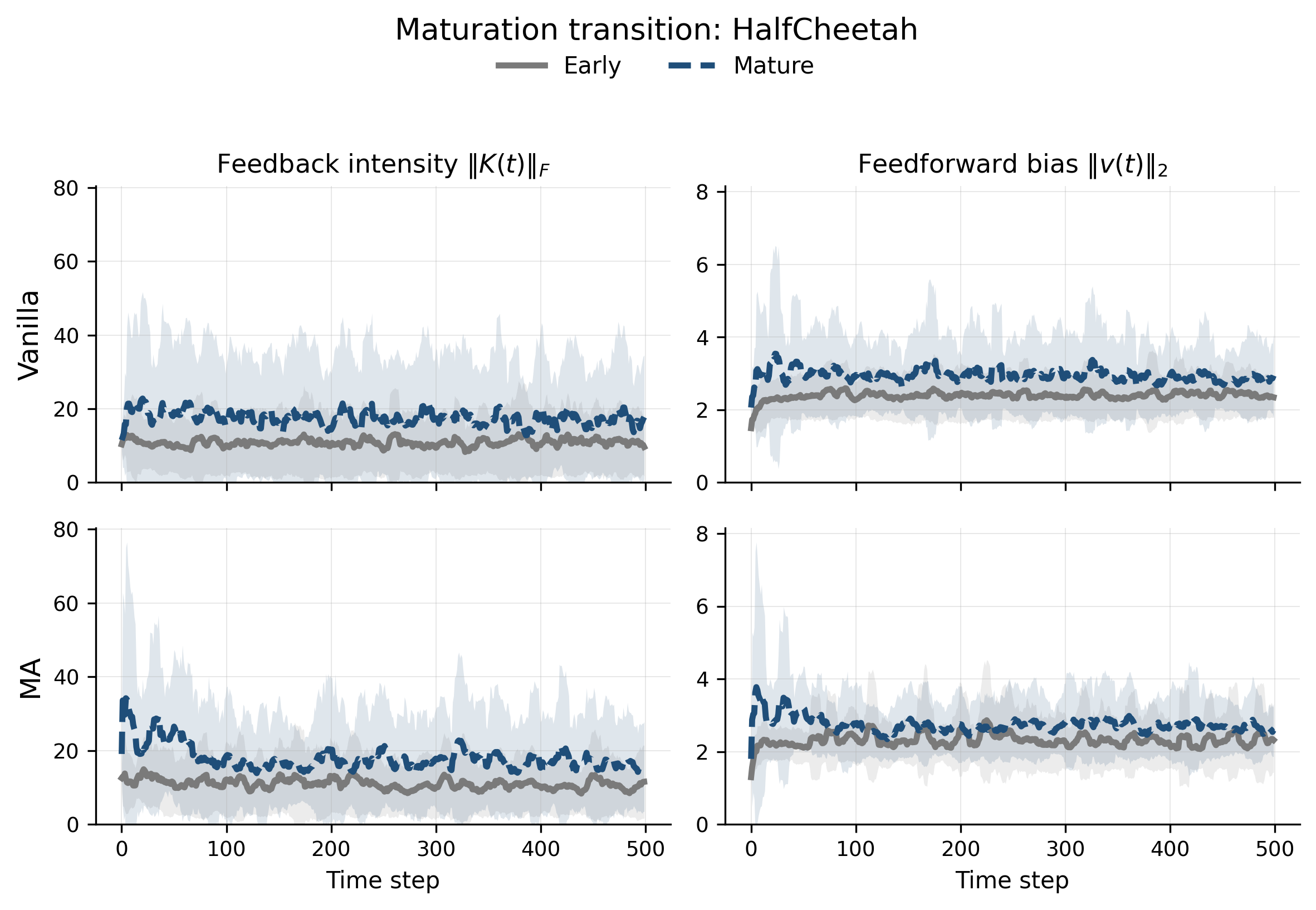}
    \caption{
    Feedback--feedforward decomposition during policy maturation for
    HalfCheetah-v3. We visualize $\|K(t)\|_F$ and $\|v(t)\|_2$ in the
    control law $u(x,t)=K(t)x+v(t)$ for early and mature policies. The
    comparison shows how training changes both the feedback and feedforward
    components over rollout time. MA regularization produces a more
    constrained dispersion profile, suggesting that the learned controller
    matures toward a less variable control structure.
    }
    \label{fig:kv_maturation}
\end{figure*}

Figure \ref{fig:kv_maturation} compares early and mature policies. Across both
vanilla and minimum-attention (MA) policies, maturation changes the magnitude
and variability of both feedback and feedforward components, indicating that
the learned controller does not simply converge in reward space but also
develops a more structured control profile over rollout time. In the
MA-regularized policy, this maturation is accompanied by a more constrained
dispersion profile, consistent with the role of minimum attention as a
regularizer on variations in control.

\begin{figure*}[t]
    \centering
    \includegraphics[width=0.95\textwidth]{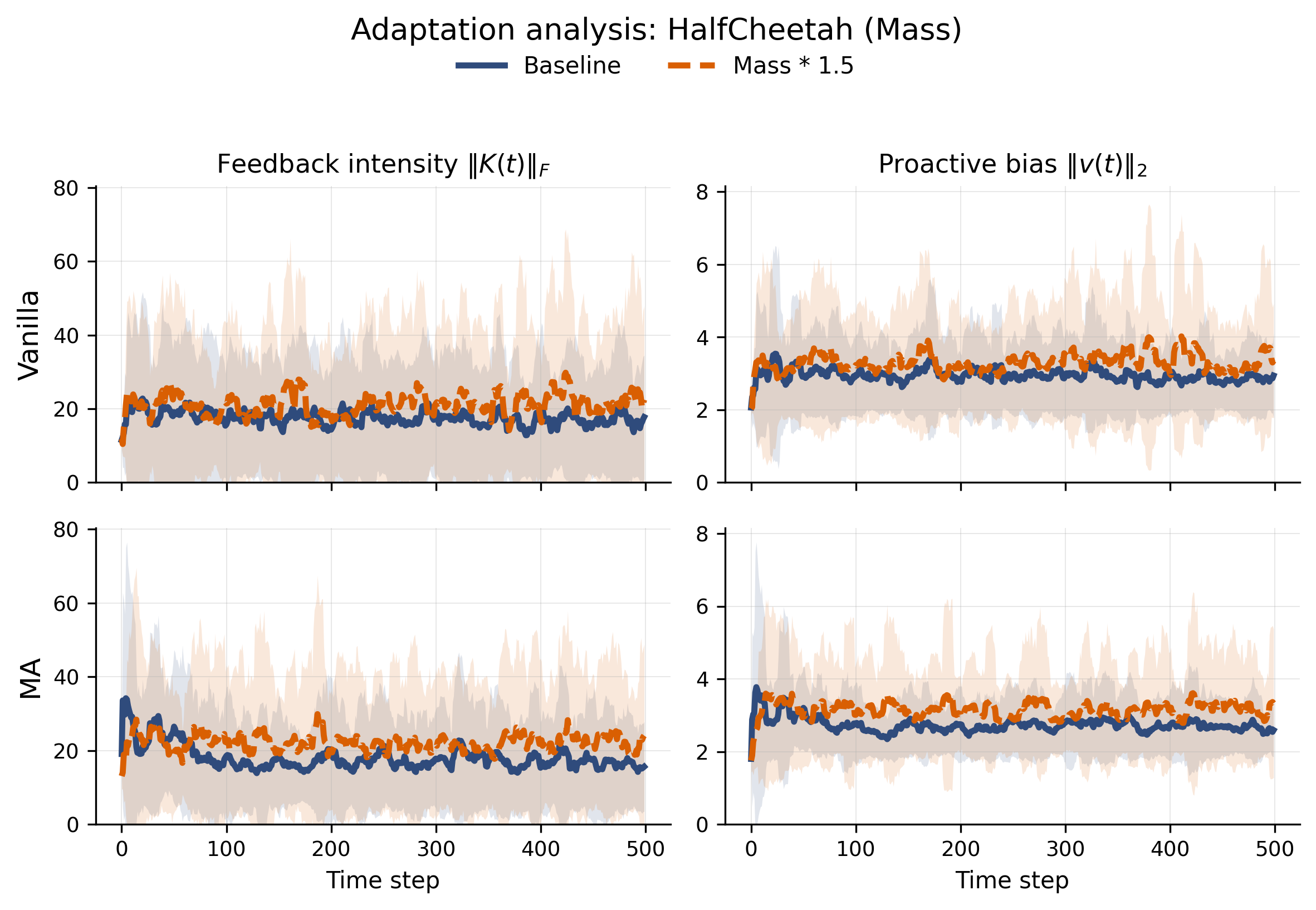}
    \caption{
    Feedback--feedforward decomposition for HalfCheetah-v3 under mass
    perturbation, where the body mass is scaled by $1.5$. Rows compare
    vanilla and minimum-attention (MA) policies, and columns show the feedback
    intensity $\|K(t)\|_F$ and feedforward bias $\|v(t)\|_2$. The trends are
    consistent with the crippled-back perturbation in the main paper: MA
    maintains a more regularized feedback--feedforward profile while preserving
    temporal adjustments needed to adapt to altered dynamics.
    }
    \label{fig:kv_mass15}
\end{figure*}

Figure \ref{fig:kv_mass15} shows the corresponding adaptation analysis under a
mass perturbation, where the HalfCheetah body mass is scaled by $1.5$. The
qualitative trends are consistent with the crippled-back case in the main
paper: the perturbed controller requires temporal adjustment in both feedback
and feedforward components, while MA maintains a more regularized
feedback--feedforward profile. This supports the interpretation that MA reduces
unnecessary variability without removing the corrective responses required for
adaptation to altered dynamics.

\begin{figure}[hbt]
\begin{center}
\includegraphics[angle=0, width=1\textwidth]{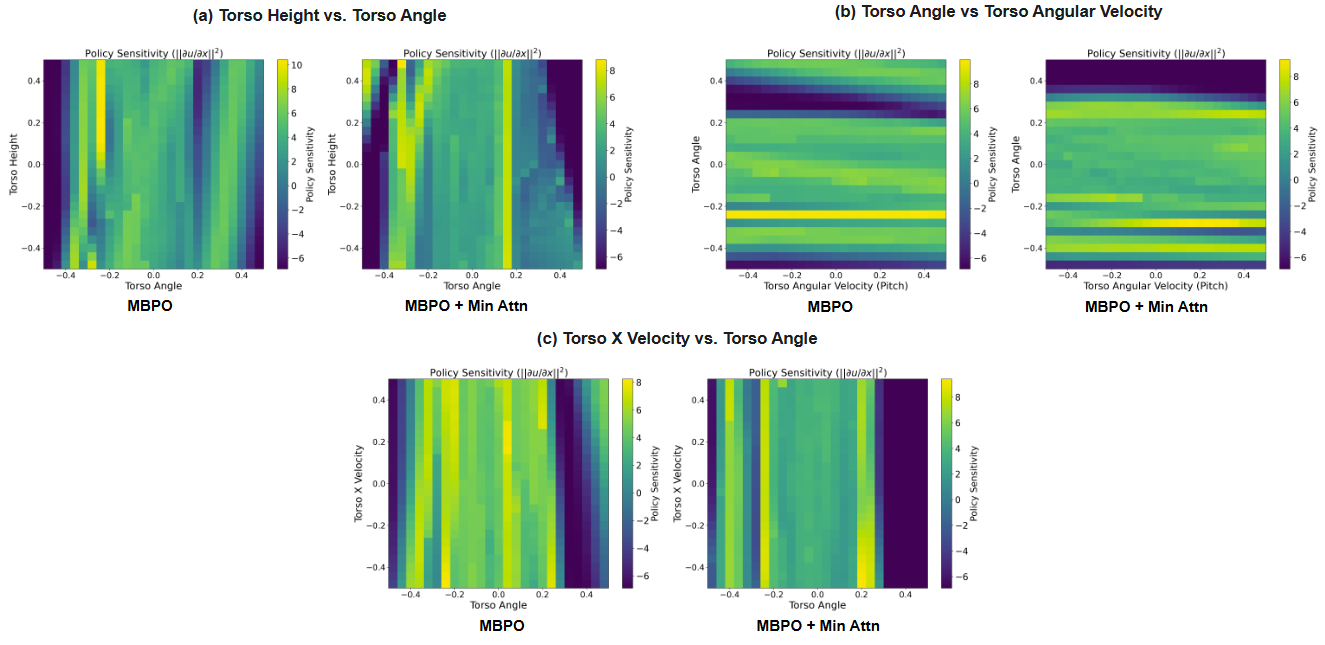}
\end{center}
\caption{Hopper: Heatmap of feedback. The feedback ($||\partial u / \partial x||^2$) in the projection of the states to (a) Torso-angle versus Torso-height, (b) Torso-angle versus Torso X-velocity, and (c) Torso-angular velocity versus Torso-angle.} \label{fig:hopper_heatmap}
\end{figure}

\begin{figure}[hbt]
\begin{center}
\includegraphics[angle=0, width=1\textwidth]{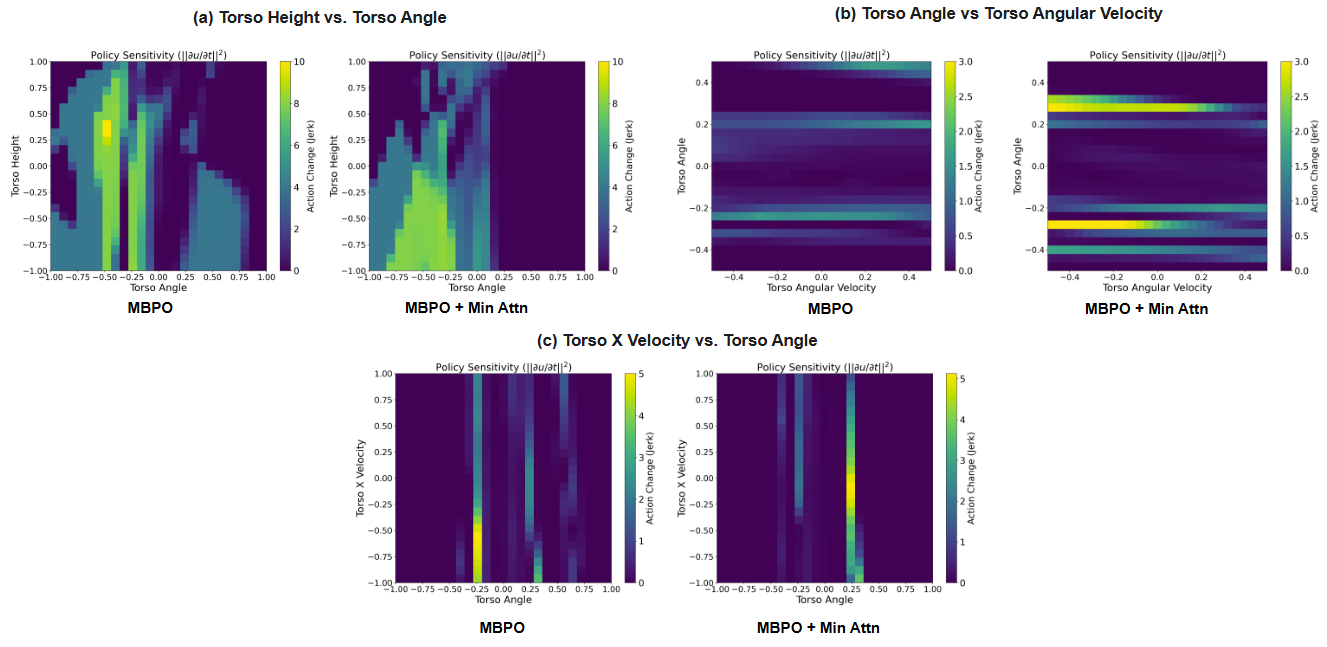}
\end{center}
\caption{Hopper: Heatmap of feedforward. The feedforward ($||\partial u / \partial t||^2$) in the projection of the states to (a) Torso-angle versus Torso-height, (b) Torso-angle versus Torso X-velocity, and (c) Torso-angular velocity versus Torso-angle.} \label{fig:hopper_heatmap_jerk}
\end{figure}

\begin{figure}[hbt]
\centering
%\begin{subfigure}[b]{0.325\textwidth}
\begin{subfigure}[b]{0.48\textwidth} 
   \centering
   \includegraphics[width=\textwidth]{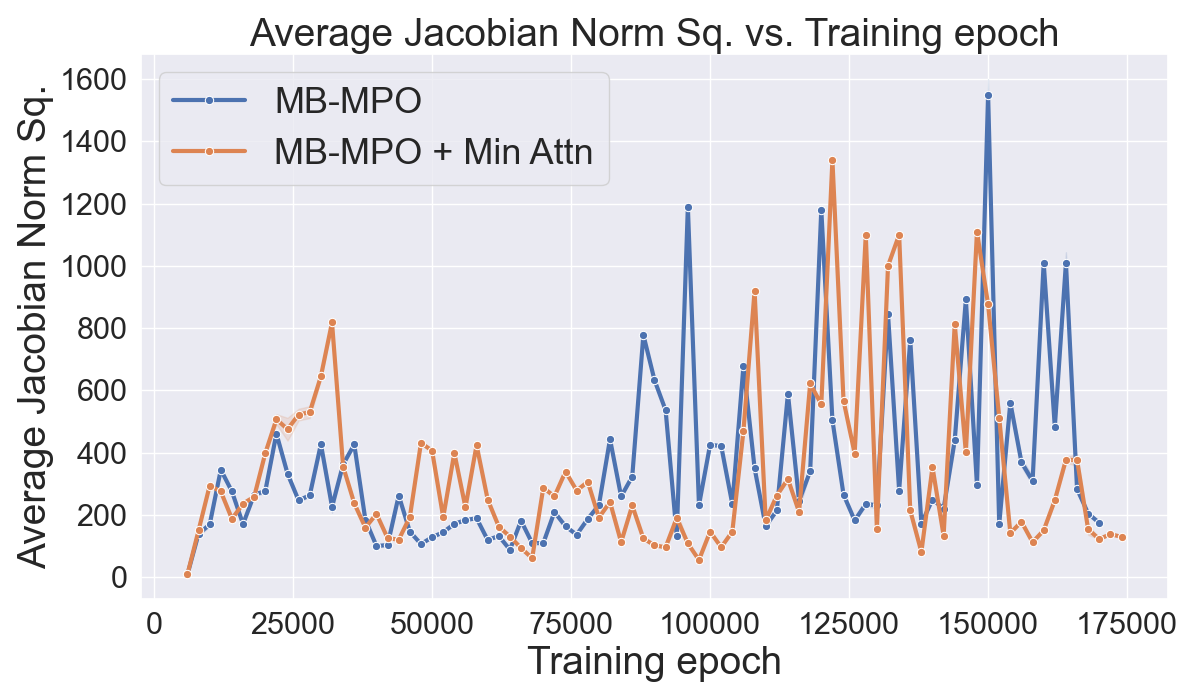}
   \label{fig:figure3a}
\end{subfigure}
\hfill 
%\begin{subfigure}[b]{0.325\textwidth}
\begin{subfigure}[b]{0.48\textwidth} 
   \centering
   \includegraphics[width=\textwidth]{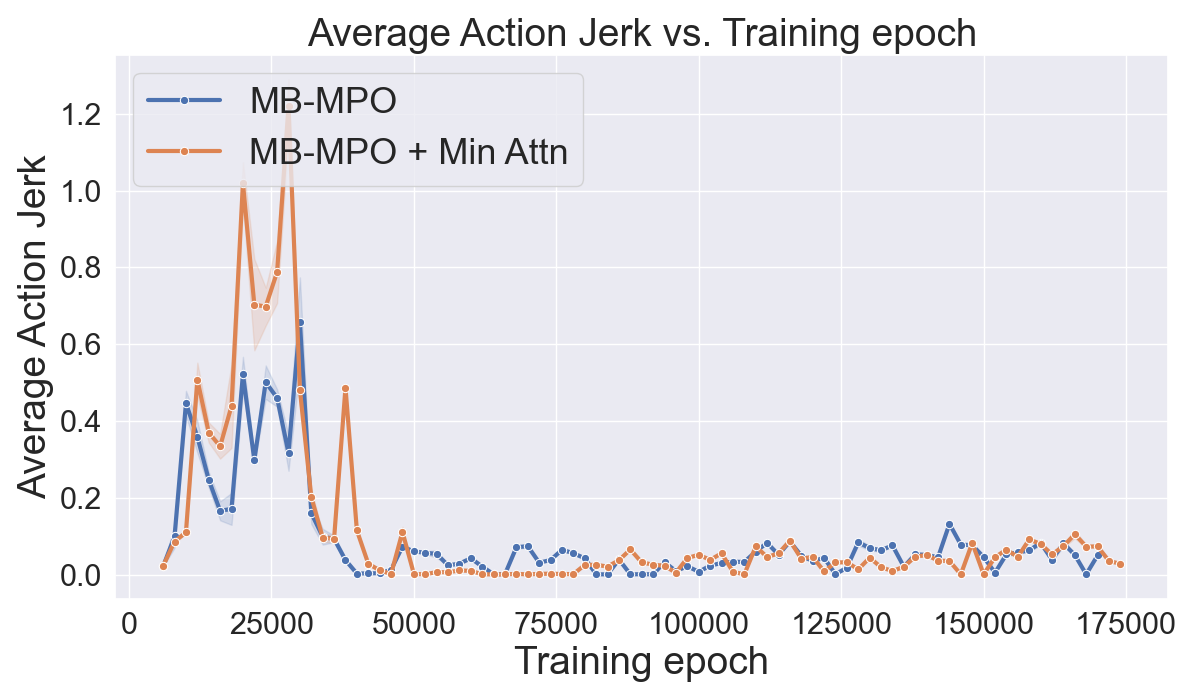} 
   \label{fig:figure3b}
\end{subfigure}
\hfill 
%\begin{subfigure}[b]{0.325\textwidth}
\begin{subfigure}[b]{0.48\textwidth} 
   \centering
   \includegraphics[width=\textwidth]{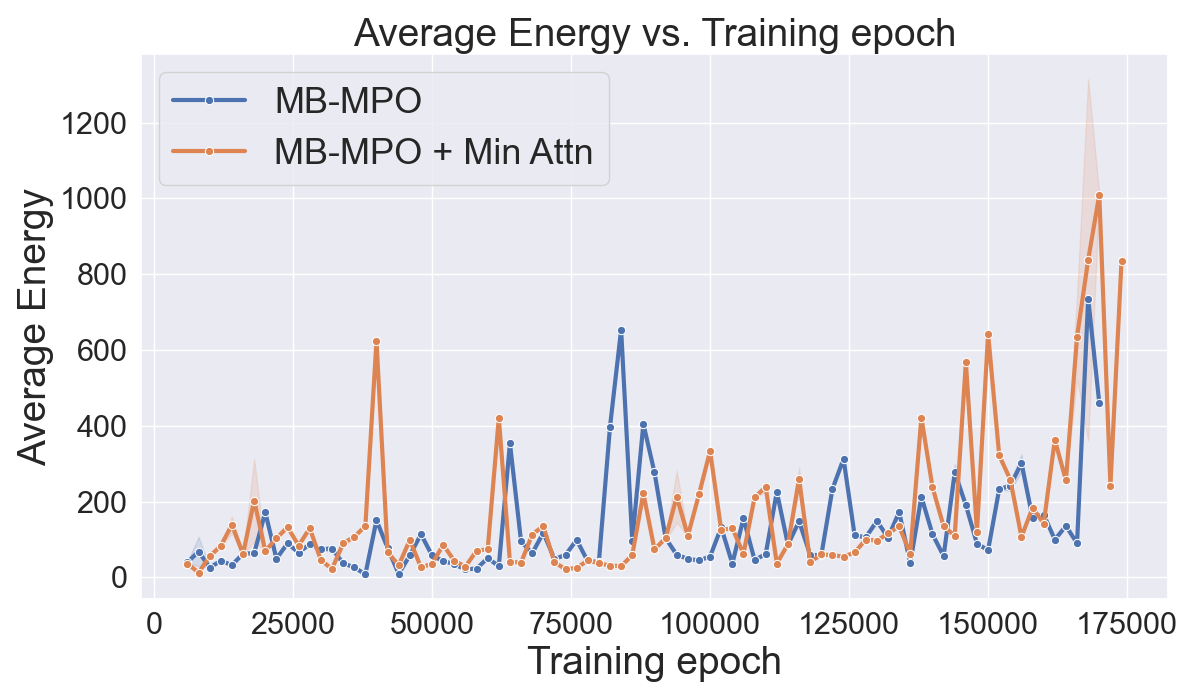} 
   \label{fig:heatmap_hopper}
\end{subfigure}

\caption{Hopper: Time profiles for the feedback, feedforward, and energy in meta-training. The comparison of feedback ($||\partial u / \partial x||^2$, Jacobian), feedforward ($||\partial u/ \partial t||^2$, Jerk), and energy ($||u||^2$) between MB-MPO and MB-MPO with minimum attention ($\alpha = 1.0$).}
\label{time_profile_hopper}
\end{figure}

\subsection{Hopper (Figure \ref{fig:hopper_heatmap}, \ref{fig:hopper_heatmap_jerk}, \ref{time_profile_hopper}, \ref{fig:hopper_trajectory}, Table \ref{table:hopper:meta-training}, \ref{table:hopper:meta-testing})}

For the Hopper, the regularizer enforces a more structured and efficient balancing strategy.
\begin{itemize}
    \item \textbf{(a) Torso angle vs. height:} The vanilla policy is highly reactive when the Hopper is leaning backward (negative Torso Angle), a critical state for falling. The regularized policy significantly dampens this peak sensitivity, suggesting it has learned a smoother, less ``panicked'' method for balance recovery.
    \item \textbf{(b) Torso angle vs. angular velocity:} In the critical balancing phase space, the regularized policy shows more focused, distinct bands of sensitivity. This is in contrast to the broader, more diffuse sensitivity of the vanilla policy, indicating a more specialized and precise reactive control for managing angular momentum.
    \item \textbf{(c) Torso X-velocity vs. torso angle:} The sensitivity patterns of the regularized policy are visibly cleaner and more structured, suggesting a more coherent and principled relationship between forward speed and balancing posture compared to the patchy, reactive strategy learned by the baseline.
\end{itemize}

\begin{table}[hbt]
\caption{Hopper: Meta-training} \label{table:hopper:meta-training}
\begin{center}
  \small
  \begin{tabular}{|l|c|c|}
    \hline
           & MB-MPO & Ours ($\alpha = 1.0$) \\
    \hline
    Early stage of epoch: 10K & & \\
    \hline    
      Average total rewards   & 47   $\pm$ 2 & \textbf{232 $\pm$ 5.75} \\
     \hline
     Average feedback normalized  & \textbf{168 $\pm$ 5.26} & 290 $\pm$ 1.82 \\
    \hline
     Average feedforward normalized  & 0.45 $\pm$ 0.026 & \textbf{0.11 $\pm$ 0.003}  \\
    \hline
%     Average energy normalized  & \textbf{25.9 $\pm$ 0.27} & 55 $\pm$ 2.23   \\
        Average energy normalized  & \textbf{0.26 $\pm$ 0.0027} & 0.55 $\pm$ 0.022   \\
    \hline
    Middle stage of epoch: 100K & &\\
     \hline   
      Average total rewards   & 222   $\pm$ 0.68 & \textbf{688 $\pm$ 1.92} \\
     \hline
     Average feedback normalized  & \textbf{42.6 $\pm$ 3.47} & 55.1 $\pm$ 0.68 \\
    \hline
     Average feedforward normalized  & \textbf{0.007 $\pm$ 0.003} & 0.04 $\pm$ 0.002  \\
    \hline
%     Average energy normalized  & \textbf{55.4 $\pm$ 0.388} & 219 $\pm$ 1.66  \\
         Average energy normalized  & \textbf{0.55 $\pm$ 0.0039} & 21.9 $\pm$ 0.017  \\

    \hline
     Final stage of epoch: 200K & &\\
    \hline    
      Average total rewards   & 2475   $\pm$ 3.81 & \textbf{2825 $\pm$ 1.25} \\
     \hline
     Average feedback normalized  & 507 $\pm$ 2.23 & \textbf{153 $\pm$ 0.73} \\
    \hline
     Average feedforward normalized  & 0.05 $\pm$ 0.008 & \textbf{0.04 $\pm$ 0.002}  \\
    \hline
%     Average energy normalized  & 573 $\pm$ 2.28 & \textbf{330 $\pm$ 0.93}   \\
        Average energy normalized  & 5.73 $\pm$ 0.023 & \textbf{3.30 $\pm$ 0.0093}   \\

    \hline        
  \end{tabular}
\end{center}
\end{table}
\begin{table}[hbt]
\caption{Hopper: Meta-testing} \label{table:hopper:meta-testing}
\begin{center}
  \small
  \begin{tabular}{|l|c|c|}
    \hline
           & MB-MPO & Ours ($\alpha = 1.0$) \\
    \hline
        Early stage of epoch: 10K    &  &\\  
    \hline
      Average total rewards   & 234   $\pm$ 3.39 & \textbf{278 $\pm$ 3.15} \\
     \hline
     Average feedback normalized  & \textbf{99.7 $\pm$ 2.15} & 137 $\pm$ 5.25 \\
    \hline
     Average feedforward normalized  & 0.11 $\pm$ 0.003 & \textbf{0.03 $\pm$ 0.001}  \\
    \hline
%     Average energy normalized  & 152 $\pm$ 5.75 & \textbf{114 $\pm$ 1.54}   \\
        Average energy normalized  & 1.52 $\pm$ 0.058 & \textbf{1.14 $\pm$ 0.015}   \\
    \hline  
      Middle stage of epoch: 100K & &\\
    \hline
      Average total rewards   & 323   $\pm$ 7 & \textbf{394 $\pm$ 6.85} \\
     \hline
     Average feedback normalized  & \textbf{49 $\pm$ 3.67} & 89 $\pm$ 1.26 \\
    \hline
     Average feedforward normalized  & 0.11 $\pm$ 0.004 & \textbf{0.02 $\pm$ 0.003}  \\
    \hline
%     Average energy normalized  & 148 $\pm$ 5.11 & \textbf{122 $\pm$ 10.11}   \\
        Average energy normalized  & 1.48 $\pm$ 0.051 & \textbf{1.22 $\pm$ 0.10}   \\

    \hline  
     Final stage of epoch: 200K & &\\      
    \hline
      Average total rewards   & 467   $\pm$ 100 & \textbf{485 $\pm$ 61} \\
     \hline
     Average feedback normalized  & 162 $\pm$ 27 & \textbf{140 $\pm$ 10.6} \\
    \hline
     Average feedforward normalized  & 0.04 $\pm$ 0.01 & \textbf{0.02 $\pm$ 0.002}  \\
    \hline
%     Average energy normalized  & 258 $\pm$ 38 & \textbf{77 $\pm$ 1.1}   \\
        Average energy normalized  & 2.58 $\pm$ 0.38 & \textbf{0.77 $\pm$ 0.011}   \\

    \hline        
  \end{tabular}
\end{center}
\end{table}

\subsubsection*{Dynamic consequences in Hopper}
\begin{figure}[hbt!]
    \centering
 \begin{subfigure}[b]{\textwidth} 
    \centering
    \includegraphics[width=\textwidth]{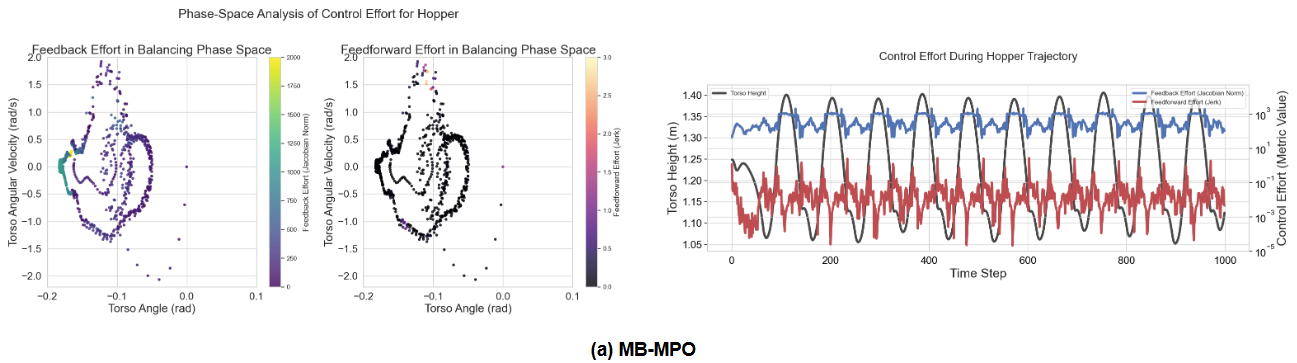}
\end{subfigure}
\hfill 
\begin{subfigure}[b]{\textwidth} 
    \centering
    \includegraphics[width=\textwidth]{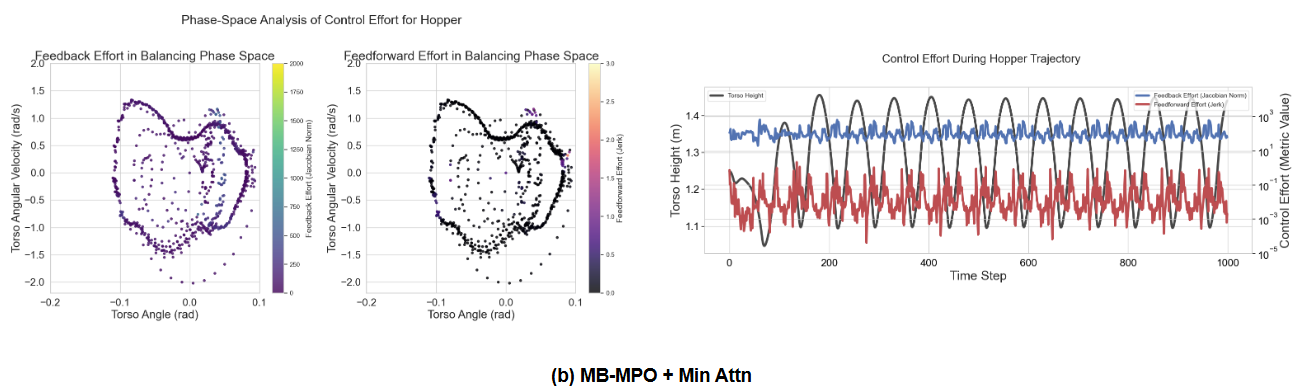} 
\end{subfigure}    
    \caption{Hopper. Left Column: Phase-space analysis of control effort for Hopper's balancing motion. \textbf{Top Row (MB-MPO):} A "panic" strategy with an extreme feedback hotspot ($>$ 2000) for backward leans. \textbf{Bottom Row (Ours):} A smooth, proactive strategy with distributed, low-gain feedback, eliminating the need for high-gain emergency corrections. Right Column: Control effort during a Hopper trajectory. \textbf{Top (MB-MPO):} Shows massive, periodic feedback spikes at the bottom of each hop. \textbf{Bottom (Ours):} Exhibits a more regular gait with dramatically lower and smoother control effort.}
    \label{fig:hopper_trajectory}
\end{figure}

These fundamental changes in the control landscape lead to qualitatively different and more stable behaviors. We illustrate this using phase-space and time-series analyses of the Hopper agent in Figure \ref{fig:hopper_trajectory}. This figure provides both a phase-space portrait of the balancing strategy and a time-series view of the control effort throughout the hopping gait.\\

\noindent The top row reveals the vanilla MB-MPO's brittle, "panic correction" strategy. The phase-space plot (top left) is dominated by an extreme hotspot of feedback effort (Jacobian norm $>$ 2000), which corresponds to a last-ditch, high-gain reaction to prevent falling when the agent is leaned far back. The time-series plot (top right) shows the dynamic signature of this strategy: a somewhat irregular hopping pattern punctuated by massive, periodic spikes in feedback effort that align perfectly with the bottom of each hop.\\

\noindent In stark contrast, the bottom row demonstrates the superior strategy learned by our regularized agent. The phase-space portrait (bottom left) is entirely free of the "panic" hotspot; feedback effort is now distributed smoothly and proactively across the balancing cycle with a much lower peak magnitude ($<$700). The corresponding time-series (bottom right) shows the result: a more regular, higher, and more stable hopping gait, achieved with dramatically lower and smoother control effort. This comparison clearly visualizes the shift from a novice's reactive, high-impact hopping to an expert's efficient, spring-like motion, directly confirming the benefits of the minimum attention.\\

\noindent These evolutionary snapshots reveal that the minimum attention does more than just suppress final sensitivity values. It fundamentally alters the learning process, guiding the policy away from generic, reactive solutions and towards the discovery of specialized, structured, and efficient motor skills that distinguish expert control from novice behavior.

\begin{figure}[hbt]
\begin{center}
\includegraphics[angle=0, width=1.0\textwidth]{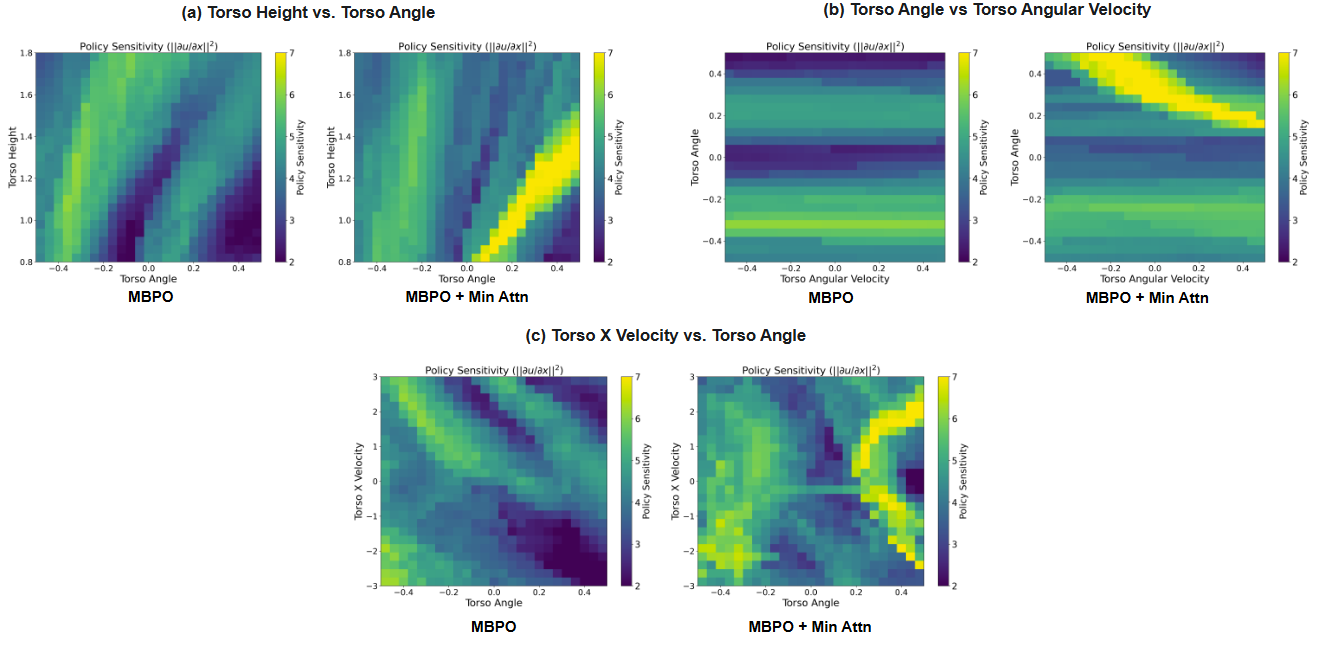}
\end{center}
\caption{Walker2D: Heatmap of feedback. The feedback ($||\partial u / \partial x||^2$) in the projection of the states to (a) Torso-angle versus Torso-height, (b) Torso-angle versus Torse X-velocity, and (c) Torso-angular velocity versus Torso-angle.}
\label{fig:walker2d_heatmap}
\end{figure}

\begin{figure}[hbt]
\begin{center}
\includegraphics[angle=0, width=1.0\textwidth]{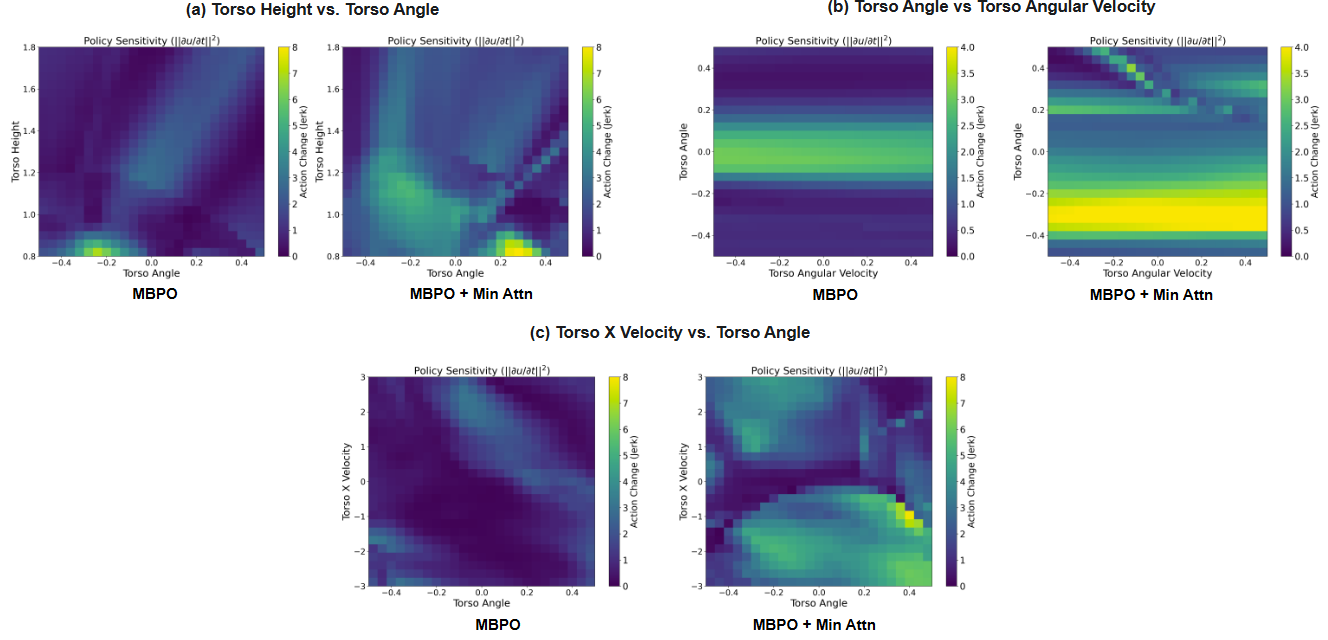}
\end{center}
\caption{Walker2D: Heatmap of feedforward. The feedforward ($||\partial u / \partial t||^2$) in the projection of the states to (a) Torso-angle versus Torso-height, (b) Torso-angle versus Torse X-velocity, and (c) Torso-angular velocity versus Torso-angle.}
\label{fig:walker2d_heatmap_jerk}
\end{figure}

\begin{figure}[hbt]
\centering
%\begin{subfigure}[b]{0.325\textwidth}
\begin{subfigure}[b]{0.48\textwidth} 
   \centering
   \includegraphics[width=\textwidth]{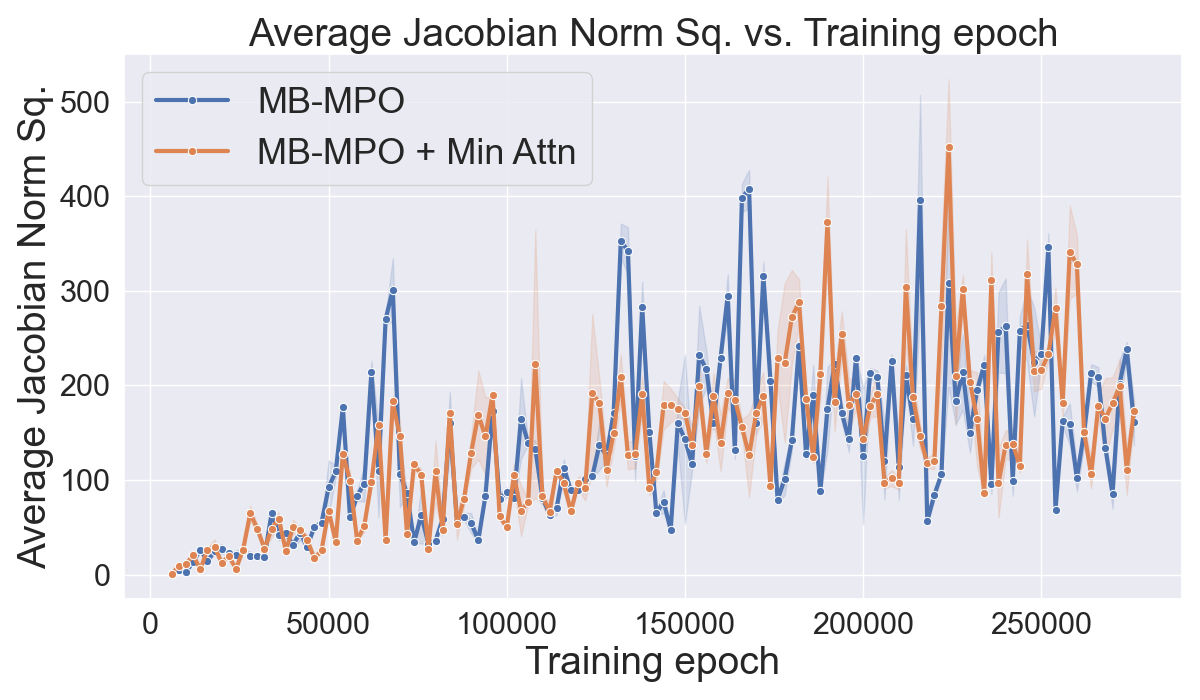}
   \label{fig:figure3a}
\end{subfigure}
\hfill 
%\begin{subfigure}[b]{0.325\textwidth}
\begin{subfigure}[b]{0.48\textwidth} 
   \centering
   \includegraphics[width=\textwidth]{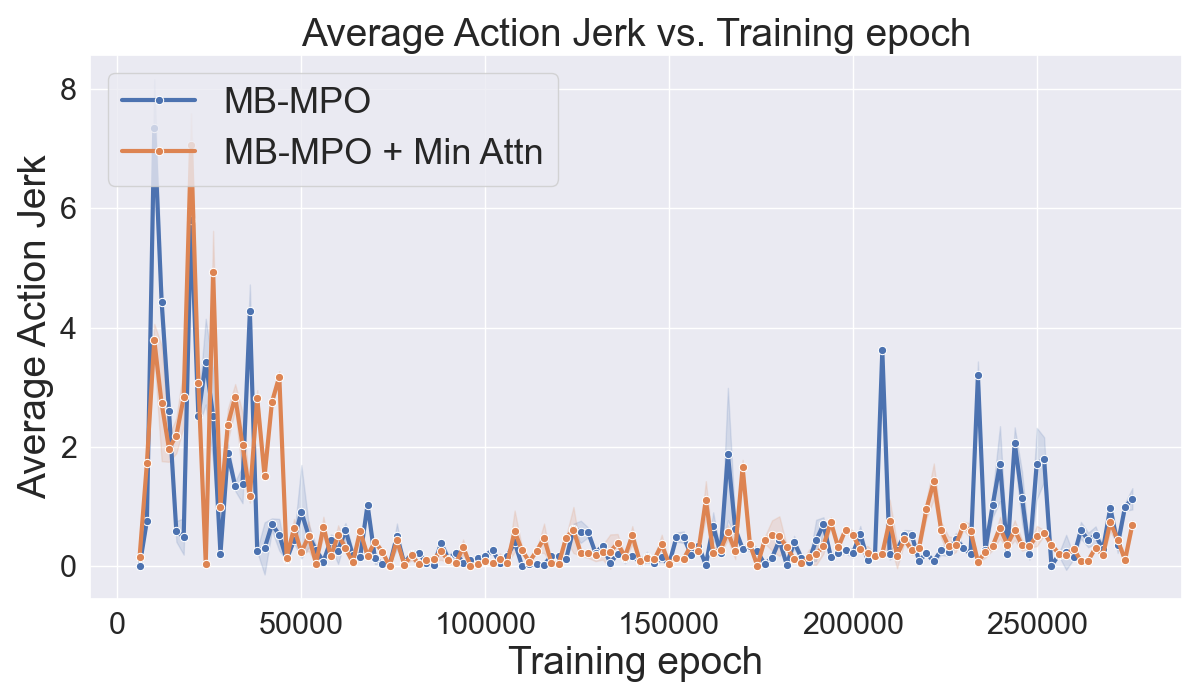} 
   \label{fig:figure3b}
\end{subfigure}
\hfill 
%\begin{subfigure}[b]{0.325\textwidth}
\begin{subfigure}[b]{0.48\textwidth} 
   \centering
   \includegraphics[width=\textwidth]{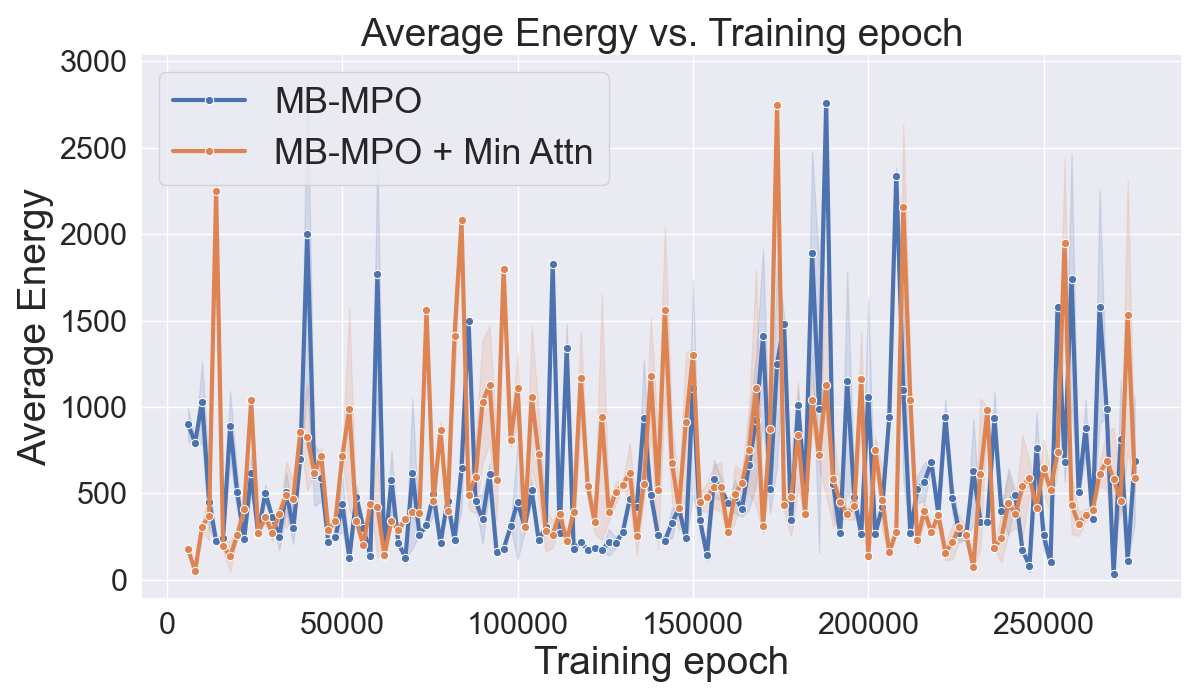} 
   \label{fig:heatmap_walker}
\end{subfigure}
\caption{Walker2D: Time profiles for the feedback, feedforward, and energy in meta-training. The comparison of feedback ($||\partial u / \partial x||^2$, Jacobian), feedforward ($||\partial u/ \partial t||^2$, Jerk), and energy ($||u||^2$) between MB-MPO and MB-MPO with minimum attention ($\alpha = 1.0$).}
\label{time_profile_walker2d}
\end{figure}

\subsection{Walker2D (Figure \ref{fig:walker2d_heatmap}, \ref{fig:walker2d_heatmap_jerk}, \ref{time_profile_walker2d}, Table \ref{table:Walker2D:meta-training}, \ref{table:Walker2D:meta-testing})}

In the Walker2D environment, the regularization induces a fundamental shift in control strategy, moving from reactive fall-prevention to proactive gait management.
\begin{itemize}
    \item \textbf{(a) Torso angle vs. height:} The most striking change is that our regularized policy almost completely eliminates the high sensitivity to backward leans that dominates the vanilla policy. Instead, it develops a new, strong diagonal band of sensitivity corresponding to states with \textit{forward lean} and changing torso height. This suggests a strategic shift from simple fall correction to proactive control of posture during the forward phase of a stride.
    \item \textbf{(b) Torso angle vs. angular velocity:} The regularized policy concentrates its sensitivity in a distinct region of high forward lean and positive angular velocity. This indicates it has learned to pay specific attention to controlling its posture at the apex of a forward stride, a hallmark of a more advanced and efficient gait.
    \item \textbf{(c) Torso X-velocity vs. torso angle:} Again, the regularized policy's sensitivity map is significantly more structured, reinforcing the conclusion that it has learned a smoother, more principled gait strategy that better coordinates speed and balance, rather than relying on a patchwork of reactive corrections.
\end{itemize}

\begin{table}[hbt]
\caption{Walker2D: Meta-training} \label{table:Walker2D:meta-training}
\begin{center}
  \small
  \begin{tabular}{|l|c|c|}
    \hline
           & MB-MPO & Ours ($\alpha = 1.0$) \\
    \hline
    Early stage of epoch: 10K & & \\
    \hline    
      Average total rewards   & 325   $\pm$ 226 & \textbf{615 $\pm$ 45.76} \\
     \hline
     Average feedback normalized  & \textbf{3 $\pm$ 0.33} & 60.6 $\pm$ 1.74 \\
    \hline
     Average feedforward normalized  & 7.76 $\pm$ 0.84 & \textbf{0.35 $\pm$ 0.07}  \\
    \hline
%     Average energy normalized  & 969 $\pm$ 248 & \textbf{798 $\pm$ 84}   \\
        Average energy normalized  & 9.69 $\pm$ 2.48 & \textbf{7.98 $\pm$ 0.84}   \\
    \hline
    Middle stage of epoch: 100K & & \\
    \hline    
      Average total rewards    & 424   $\pm$ 108 & \textbf{1453 $\pm$ 96} \\
     \hline
     Average feedback normalized  & 88 $\pm$ 10.4 & \textbf{49.9 $\pm$ 2.34} \\
    \hline
     Average feedforward normalized  & 0.15 $\pm$ 0.04 & \textbf{0.09 $\pm$ 0.02}  \\
    \hline
%     Average energy normalized  & \textbf{412 $\pm$ 250} & 992 $\pm$ 295   \\
        Average energy normalized  & \textbf{4.12 $\pm$ 2.50} & 9.92 $\pm$ 2.95   \\

    \hline
    Final stage of epoch: 200K & & \\
     \hline   
      Average total rewards   & 2399   $\pm$ 694 & \textbf{3038 $\pm$ 148} \\
     \hline
     Average feedback normalized  & \textbf{129 $\pm$ 5.25} & 210 $\pm$ 10 \\
    \hline
     Average feedforward normalized & \textbf{ 0.36 $\pm$ 0.08} & 0.37 $\pm$ 0.06  \\
    \hline
%     Average energy normalized  & 2250 $\pm$ 628 & \textbf{1860 $\pm$ 52.2}   \\
        Average energy normalized  & 22.5 $\pm$ 6.28 & \textbf{18.6 $\pm$ 0.52}   \\
    \hline        
  \end{tabular}
\end{center}
\end{table}
\begin{table}[hbt]
\caption{Meta-testing of Walker2D} \label{table:Walker2D:meta-testing}
\begin{center}
  \small
  \begin{tabular}{|l|c|c|}
    \hline
           & MB-MPO & Ours ($\alpha = 1.0$) \\
    \hline
    Early stage of epoch: 10K & &\\
    \hline    
      Average total rewards   & 51   $\pm$ 29 & \textbf{45 $\pm$ 7.91} \\
     \hline
     Average feedback normalized  & \textbf{4.62 $\pm$ 0.21} & 3.1 $\pm$ 0.7 \\
    \hline
     Average feedforward normalized   & 6.47 $\pm$ 0.25 & \textbf{3.05 $\pm$ 0.19}  \\
    \hline
%     Average energy normalized  & 500 $\pm$ 42 & \textbf{394 $\pm$ 29}   \\
        Average energy normalized  & 5.00 $\pm$ 0.42 & \textbf{3.94 $\pm$ 0.29}   \\
    \hline
    Middle stage of epoch: 100K & & \\
     \hline   
      Average total rewards   & 662   $\pm$ 194 & \textbf{729 $\pm$ 132} \\
     \hline
     Average feedback normalized  & 110 $\pm$ 7.1 & \textbf{77.8 $\pm$ 33.2} \\
    \hline
     Average feedforward normalized  & 0.78 $\pm$ 0.25 & \textbf{0.25 $\pm$ 0.02}  \\
    \hline
%     Average energy normalized  & \textbf{436 $\pm$ 214} & 487 $\pm$ 141   \\
        Average energy normalized  & \textbf{4.36 $\pm$ 2.14} & 4.87 $\pm$ 1.41   \\
    \hline        
     Final stage of epoch: 200K & & \\
    \hline
      Average total rewards   & 523   $\pm$ 174 & \textbf{1123 $\pm$ 152} \\
     \hline
     Average feedback normalized  & \textbf{150 $\pm$ 29} & 197 $\pm$ 10.6 \\
    \hline
     Average feedforward normalized  & \textbf{0.01 $\pm$ 0.06} & 0.07 $\pm$ 0.07  \\
    \hline
%     Average energy normalized  & \textbf{624 $\pm$ 106} & 1230 $\pm$ 323   \\
        Average energy normalized  & \textbf{6.24 $\pm$ 1.06} & 12.30 $\pm$ 3.23   \\
    \hline        
  \end{tabular}
\end{center}
\end{table}

\begin{figure}[hbt]
\begin{center}
\includegraphics[angle=0, width=1.0\textwidth]{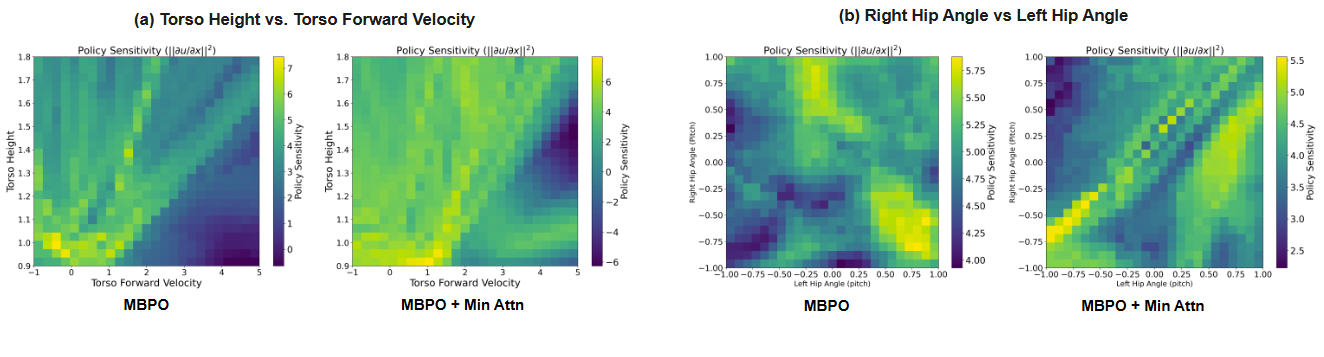}
\end{center}
\caption{Humanoid: Heatmap of feedback. The feedback ($||\partial u / \partial x||^2$) in the projection of the states to (a) Torso-height versus Torso-forward-velocity, and (b) Right-hip angle versus left-hip angle.}
\label{fig:humanoid_heatmap}
\end{figure}

\begin{figure}[hbt]
\begin{center}
\includegraphics[angle=0, width=1.0\textwidth]{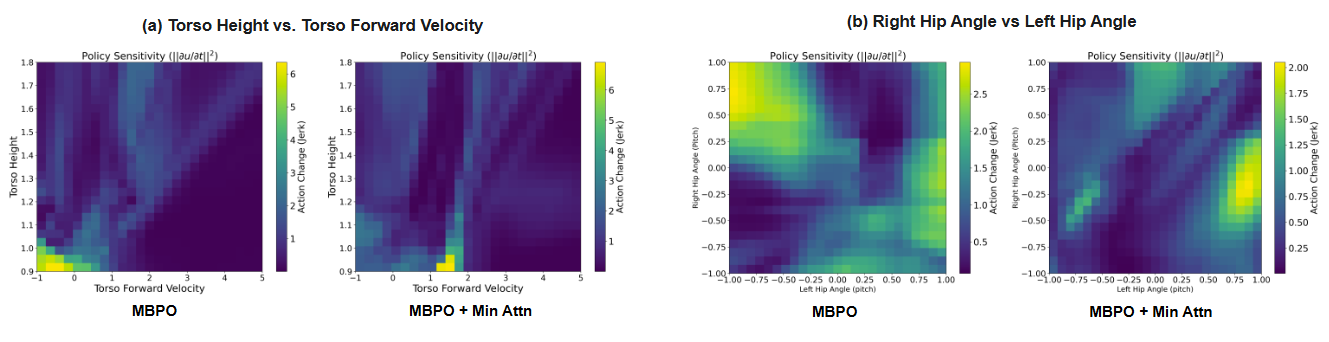}
\end{center}
\caption{Humanoid: Heatmap of feedforward. The feedforward ($||\partial u / \partial t||^2$) in the projection of the states to a) Torso-height versus Torso-forward-velocity, and b) Right-hip angle versus left-hip angle.}
\label{fig:humanoid_heatmap_jerk}
\end{figure}

\begin{figure}[hbt]
\centering
%\begin{subfigure}[b]{0.325\textwidth}
\begin{subfigure}[b]{0.48\textwidth}  
   \centering
   \includegraphics[width=\textwidth]{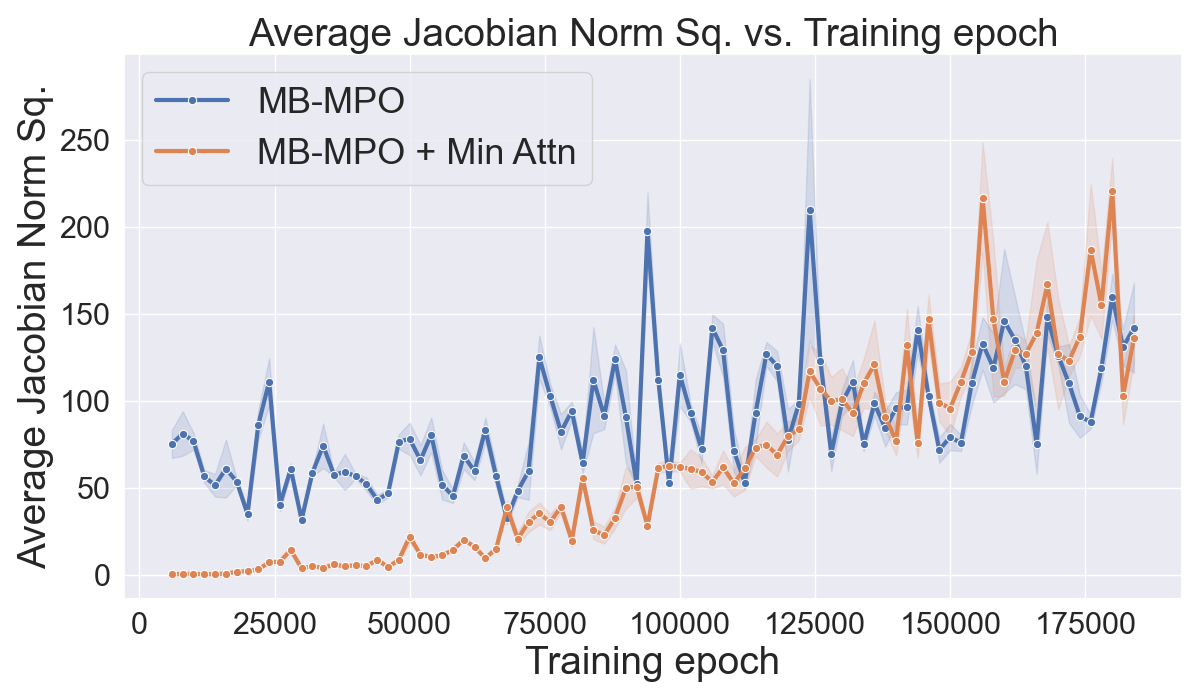}
   \label{fig:figure3a}
\end{subfigure}
\hfill 
%\begin{subfigure}[b]{0.325\textwidth}
\begin{subfigure}[b]{0.48\textwidth} 
   \centering
   \includegraphics[width=\textwidth]{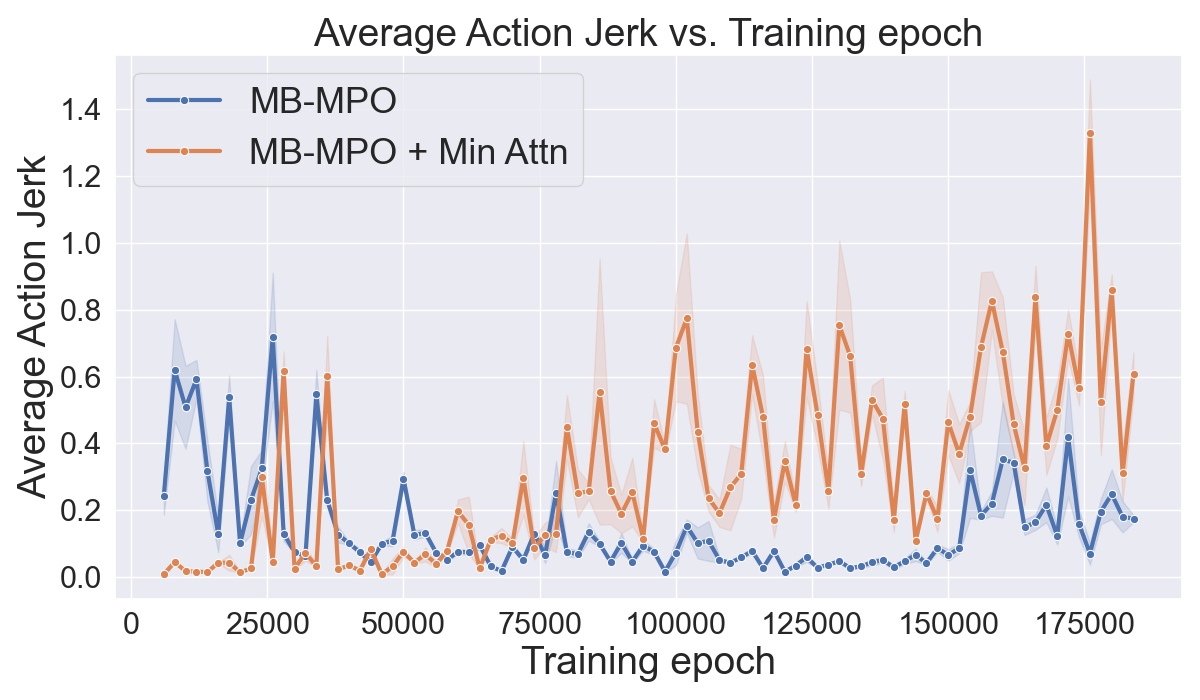} 
   \label{fig:figure3b}
\end{subfigure}
\hfill 
%\begin{subfigure}[b]{0.325\textwidth}
\begin{subfigure}[b]{0.48\textwidth} 
   \centering
   \includegraphics[width=\textwidth]{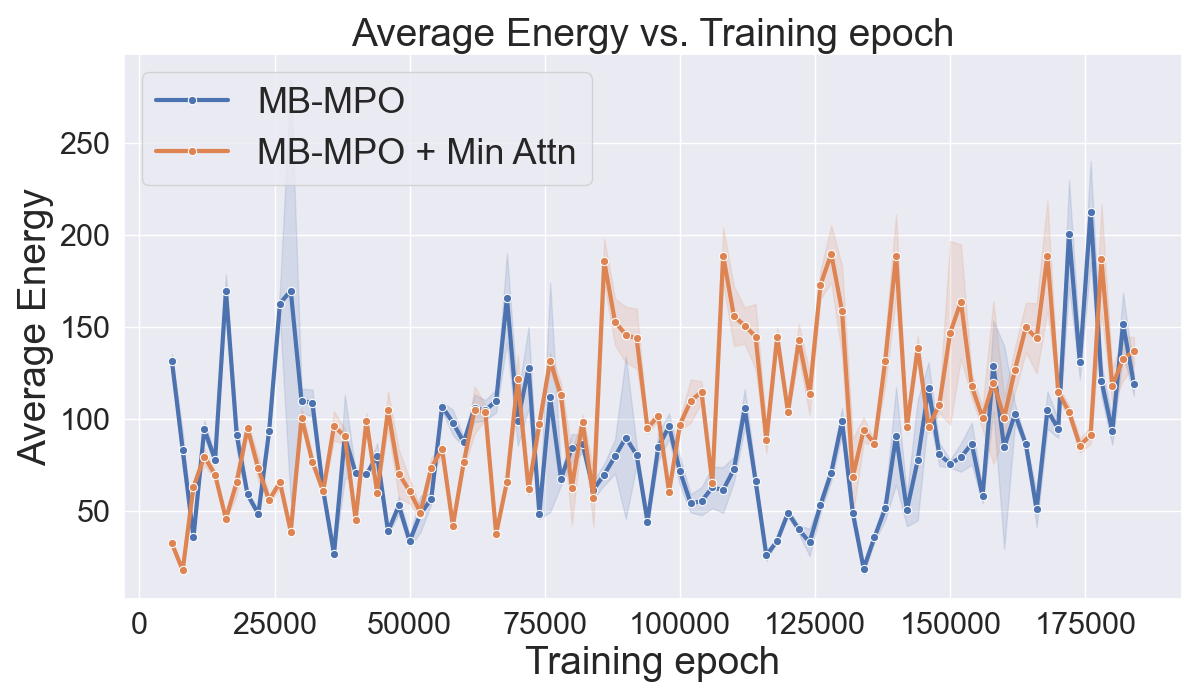} 
   \label{fig:heatmap_humanoid}
\end{subfigure}
\caption{Humanoid: Time profiles for the feedback, feedforward, and energy in meta-training. The comparison of feedback ($||\partial u / \partial x||^2$, Jacobian), feedforward ($||\partial u/ \partial t||^2$, jerk), and energy ($||u||^2$) between MB-MPO and MB-MPO with minimum attention ($\alpha = 1.0$).}
\label{time_profile_humanoid}
\end{figure}

\subsection{Humanoid (Figure \ref{fig:humanoid_heatmap}, \ref{fig:humanoid_heatmap_jerk}, \ref{time_profile_humanoid}, Table \ref{humanoid:meta_training}, \ref{humanoid:meta_testing})}

For the highly complex Humanoid, the regularization drastically simplifies the control problem by encouraging the discovery of more efficient and specialized gait mechanics.
\begin{itemize}
    \item \textbf{(a) Torso height vs. forward velocity:} The vanilla policy exhibits scattered, high-sensitivity ``islands,'' indicating a complex, reactive strategy. Our regularized policy discovers a remarkably simpler solution: its sensitivity is almost entirely focused in a narrow band at maximum torso height. This corresponds to the critical transition phase at the apex of a step, allowing the policy to be ``calm'' and smooth everywhere else, demonstrating a highly efficient gait.
    \item \textbf{(b) Hip angle coordination:} The vanilla policy shows broad sensitivity during stride transitions (when legs are in opposition). Our regularized policy dramatically focuses this sensitivity into a single, sharp hotspot. This suggests a shift from a novice, generally reactive gait to an expert one that applies precise control only at the most critical moment of weight transfer and power generation, thereby increasing efficiency and stability.
\end{itemize}

\begin{table}[hbt]
\caption{Humanoid: Meta-training} \label{humanoid:meta_training}
\begin{center}
  \small
  \begin{tabular}{|l|c|c|}
    \hline
           & MB-MPO & Ours ($\alpha = 1.0$) \\
    \hline
    Early stage of epoch: 10K & & \\
    \hline    
      Average total rewards   & 297   $\pm$ 10 & \textbf{245 $\pm$ 10} \\
     \hline
     Average feedback normalized  & 39.9 $\pm$ 1.7 & \textbf{2.48 $\pm$ 0.13} \\
    \hline
     Average feedforward normalized  & \textbf{0.13 $\pm$ 0.006} & 0.877 $\pm$ 0.06  \\
    \hline
%     Average energy normalized  & \textbf{108 $\pm$ 3.58} & 52.1 $\pm$ 1.6   \\
        Average energy normalized  & \textbf{1.08 $\pm$ 0.036} & 0.52 $\pm$ 0.016   \\
    \hline
    Middle stage of epoch: 100K & & \\
    \hline    
      Average total rewards    & \textbf{332   $\pm$ 26} & 267 $\pm$ 82 \\
     \hline
     Average feedback normalized  & 119.2 $\pm$ 3.82 & \textbf{57.5 $\pm$ 10.8} \\
    \hline
     Average feedforward normalized  & \textbf{0.09 $\pm$ 0.01} & 0.42 $\pm$ 0.07  \\
    \hline
%     Average energy normalized  & \textbf{64.3 $\pm$ 4.99} & 107 $\pm$ 8.89   \\
        Average energy normalized  & \textbf{0.64 $\pm$ 0.050} & 1.07 $\pm$ 0.089   \\
    \hline
    Final stage of epoch: 200K & & \\
     \hline   
      Average total rewards   & \textbf{575   $\pm$ 177} & 498 $\pm$ 38 \\
     \hline
     Average feedback normalized  & 187 $\pm$ 16 & \textbf{79.1 $\pm$ 5.49} \\
    \hline
     Average feedforward normalized  & 0.516 $\pm$ 0.05 & \textbf{0.198 $\pm$ 0.01}  \\
    \hline
%     Average energy normalized  & 144 $\pm$ 42.4 & \textbf{136 $\pm$ 8.95}   \\
        Average energy normalized  & 1.44 $\pm$ 0.42 & \textbf{1.36 $\pm$ 0.090}   \\
    \hline        
  \end{tabular}
\end{center}
\end{table}
\begin{table}[hbt]
\caption{Meta-testing of Humanoid}  \label{humanoid:meta_testing}
\begin{center}
  \small
  \begin{tabular}{|l|c|c|}
    \hline
           & MB-MPO & Ours ($\alpha = 1.0$) \\
    \hline
    Early stage of epoch: 10K & &\\
    \hline    
      Average total rewards   & 314   $\pm$ 19 & \textbf{359 $\pm$ 27} \\
     \hline
     Average feedback normalized  & 59.1 $\pm$ 5.24 & \textbf{15.9 $\pm$ 1.06} \\
    \hline
     Average feedforward normalized   & 0.285 $\pm$ 0.003 & \textbf{0.136 $\pm$ 0.002}  \\
    \hline
%     Average energy normalized  & 101 $\pm$ 8.21 & \textbf{53.3 $\pm$ 0.811}   \\
        Average energy normalized  & 1.01 $\pm$ 0.082 & \textbf{0.053 $\pm$ 0.0081}   \\
    \hline
    Middle stage of epoch: 100K & & \\
     \hline   
      Average total rewards   & 351   $\pm$ 91 & \textbf{395 $\pm$ 27} \\
     \hline
     Average feedback normalized  & 76.3 $\pm$ 13.6 & \textbf{36 $\pm$ 4.32} \\
    \hline
     Average feedforward normalized  & \textbf{0.13 $\pm$ 0.03} & 0.16 $\pm$ 0.09  \\
    \hline
%     Average energy normalized  & \textbf{73.1 $\pm$ 21.9} & 140 $\pm$ 9.42   \\
        Average energy normalized  & \textbf{0.73 $\pm$ 0.22} & 1.40 $\pm$ 0.094   \\

    \hline        
     Final stage of epoch: 200K & & \\
    \hline
      Average total rewards   & 315   $\pm$ 67 & \textbf{580 $\pm$ 34} \\
     \hline
     Average feedback normalized  & 277 $\pm$ 64.1 & \textbf{76.3 $\pm$ 11.7} \\
    \hline
     Average feedforward normalized  & 0.332 $\pm$ 0.11 & \textbf{0.09 $\pm$ 0.01}  \\
    \hline
%     Average energy normalized  & \textbf{11 $\pm$ 16} & 188 $\pm$ 20.8  \\
        Average energy normalized  & \textbf{0.11 $\pm$ 0.16} & 1.88 $\pm$ 0.21  \\
    \hline        
  \end{tabular}
\end{center}
\end{table}


\end{document}